%% file: main.tex
\pdfoutput=1
\documentclass[11pt]{article}
\input{glyphtounicode}
\usepackage[dvipsnames]{xcolor}
\usepackage{algorithm}
\usepackage{algpseudocode}
\usepackage{newtxtext}
\usepackage[left=1.25in, top=1in, bottom=1in, right=1.25in]{geometry}
\usepackage[USenglish]{babel}
\usepackage[utf8]{inputenc}
\usepackage[T1]{fontenc}
\usepackage{csquotes}
\usepackage{url}
\usepackage{booktabs}
\usepackage{amsfonts}
\usepackage{amsmath}
\usepackage{amssymb}
\usepackage{mathtools}
\usepackage{amsthm}
\usepackage{bm}
\usepackage{nicefrac}
\usepackage{microtype}
\DisableLigatures[i,I]{encoding = T1}
\usepackage{enumitem}
\usepackage{wrapfig}
\usepackage{dsfont}
\usepackage{graphicx}
\usepackage{subcaption}
\usepackage[most]{tcolorbox}
\usepackage{natbib}
\definecolor{myblue}{rgb}{0,0.08,0.75}
\usepackage[colorlinks,citecolor=myblue,linkcolor=orange,urlcolor=myblue]{hyperref}
\usepackage{xurl}
\usepackage[capitalize,noabbrev]{cleveref}

\newtcolorbox{wrongbox}[1][]{
  enhanced,
  colback=red!3,
  colframe=red!50!black,
  boxrule=0pt,
  leftrule=3pt,
  arc=0pt,
  outer arc=0pt,
  left=8pt,
  right=8pt,
  top=6pt,
  bottom=6pt,
  fonttitle=\bfseries,
  coltitle=red!50!black,
  attach title to upper,
  after title={:\ },
  title={#1}
}

\newtcolorbox{correctbox}[1][]{
  enhanced,
  colback=green!3,
  colframe=green!50!black,
  boxrule=0pt,
  leftrule=3pt,
  arc=0pt,
  outer arc=0pt,
  left=8pt,
  right=8pt,
  top=6pt,
  bottom=6pt,
  fonttitle=\bfseries,
  coltitle=green!50!black,
  attach title to upper,
  after title={:\ },
  title={#1}
}

\newtcolorbox{abstainbox}[1][]{
  enhanced,
  colback=blue!3,
  colframe=blue!50!black,
  boxrule=0pt,
  leftrule=3pt,
  arc=0pt,
  outer arc=0pt,
  left=8pt,
  right=8pt,
  top=6pt,
  bottom=6pt,
  fonttitle=\bfseries,
  coltitle=blue!50!black,
  attach title to upper,
  after title={:\ },
  title={#1}
}

\newtcolorbox{abandonbox}[1][]{
  enhanced,
  colback=orange!5,
  colframe=orange!70!black,
  boxrule=0pt,
  leftrule=3pt,
  arc=0pt,
  outer arc=0pt,
  left=8pt,
  right=8pt,
  top=6pt,
  bottom=6pt,
  fonttitle=\bfseries,
  coltitle=orange!70!black,
  attach title to upper,
  after title={:\ },
  title={#1}
}

\newtcolorbox{examplebox}[1][]{
  enhanced,
  colback=gray!4,
  colframe=gray!50,
  boxrule=0pt,
  leftrule=3pt,
  arc=0pt,
  outer arc=0pt,
  left=8pt,
  right=8pt,
  top=6pt,
  bottom=6pt,
  fonttitle=\bfseries,
  coltitle=gray!70!black,
  title={#1}
}

\newtcolorbox{promptbox}[1][]{
  enhanced,
  colback=gray!2,
  colframe=gray!45,
  boxrule=0.4pt,
  arc=0pt,
  outer arc=0pt,
  left=8pt,
  right=8pt,
  top=6pt,
  bottom=6pt,
  fonttitle=\bfseries,
  coltitle=black,
  attach title to upper,
  after title={:\ },
  title={#1}
}

\newtcolorbox{casebox}[1][]{
  enhanced,
  colback=gray!2,
  colframe=gray!45,
  boxrule=0.4pt,
  arc=0pt,
  outer arc=0pt,
  left=8pt,
  right=8pt,
  top=6pt,
  bottom=6pt,
  fonttitle=\bfseries,
  coltitle=black,
  attach title to upper,
  after title={:\ },
  title={#1}
}

\theoremstyle{plain}

\theoremstyle{definition}

\theoremstyle{remark}

\title{Honesty over Accuracy:\\Trustworthy Language Models through Reinforced Hesitation}
\date{}
\author{
Mohamad Amin Mohamadi\thanks{%
    Toyota Technological Institute at Chicago
    \; $^\dagger$University of California, San Diego
\\Correspondence to \texttt{mohamadamin@ttic.edu}.%
    }
\quad Tianhao Wang\footnotemark[2]
\quad Zhiyuan Li\footnotemark[1]
}

\begin{document}
\maketitle
\input{sections/abstract}
\input{sections/introduction}
\input{sections/background}
\input{sections/method}
\input{sections/inference_compute}
\input{sections/conclusion}
\input{sections/acknowledgements}
\bibliography{bibliography}

\newpage
\appendix
\section*{Appendix Contents}

\vspace{-1mm}
\begingroup
\normalsize
\setlength{\parskip}{0pt}
\makeatletter
\newcommand{\appcontentssection}[3]{%
  \par\noindent\hb@xt@\textwidth{%
    \hyperref[#3]{\makebox[1.8em][l]{\textbf{#1}}\textbf{#2}}%
    \hfil\hyperref[#3]{\textbf{\pageref*{#3}}}}%
  \par\vspace{1.2mm}%
}
\newcommand{\appcontentssubsection}[3]{%
  \@dottedtocline{2}{1.8em}{3.2em}{\hyperref[#3]{#1\quad #2}}{\hyperref[#3]{\pageref*{#3}}}%
}
\appcontentssection{A}{Related Work}{sec:related_work}
\appcontentssection{B}{Experimental Setup and Training Implementation}{sec:appendix_training_implementation}
\appcontentssubsection{B.1}{Training Suites and Shared Configuration}{sec:appendix_hyperparameters}
\appcontentssubsection{B.2}{Knights \& Knaves Dataset Construction}{sec:appendix_dataset}
\appcontentssubsection{B.3}{MATH Dataset, Protocol, and Checkpoint Selection}{sec:appendix_math_protocol}
\appcontentssubsection{B.4}{Training Prompts}{sec:appendix_training_prompts}
\appcontentssection{C}{Penalty-Blind Prompting Study Details}{sec:appendix_frontier_eval}
\appcontentssubsection{C.1}{Evaluation Prompts}{sec:appendix_prompts}
\appcontentssubsection{C.2}{MedQA, GPQA, and All-Dataset Results}{sec:appendix_extended_results}
\appcontentssubsection{C.3}{Evaluation Configuration}{sec:appendix_frontier_config}
\appcontentssubsection{C.4}{Uncertainty Acknowledgment and Abstention Decisions}{sec:appendix_uncertainty}
\appcontentssection{D}{Fixed-Penalty RH Diagnostics}{sec:appendix_fixed_penalty_diagnostics}
\appcontentssubsection{D.1}{Knights \& Knaves Behavioral Evolution}{sec:appendix_kk_behavioral_evolution}
\appcontentssubsection{D.2}{Knights \& Knaves Response Length and Format Diagnostics}{sec:appendix_format_penalty}
\appcontentssubsection{D.3}{MATH Fixed-Penalty Results}{sec:appendix_math_extra_results}
\appcontentssection{E}{Baselines and Inference Diagnostics}{sec:appendix_baselines_inference}
\appcontentssubsection{E.1}{RLCR Comparison on MATH}{sec:appendix_rlcr_comparison}
\appcontentssubsection{E.2}{RH vs RLCR: Capacity Allocation and Equivalence Analysis}{sec:appendix_rlcr_objective}
\appcontentssubsection{E.3}{Knights \& Knaves Majority-Vote Baseline}{sec:appendix_kk_majority_vote}
\appcontentssubsection{E.4}{MATH Inference Baselines}{sec:appendix_math_inference}
\appcontentssubsection{E.5}{MATH Inference Methodology Disclosures and Diagnostics}{sec:appendix_math_inference_diagnostics}
\appcontentssection{F}{Dynamic Penalty Conditioning on MATH}{sec:appendix_math_dynamic}
\appcontentssection{G}{Qualitative Analysis of Learned Abstention Behavior on Knights \& Knaves}{sec:appendix_qualitative}
\appcontentssubsection{G.1}{From Broad Answering to Calibrated Abstention}{sec:appendix_qualitative_broad_answering}
\appcontentssubsection{G.2}{Selective Coverage at High Penalties}{sec:appendix_qualitative_high_penalties}
\appcontentssubsection{G.3}{How Models Frame Their Abstentions}{sec:appendix_qualitative_framing}
\appcontentssubsection{G.4}{Summary}{sec:appendix_qualitative_summary}
\appcontentssection{H}{Medical QA Training Suite}{sec:appendix_medical_suite}
\appcontentssubsection{H.1}{Medical Suite Details}{sec:appendix_medical}
\appcontentssection{I}{Targeting of Learned Abstentions}{sec:appendix_abstention_precision}
\makeatother
\endgroup
\vspace{-1mm}

\input{appendix/related_work}
\input{appendix/appendix}
\input{appendix/medical}

\input{appendix/abstention_precision}

\end{document}

%% file: sections/abstract.tex
\begin{abstract}
    Modern language models fail a fundamental requirement of trustworthy intelligence: knowing when not to answer. Despite achieving impressive accuracy on benchmarks, these models produce confident hallucinations, even when wrong answers carry catastrophic consequences. Our evaluations on GSM8K, MedQA and GPQA show frontier models almost never abstain despite explicit warnings of severe penalties, suggesting that prompts cannot override training that rewards any answer over no answer. As a remedy, we propose Reinforced Hesitation (RH): a modification to Reinforcement Learning from Verifiable Rewards (RLVR) to use ternary rewards (+1 correct, 0 abstention, -$\lambda$ error) instead of binary. Controlled experiments on logic puzzles reveal that varying $\lambda$ produces distinct models along a Pareto frontier, where each training penalty yields the optimal model for its corresponding risk regime: low penalties produce aggressive answerers, high penalties conservative abstainers. The same frontier holds on MATH Levels 4--5 and on medical QA, where it transfers to an unseen dataset. We then introduce two inference strategies that exploit trained abstention as a coordination signal: cascading routes queries through models with decreasing risk tolerance, while self-cascading re-queries the same model on abstention. Both outperform majority voting  with lower computational cost. These results establish abstention as a first-class training objective that transforms ``I don't know'' from failure into a coordination signal, enabling models to earn trust through calibrated honesty about their limits.
\end{abstract}

%% file: sections/introduction.tex
\section{Introduction}
\vspace{-2mm}

Language models are increasingly embedded in high-stakes workflows from medical diagnosis \citep{Thirunavukarasu2023LLMMedicine,li2024mediq} to financial advisory \citep{Wu2023BloombergGPT}, legal research \citep{Guha2023LegalBench}, and infrastructure control, where a single confidently stated falsehood can outweigh many correct predictions \citep{Krishnan2024EnhancingTrust,Yang2024CanWeTrust,liu2024trustworthy,mazeika2024harmbench}. Yet current evaluation and training paradigms remain fixated on maximizing accuracy \citep{srivastava2022beyond,Hendrycks2021MMLU,Glazer2024FrontierMath}, treating mistakes as equivalent whether the model is solving trivia or advising a medical intervention. Reinforcement learning from verifiable rewards (RLVR) \citep{DeepSeekAI2025R1,lambert2025tulu3,Muennighoff2025s1,jaech2024o1,google2025gemini25Pro} reinforces this misalignment: models receive +1 for correct answers and 0 for wrong ones, so guessing is never worse than abstaining under the training reward \citep{chen2025reasoningfaithfulness,lin-etal-2022-truthfulqa,Kalai2025WhyHallucinate}. The binary signal also ignores verification costs that range from milliseconds for trivia to hours for formal proofs \citep{chen2025seedproverdeepbroadreasoning} or expert review for medical decisions. This misalignment between training objectives and deployment requirements has created systems that race up leaderboards while failing the basic requirement of trustworthy intelligence: \emph{knowing when not to answer} \citep{RS88b}.

\begin{figure*}[t]
    \centering
    \begin{subfigure}[c]{0.30\textwidth}
        \centering
        \includegraphics[width=.9\textwidth]{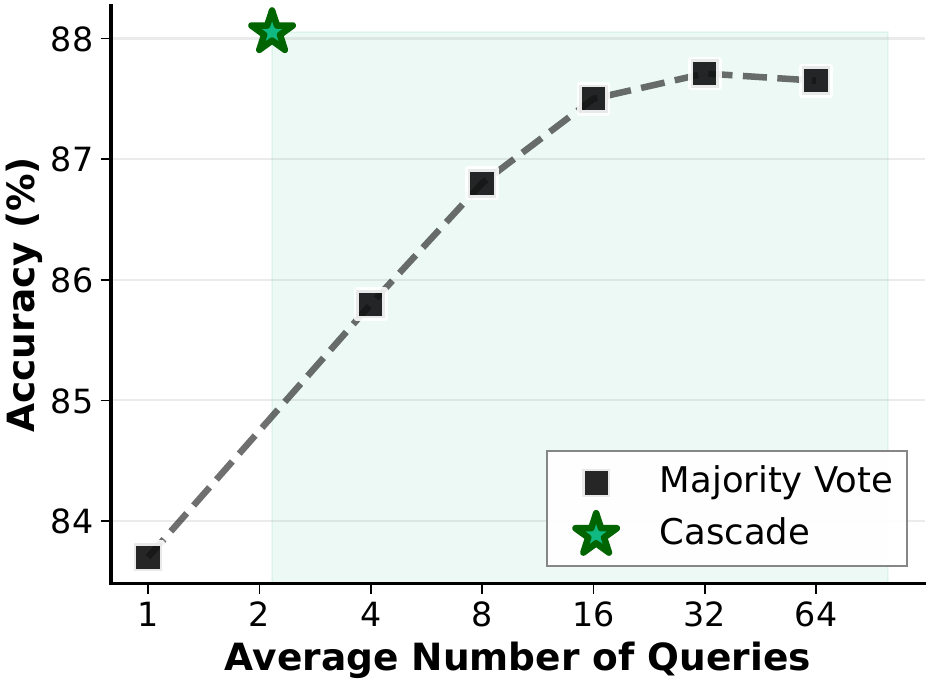}
        \label{fig:cascading}
        \hspace{1mm}
        \includegraphics[width=.9\textwidth]{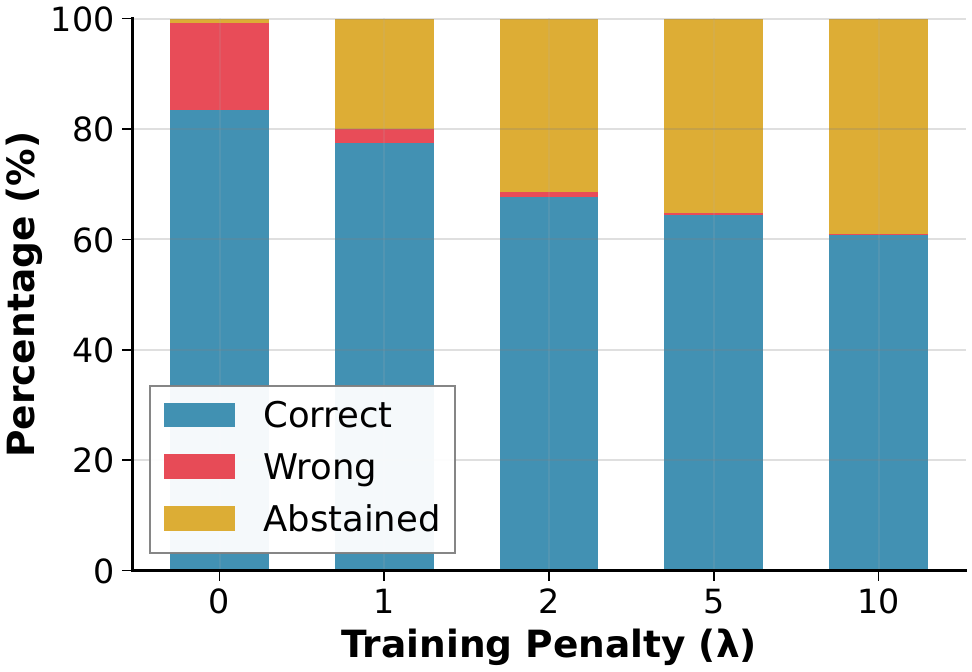}
        \label{fig:pareto_conditional}
    \end{subfigure}
    \begin{subfigure}[c]{0.48\textwidth}
        \centering
        \vspace{4mm}
        \includegraphics[width=\textwidth]{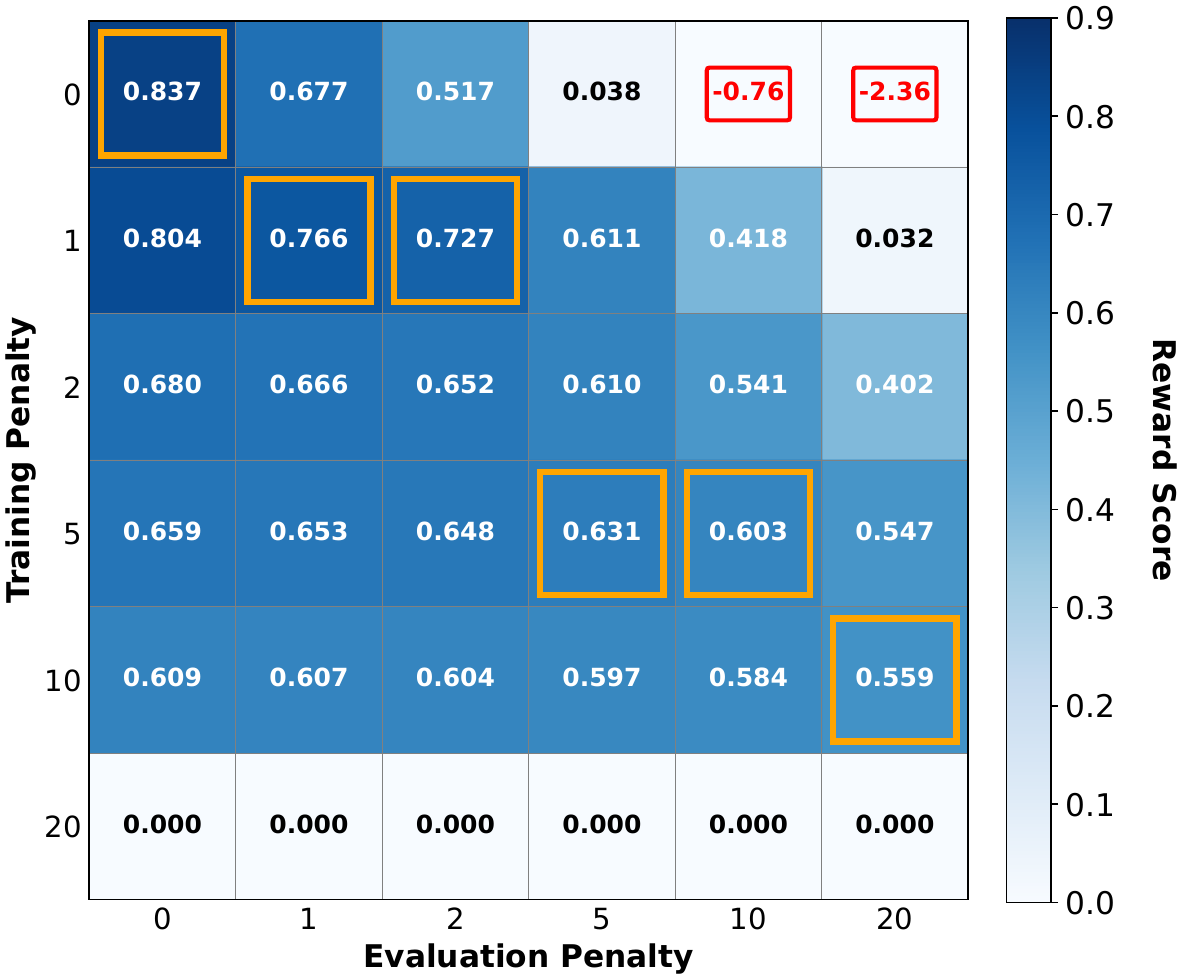}
        \label{fig:pareto_heatmap}
    \end{subfigure}
    \caption{\textbf{Reinforced Hesitation creates an ordered risk frontier on Knights \& Knaves, reducing error rates through calibrated abstentions and enabling adaptive inference.} All panels in this summary figure use Knights \& Knaves. \textbf{Right:} Cross-penalty evaluation reveals risk specialization across our model family: each model achieves superior performance under specific evaluation penalties, with optimal models (orange) clustering near the diagonal where training and evaluation contexts align. This demonstrates that each training penalty produces a useful operating point for a corresponding risk regime. \textbf{Top left:} Cascading through models with decreasing risk aversion ($\lambda=10\rightarrow5\rightarrow2\rightarrow1\rightarrow0$) achieves efficient triage where each specialist handles problems matching its confidence regime. \textbf{Bottom left:} Models trained with different penalties form a frontier where higher $\lambda$ achieves both lower error rates and higher conditional accuracy.}
    \label{fig:pareto_frontier}
    \vspace{-3mm}
\end{figure*}

This misalignment raises a natural question: can instruction-following alone induce appropriate abstention? To test this, in \hyperref[sec:penalty_blind]{\textbf{Section 2}}, we evaluate a series of frontier models on standard benchmarks with explicit abstention instructions and severe penalty warnings. The results reveal a structural failure: these models typically abstain less than 1\% of the time while maintaining error rates above 10\%, even when abstention would mathematically dominate. Consistent with \citep{kirichenko2025abstentionbenchreasoningllmsfail, yao-etal-2025-are, Kapoor2024LLMMustBeTaught, wu2025answer, tong2025measuring}, RLVR-trained models perform worse at abstention than their base counterparts, even when explicitly recognizing uncertainty in reasoning chains. These results suggest that the lack of effective abstention is not a missing capability but an ingrained behavior: prompts cannot override gradient-driven priors from thousands of training steps promoting any answer over no answer. This failure reveals that effective abstention requires training-time intervention, not inference-time instruction.

In \hyperref[sec:rh_method]{\textbf{Section 3}}, we propose \textbf{Reinforced Hesitation} (RH): a minimal modification that addresses the problem at its source during training. By transforming RLVR's correct-versus-wrong reward into a ternary structure $(+1, 0, -\lambda)$ for correct answers, abstentions, and wrong answers respectively, we make hesitation explicitly valuable rather than merely possible. The penalty parameter $\lambda\geq0$ encodes both domain-specific consequences and verification costs, explicitly trading off mistakes against abstentions \citep{sayedi2010trading}: high values for domains where wrong answers are costly, low values for tasks where exploration matters. Through a series of controlled RLVR experiments with Qwen3-1.7B on a dataset of Knights \& Knaves logic puzzles of varying complexity, we demonstrate that this teaches models to develop discrimination between problems they can reliably solve and those where guessing would be reckless: models trained with $\lambda=1$ selectively abstain on 60\% of logically complex problems while abstaining on only 10\% of simpler ones, reducing overall error rates from 15\% to below 2\% in comparison to the baseline RLVR training paradigm. We further test whether this behavior extends beyond synthetic logic. On MATH with Qwen3.5-4B, increasing $\lambda$ reduces wrong answers from 11.4\% to 2.75\% and raises conditional accuracy from 88.6\% to 95.8\%, at the cost of coverage. The same frontier holds on medical QA (MedMCQA), including transfer to the unseen MedQA test set (Appendix~\ref{sec:appendix_medical_suite}). Across fixed-penalty specialists, increasing $\lambda$ produces distinct behavioral regimes: aggressive answering with persistent errors ($\lambda=0$), balanced trade-offs with calibrated abstention ($\lambda \in \{1,2,5\}$), and conservative abstention with near-zero errors ($\lambda \geq 10$), each representing a valid specialist for different deployment needs. We also study a single prompt-conditioned $\lambda$ model as a practical alternative; it responds monotonically but weakly to the prompted penalty and does not match the specialist frontier.

These abstentions are not terminal failures but exploitable coordination signals. When a model says "I don't know," it precisely identifies problems beyond its confidence boundary. We can leverage this information to route queries to appropriate specialists or alternative approaches, transforming uncertainty into productive collaboration.
In \hyperref[sec:rh_inference]{\textbf{Section 4}}, we demonstrate how to exploit these abstention signals through two inference strategies. \textbf{Cascading} routes queries through models trained with decreasing risk tolerance ($\lambda=10 \to \cdots \to 0$) where each abstention triggers delegation to the next specialist. This architectural approach achieves 88.1\% accuracy with only 2.2 average queries on Knights \& Knaves and reaches 90.0\% on MATH at 1.58 average queries, close to 16-sample majority voting at much lower query cost. \textbf{Self-cascading} exploits the inherent nondeterminism in language model inference: when a model abstains, we re-query the same model with the same prompt, allowing alternative reasoning paths to emerge through random sampling in autoregressive models. Through this method, a model trained with $\lambda=1$ improves from 77.5\% to 92.5\% accuracy on Knights \& Knaves, with smaller but useful gains on MATH. Both approaches demonstrate how abstention is not terminal failure but a coordination mechanism enabling adaptive computation. In a broader perspective, an abstention not only enables collaboration between models, but ultimately between AI systems and human experts, which can be far more valuable. Related work is discussed in Appendix~\ref{sec:related_work}.

%% file: sections/background.tex
\section{Penalty-Blind Abstention in Frontier Models}\label{sec:penalty_blind}
\vspace{-2mm}

\begin{figure*}[t]
  \centering
  \includegraphics[width=\textwidth]{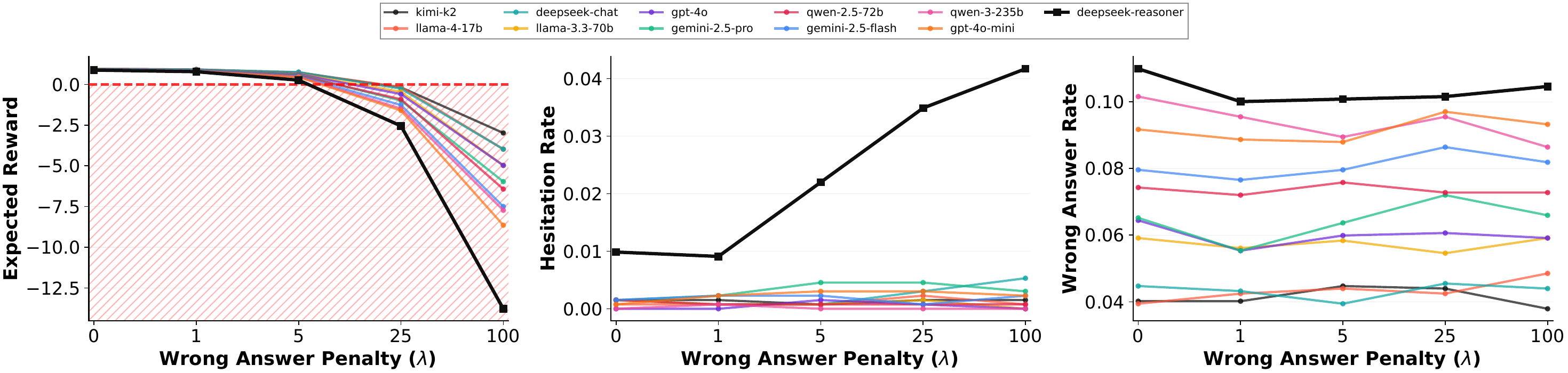}
  \caption{\textbf{Penalty sensitivity of frontier models on GSM8K.}
\textbf{Left:} Expected reward $r(\lambda)=p(\text{correct})-\lambda\,p(\text{wrong})$ for $\lambda\in\{1,5,25,100\}$; red dashed line marks $r=0$ baseline.
\textbf{Middle:} Frontier models rarely choose to abstain, even when faced with penalties of magnitude 100. \textbf{Right:} Despite the high penalty values, the rate of wrong answers remains high across various models.}
    \label{fig:frontier_eval_gsm8k}
\end{figure*}

Modern post-training has a shared blind spot. RLHF optimizes learned human preferences, while RLVR maximizes sparse terminal rewards on verifiable tasks; neither directly trains \emph{when not to answer}. RLHF can treat abstention as unhelpful, and standard RLVR gives no special benefit to abstention over a wrong answer. The outcome is similar: models learn fluent reasoning without a reliable policy for admitting uncertainty.

Can instruction-following supply this missing policy? We tell models the wrong-answer penalty, allow abstention, and ask them to maximize expected score. Prior work studies prompting-based uncertainty and abstention \citep{Kapoor2024LLMMustBeTaught, wu2025answer, yunis2024reducing}, but explicit penalty sweeps across recent RLVR-trained models remain limited. We evaluate 11 frontier models on GSM8K \citep{Cobbe2021TrainingVerifiers}, MedQA \citep{jin2021medqa}, and GPQA \citep{Rein2024GPQA}; models rarely convert stated penalties into reliable abstention.

\textbf{Setup.} We evaluate 11 frontier models on GSM8K, MedQA, and GPQA under $\lambda \in \{1,5,25,100\}$ plus a no-penalty control; exact prompt modifications and evaluation details are in Appendix~\ref{sec:appendix_prompts}.
\vspace{-2mm}

\textbf{Results.} The primary GSM8K results are shown in \Cref{fig:frontier_eval_gsm8k}; MedQA/GPQA and all-dataset plots are in Appendix~\ref{sec:appendix_extended_results}. Across models and penalty settings, \textbf{hesitation remains vanishingly rare while accuracy is largely invariant to the wrong-answer penalty}. Abstention rates stay near zero and wrong-answer rates remain effectively flat despite increasing \(\lambda\) from 1 to 100. A risk-sensitive policy should hesitate more as \(\lambda\) grows; instead expected reward quickly falls below the always-abstain baseline. This pattern holds across parameter scales and model families, suggesting that non-responsiveness to external stakes is structural rather than a missing capability.

Reasoning-tuned systems such as Gemini 2.5 Pro, Kimi-K2, and DeepSeek-Reasoner show no special advantage and can be among the least penalty-responsive. Their hesitation curves remain flat as \(\lambda\) grows, and in some cases accuracy degrades under higher penalties, suggesting that accuracy-maximization can override epistemic signals at inference time (Appendix~\ref{sec:appendix_uncertainty}). This aligns with observations that RLVR-style training can reduce abstention behavior \citep{kirichenko2025abstentionbenchreasoningllmsfail,yao-etal-2025-are,Kapoor2024LLMMustBeTaught,wu2025answer,tong2025measuring}.

The broad failure across models, scales, and penalties indicates this is not merely a prompt-format issue. Models can follow the format, acknowledge penalties, and sometimes reason about uncertainty, but they lack a reliable learned policy for deciding when abstention beats answering. Prompt-only penalty instructions are therefore insufficient, motivating the training-time intervention in \hyperref[sec:rh_method]{Section~3}, including direct RH training on MATH.
\vspace{-2mm}

%% file: sections/method.tex
\section{Reinforced Hesitation: Teaching Models When Not to Answer}\label{sec:rh_method}
\vspace{-2mm}

Reinforced Hesitation formalizes the intuition that hesitation should be valuable through a simple ternary reward: $+1$ for correct answers, $0$ for abstention (``I don't know''), and $-\lambda$ for wrong answers, where $\lambda \geq 0$ encodes the domain-specific cost of errors. Under these rewards, a rational agent abstains when expected utility falls below zero, at confidence threshold $\frac{\lambda}{1+\lambda}$. This makes $\lambda$ a control knob that can be set high for high-risk domains (e.g., $\lambda = 100$ for medical diagnosis) or low for low-risk tasks (e.g., $\lambda = 1$ for homework help), with $\lambda=0$ recovering the always-answering RLVR baseline.

\begin{algorithm}[!ht]
    \caption{Reinforced Hesitation RLVR}
    \label{alg:rh}
    \begin{algorithmic}[1]
    \Require Initial policy $\pi_0$, Dataset $\mathcal{D}$, Penalty $\lambda$, Iterations $T$
    \For{iteration $t = 1$ to $T$}
        \State Sample batch $B \sim \mathcal{D}$
        \For{each problem $p \in B$}
            \State $p' \gets p + $ abstention instruction
            \State Generate response: $y \sim \pi_t(\cdot|p')$
            \State Parse answer from $y$ using the task-specific output format
            \State Calculate reward: $r \gets R_{\text{total}}(y, y^*)$
        \EndFor
        \State Update $\pi_{t+1}$ with an RLVR optimizer using rewards $\{r\}_{p \in B}$
    \EndFor
    \State \Return $\pi_T$
    \end{algorithmic}
\end{algorithm}

We implement RH as a modification to the RLVR stage of post-training, restructuring the standard correct-versus-wrong reward into our ternary system without architectural changes (see Appendix~\ref{sec:appendix_hyperparameters} for full details). The total reward decomposes as $R_{\text{total}}(y, y^*) = R_{\text{content}}(y, y^*) + R_{\text{format}}(y)$, where the content reward evaluates answer accuracy:
\begin{equation}
R_{\text{content}}(y, y^*) = \begin{cases}
+1 & \text{if } y = y^* \text{ (correct answer)} \\
0 & \text{if } y = \text{``I don't know''} \\
-\lambda & \text{if } y \neq y^* \text{ (incorrect answer)}
\end{cases}
\end{equation}
and the format penalty $R_{\text{format}}(y) = -0.5\lambda$ for violations (0 otherwise) ensures proper output structure (see Appendix~\ref{sec:appendix_format_penalty} for the rationale). Algorithm~\ref{alg:rh} shows the complete procedure; the key modification is replacing binary rewards with our ternary structure.

To investigate the efficacy of this approach, we present a series of RLVR experiments that test how the penalty parameter $\lambda$ shapes model behavior. We use Knights \& Knaves as a controlled mechanism study and MATH as a larger, non-synthetic generalization setting. Our empirical investigation reveals transient dynamics where models initially learn to solve the problem and then learn to selectively abstain, ordered risk frontiers across penalties, a direct comparison to confidence-threshold abstention, and a limitation of naive prompt-conditioned dynamic penalties.

\subsection{Experimental Design}
\vspace{-1mm}

We validate RH through controlled experiments on Knights \& Knaves logic puzzles \citep{xie2025logicRL, xie2024memorization}, where ground-truth solutions enable clean evaluation of abstention decisions. These puzzles require models to maintain logical consistency across interdependent statements, with complexity scaling exponentially with the number of inhabitants (see Appendix~\ref{sec:appendix_dataset} for examples).

Our dataset contains 80,000 training and 10,000 test samples of 5, 6, and 7-person puzzles with a 2 to 1 ratio of easy to hard split based on logical complexity. We train Qwen3-1.7B \citep{qwenTeam2025qwen3} using Dr.GRPO \citep{anonymous2025drGRPO,DeepSeekAI2025R1} with identical hyperparameters\footnote{Hyperparameters for $\lambda = 0$ are slightly different due to a need for format control, this is discussed in detail in Appendix~\ref{sec:appendix_hyperparameters}.} across all conditions varying only the penalty $\lambda \in \{0, 1, 2, 5, 10, 20\}$. This controlled design isolates $\lambda$ as the sole causal factor, preventing confounds from architecture, data, or optimization variations. Details of these hyperparameters and the dataset preparation are provided in Appendix~\ref{sec:appendix_hyperparameters} and \ref{sec:appendix_dataset}.

For Knights \& Knaves, we enforce a two-part response format with \texttt{<think>...</think>} tags for reasoning and \texttt{<answer>...</answer>} tags for the final decision (either a solution or ``I don't know''). The prompt explicitly permits abstention (see Algorithm~\ref{alg:rh}). Format violations (missing tags, malformed answers, exceeding 4096 tokens) incur a $-0.5\lambda$ penalty.

We track a set of behavioral and computational metrics throughout training. Behavioral measures disaggregate responses into four categories: correct, wrong, I don't know (abstention), and format violations. Computational metrics include mean response length, clipping frequency (truncation at 4096 tokens), and parsing errors. For single-sample training diagnostics, decoding settings are held fixed; inference-scaling experiments explicitly use multiple rollouts. Full experimental details including prompts and hardware specifications appear in Appendix~\ref{sec:appendix_training_prompts} and \ref{sec:appendix_hyperparameters}.

\textbf{MATH setup.} To test whether RH extends beyond synthetic logic puzzles, we also train Qwen3.5-4B on MATH Levels 4--5 \citep{Hendrycks2021MATH}. The MATH runs use boxed final answers with the same answer/abstain reward, and the fixed-penalty specialists use $\lambda \in \{0,1,2,5,10\}$ after initializing from a short RLVR format-warmup checkpoint. We also train a prompt-conditioned dynamic-penalty model that samples $\lambda$ uniformly from the same set and an expected-penalty ablation at $\bar{\lambda}=3.6$. MATH numbers below are validation results from 2,538 problems with 16 rollouts per problem; checkpoints are selected by validation reward, with configuration and selected-step details in Appendices~\ref{sec:appendix_hyperparameters} and~\ref{sec:appendix_math_protocol}.

\subsection{Experimental Results}
\vspace{-1mm}

\begin{figure}[t]
    \centering
    \includegraphics[width=\columnwidth]{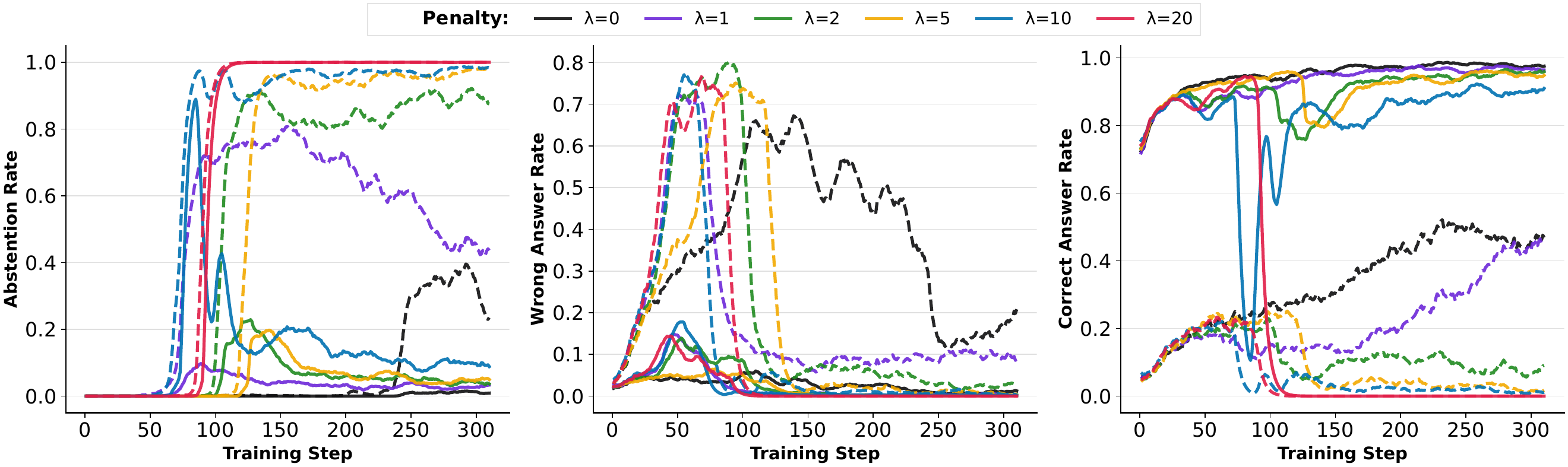}
    \caption{\textbf{Knights \& Knaves behavioral decomposition by difficulty.} Solid lines show easy logic puzzles (66\% of the Knights \& Knaves dataset), dashed lines show hard puzzles (33\%). \textbf{Left:} Models with $\lambda>0$ learn calibrated abstention: 5-15\% on easy versus 60-95\% on hard problems, while $\lambda=0$ abstains near 0\% regardless. The $\lambda=10$ transient spike to 97\% on easy problems (step 80) followed by recovery indicates behavioral recalibration. \textbf{Middle:} Wrong rates rapidly suppress to $<$2\% for all $\lambda\geq1$ while $\lambda=0$ maintains 10-20\% errors. \textbf{Right:} Correct rates reveal the coverage-safety tradeoff, with $\lambda=10$ achieving 90\% accuracy when it does answer.}
    \label{fig:behavioral_decomposition}
    \vspace{-1mm}
\end{figure}

On Knights \& Knaves, the penalty $\lambda$ determines optimal abstention thresholds, with models learning three distinct strategies to maximize expected reward (\Cref{fig:behavioral_decomposition}). With $\lambda=0$, models learn that always answering maximizes reward (0.82) despite maintaining a persistent wrong answer ratio of around 15\% with near-zero abstention. With $\lambda \in \{1,2,5\}$, models learn to trade coverage for safety: wrong answers collapse below 2\%, and models develop calibrated abstention (5-10\% on easy problems, 60-95\% on hard). With $\lambda \geq 10$, the severe error penalty teaches conservative abstention: $\lambda=10$ achieves $<$1\% wrong answers, while $\lambda=20$ collapses to universal abstention. This progression from aggressive answering through selective hesitation to conservative abstention demonstrates how RH controls the accuracy-trustworthiness trade-off.

\textbf{Training dynamics.} The most compelling evidence that models are optimizing the RH reward function rather than losing capability comes from the $\lambda=10$ model's ``transient crisis'': abstention spikes to 97\% on easy problems around step 80, before recovering to 10\% by step 300, while maintaining $>$95\% abstention on hard problems throughout (\Cref{fig:behavioral_decomposition}). This temporary over-caution followed by selective re-engagement shows that models are learning decision boundaries, not losing capabilities. All models with $\lambda>0$ demonstrate difficulty discrimination: they abstain proportionally more on hard problems (60-95\%) than easy ones (5-10\%), confirming that \textbf{models learn when abstention yields higher expected reward than attempting an uncertain answer} (full trajectories and qualitative traces are in Appendices~\ref{sec:appendix_training_dynamics} and~\ref{sec:appendix_qualitative}).

\subsection{MATH Generalization}
\vspace{-1mm}

\begin{figure}[t]
    \centering
    \includegraphics[width=0.88\textwidth]{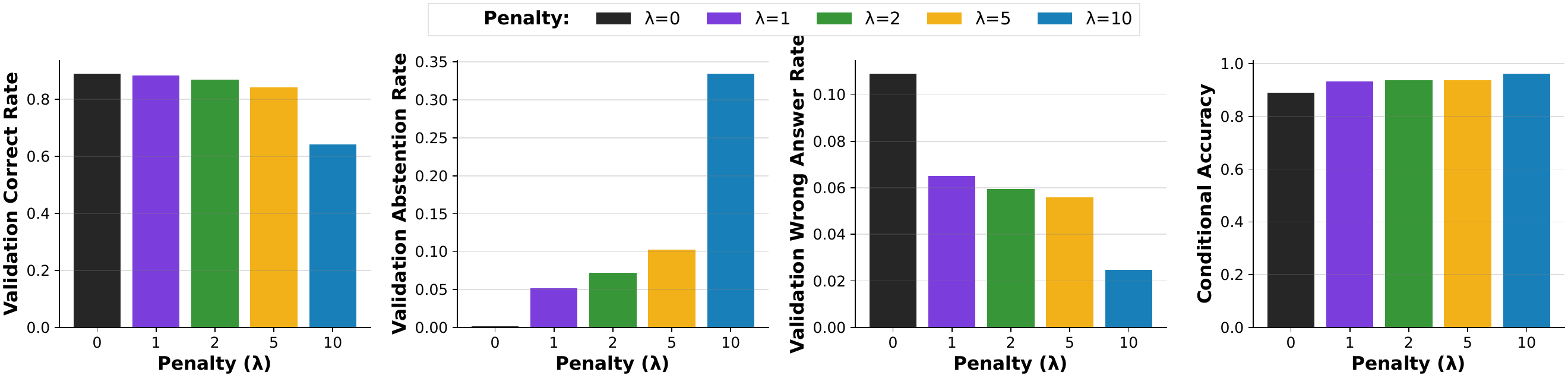}
    \caption{\textbf{MATH validation metrics.} Fixed-penalty RH specialists show the intended trade-off directly: increasing $\lambda$ lowers wrong-answer rate and raises conditional accuracy, while abstention absorbs coverage. Full outcomes, risk-frontier, difficulty, and format diagnostics are in Appendix~\ref{sec:appendix_math_extra_results}.}
    \label{fig:math_validation_metrics_main}
    \vspace{-2mm}
\end{figure}

The same qualitative trade-off appears on MATH, but the frontier is softer than in the controlled logic setting. As shown in \Cref{fig:math_validation_metrics_main}, increasing $\lambda$ from 0 to 10 cuts wrong answers from 11.4\% to 2.75\% and raises conditional accuracy from 88.6\% to 95.8\%, while abstention rises from near-zero to 34.39\%. Moderate penalties provide useful intermediate operating points: $\lambda=1$ retains 87.6\% correctness while reducing wrong answers to 6.25\%. This shows that RH is not merely exploiting puzzle-specific structure, although MATH does not exhibit the near-perfect separation seen on Knights \& Knaves.

As a direct confidence-threshold baseline, we also compare to an RLCR-style verbalized-confidence model \citep{damani2025rlcr}: RH matches it when errors are free and obtains higher utility at nonzero penalties, especially in the high-penalty regime. The matched outcome table, diagnostic plots, and decision-theoretic relationship between thresholded confidence and RH are in Appendix~\ref{sec:appendix_rlcr_comparison}.

\subsection{Risk Frontiers Across Penalties}
\vspace{-1mm}

Our experiments show that different penalties produce different accuracy-trustworthiness trade-offs, raising a natural question: does any single penalty produce a universally dominant model? We cross-evaluate each Knights \& Knaves model trained with $\lambda_{\text{train}}$ against all evaluation penalties, scoring using $\text{Reward} = \text{Correct Ratio} - \lambda_{\text{eval}} \cdot \text{Wrong Ratio}$ where $\lambda_{\text{eval}} \in \{0,1,2,5,10,20\}$. As seen in \Cref{fig:pareto_frontier}, the Knights \& Knaves results reveal a sharply ordered family: the baseline ($\lambda=0$) achieves the highest reward (0.837) when errors are free but plummets to -2.36 at $\lambda_{\text{eval}}=20$, while the $\lambda=10$ model shows the opposite pattern (0.559 at $\lambda_{\text{eval}}=20$). The near-diagonal clustering of optimal models confirms that each training penalty produces a useful specialist for its corresponding risk regime. Although the diagonal is not perfectly optimal\footnote{Which we hypothesize is due to randomness in training and imperfect optimization.}, no single model uniformly replaces the family: each has learned a distinct strategy for balancing accuracy and trustworthiness.

On MATH, the same structure is present but less exact: 3 of 5 evaluation-penalty columns are maximized by their diagonal fixed-penalty specialist, while neighboring penalties compete closely in the remaining columns (Appendix~\ref{sec:appendix_math_training}). We therefore use the stronger KK result as the mechanism study and the MATH result as evidence for an ordered risk frontier in a harder setting. This structure indicates a need for rethinking evaluation: rather than racing for a universal accuracy leaderboard, deployment should choose operating points according to the observable cost of wrong answers, abstentions, and verification.

\subsection{Dynamic Penalty Conditioning}
\vspace{-1mm}

A natural alternative to training one model per $\lambda$ is to train a single $\lambda$-conditioned policy. We test this by sampling $\lambda$ uniformly from $\{0,1,2,5,10\}$ during MATH training and including the sampled value in the prompt. The dynamic model is not fully penalty-blind: its IDK rate increases monotonically with prompted $\lambda$. However, the effect is weak, spanning only 8 percentage points, about 24\% of the 34-point abstention span achieved by separately trained specialists.

A fixed expected-penalty ablation at $\bar{\lambda}=\mathbb{E}[\Lambda]=3.6$ slightly outperforms the dynamic model, suggesting that current RLVR optimization treats sampled penalties partly as reward noise rather than learning a full risk-conditioned policy. This expected-penalty collapse is consistent with broader evidence that neural networks can initially track simple predictors before using their full nonlinear capacity \citep{hu2020surprising}. Thus, fixed specialists remain the reliable method in our experiments, while learning a single fully $\lambda$-conditioned RH policy remains an open optimization problem. Appendix~\ref{sec:appendix_math_dynamic} gives the degeneracy derivation, dynamic-versus-expected-penalty curves, and prompt-conditioned validation diagnostics.

%% file: sections/inference_compute.tex
\section{Scaling Inference Compute Through Exploiting Learned Hesitation}\label{sec:rh_inference}
\vspace{-2mm}

A natural reaction to trained abstentions might view them as limitations where coverage is lost in exchange for safety. This perspective, however, misses a practical opportunity:

\begin{center}
\parbox{0.95\columnwidth}{\centering\itshape
\textbf{``I don't know'' is not a terminal signal but an informationally rich indicator that enables entirely new inference paradigms.}}
\end{center}

\noindent Consider how human experts handle uncertainty: when a doctor says ``I need to consult a specialist,'' they're intelligently routing uncertainty through expertise. Similarly, when our models say ``I don't know,'' they provide actionable information about their learned boundaries. This section shows how to transform abstentions into coordination signals that enable adaptive computation.

\begin{figure*}[t]
    \centering
    \includegraphics[width=0.89\textwidth]{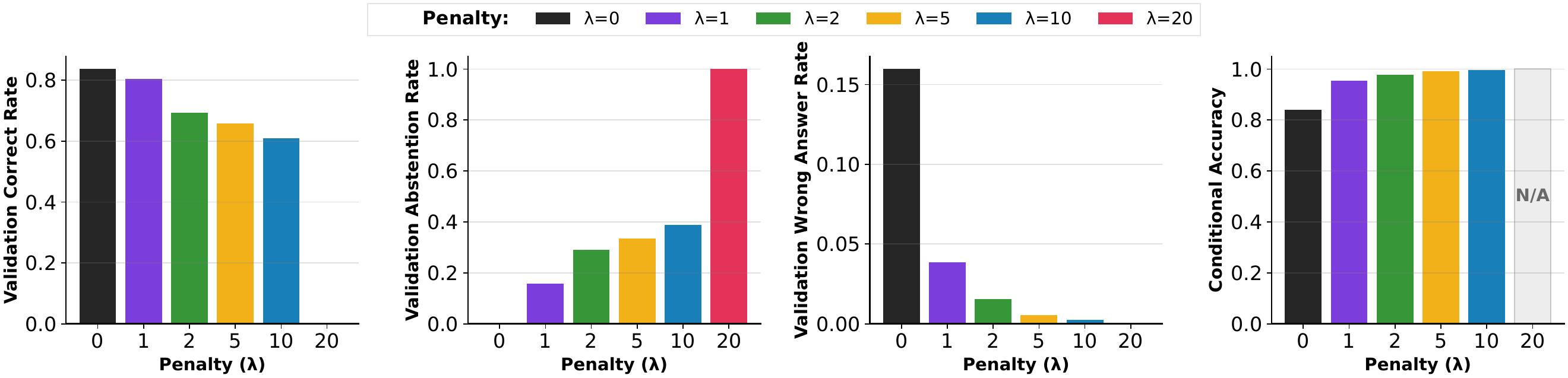}
    \caption{\textbf{Knights \& Knaves validation performance reveals learned selectivity.} As training penalty $\lambda$ increases, models trade coverage for safety on held-out logic puzzles: correct rates decrease while abstention rates rise, but wrong rates collapse dramatically. The most informative insight is conditional accuracy (rightmost panel) jumping from 84\% to $>$99\%, proving models learn to abstain precisely on problems where they would likely make mistakes.}
    \label{fig:validation_metrics}
    \vspace{-3mm}
\end{figure*}

Figure~\ref{fig:validation_metrics} shows this effect on Knights \& Knaves and reveals why these abstentions are exploitable: models trained with penalties develop genuine discrimination about their competence. The Knights \& Knaves baseline model ($\lambda=0$) achieves 84\% accuracy across all problems. But models trained with moderate penalties ($\lambda=1,2$), while abstaining on 15-30\% of problems, achieve 95-99\% accuracy \emph{on the problems they do choose to answer}. This jump in conditional accuracy shows that abstentions are targeted at error-prone cases rather than being random refusals. \textbf{The ordered relationship between penalty and error rate is what makes cascading possible: we can trust high-penalty models when they answer.} The same pattern appears on MATH, where conditional accuracy rises from 88.6\% at $\lambda=0$ to 95.8\% at $\lambda=10$. Each ``I don't know'' effectively says: \emph{My training taught me that for this problem, the expected penalty from potentially being wrong exceeds the reward from potentially being right.} This learned selectivity becomes exploitable through two properties: specialization across the trained model family and nondeterminism during language-model inference.

First, different penalty values create models with complementary behavioral regimes. What a $\lambda=10$ model abstains from, a $\lambda=1$ model might confidently attempt, mirroring medical triage from conservative gatekeepers to risk-tolerant specialists. This behavioral diversity enables routing problems architecturally through models with different risk tolerances. \emph{This is what we call \textbf{cascading}.}

Second, LLM inference is inherently nondeterministic. Sampling strategies and autoregressive dynamics mean small variations early in generation amplify through reasoning chains. For problems where models abstain, these variations might occasionally produce different outcomes, allowing successful solution paths to emerge on subsequent attempts. This suggests we can exploit nondeterminism temporally by re-querying the same model. \emph{We call this \textbf{self-cascading}}.

\subsection{Cascading: Routing for Efficiency}\label{sec:rh_cascading}
\vspace{-1mm}

Cascading leverages behavioral diversity across models to create an efficient inference pipeline. The key insight is that \textbf{conditional accuracy increases with penalty}: $\lambda=10$ achieves $>$99\%, $\lambda=2$ achieves 97\%, $\lambda=1$ achieves 95\%, and $\lambda=0$ achieves 84\% on Knights \& Knaves. Therefore, \textbf{we query models in descending penalty order to always get the most reliable available answer}. The risk frontier from Figure~\ref{fig:pareto_frontier} showed that each penalty creates different comparative advantages. Cascading transforms this apparent limitation into architectural strength by arranging models into a risk-tolerance hierarchy. When the $\lambda=10$ model abstains, it doesn't fail but delegates: \emph{``This problem requires someone more knowledgeable or willing to accept higher error risk.''}

\noindent We use a simple early-exit rule: query specialists in descending penalty order, $\lambda=10\to5\to2\to1\to0$, and return the first clean non-IDK answer; if every specialist abstains, return IDK. This sequential structure engages only the necessary models, unlike ensemble methods that query all models regardless of difficulty. Formal pseudocode and MATH accounting details are in Appendix~\ref{sec:appendix_math_inference_diagnostics}.

\begin{figure}[t]
    \centering
    \includegraphics[width=\columnwidth]{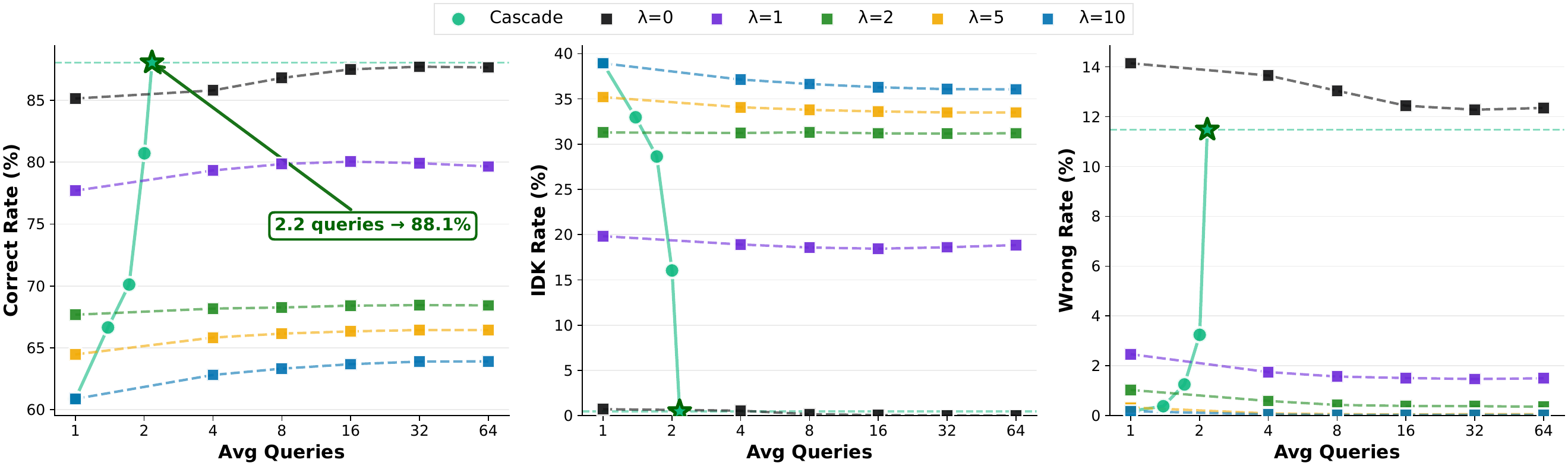}
    \caption{\textbf{Knights \& Knaves cascade is query-efficient through behavioral complementarity.} Dashed lines represent majority voting applied on models trained with different penalties, solid lines represent cascading. \textbf{Left:} Cascade (\textcolor{teal}{$\bigstar$}) achieves 88.1\% accuracy with only 2.2 average queries, outperforming individual Knights \& Knaves specialists. \textbf{Middle:} IDK rates collapse from 33\% to $<$1\% through cascading. \textbf{Right:} Wrong rates remain controlled at 11.5\%, competitive with the baseline while having higher coverage.}
    \label{fig:cascade_performance}
    \vspace{-3mm}
\end{figure}

\begin{figure}[t]
    \centering
    \includegraphics[width=\columnwidth]{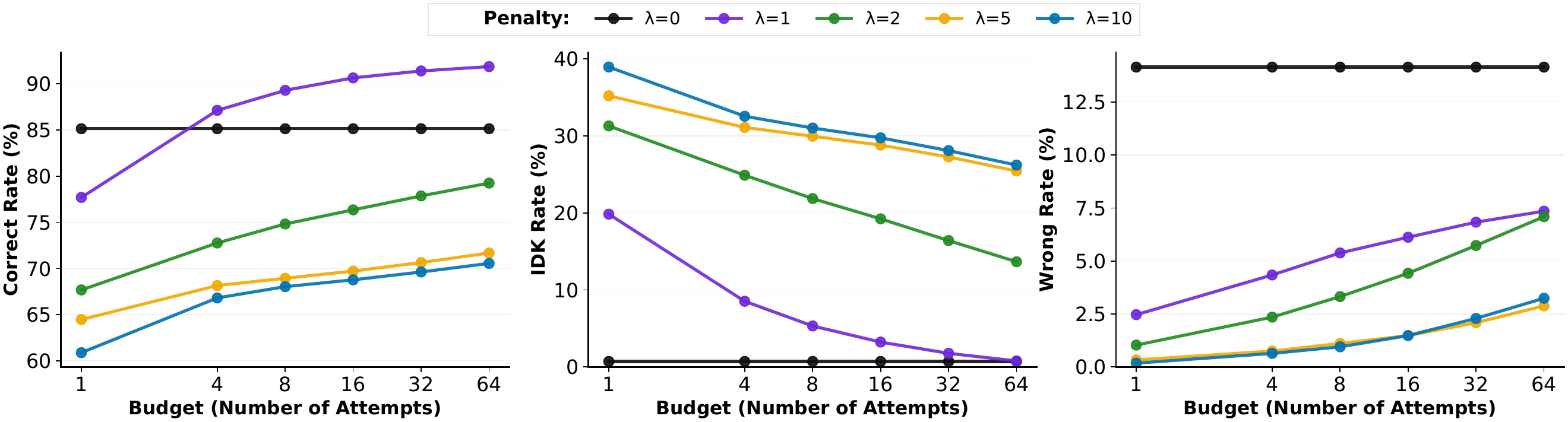}
    \caption{\textbf{Knights \& Knaves self-cascading converts abstentions to answers through nondeterminism.} \textbf{Left:} Correct rate increases with budget for models with $\lambda \geq 1$, with $\lambda=1$ showing steepest gains (77.5\%$\rightarrow$92.5\%). The $\lambda=0$ baseline remains flat as it never abstains. \textbf{Middle:} IDK rates decay with budget. \textbf{Right:} Wrong rates increase as abstentions convert to answers while mistakes remain bounded.}
    \label{fig:self_cascading}
    \vspace{-3mm}
\end{figure}

The power of cascading becomes clear in Figure~\ref{fig:cascade_performance} where it is compared with the well-known baseline of majority voting. \textbf{A five-tier cascade ($\lambda \in \{10,5,2,1,0\}$) achieves 88.1\% accuracy with only 2.2 average queries on Knights \& Knaves}. This reduces model calls while still returning at most one terminal non-IDK answer per problem. When verification is costly (human mathematicians checking proofs) or impossible to aggregate (distinct code solutions), this efficiency becomes essential. Conservative tiers, which achieve $>$99\% conditional accuracy (Figure~\ref{fig:validation_metrics}), serve as highly reliable filters for straightforward cases. Their abstentions become precise routing signals to models trained for greater risk tolerance. This transforms computational cost into an interpretable confidence indicator: tier 1 resolutions indicate straightforward problems, while tier 5 traversals signal genuine challenges requiring maximum risk tolerance.

\textbf{MATH cascade.} On MATH, the same cascade ($\lambda=10\to5\to2\to1\to0$) reaches 89.97\% correct with 1.58 average queries, competitive with 16-sample majority voting at $\lambda=0$ (91.05\% correct) while using about one tenth as many queries. Stopping before the final $\lambda=0$ model gives a safer operating point: the depth-4 cascade ($10\to5\to2\to1$) reaches 88.54\% correct, 4.60\% IDK, and 6.86\% wrong with 1.53 average queries, while the full cascade restores coverage but raises wrong answers to 9.96\%. Thus the cascade does not uniformly dominate majority voting in raw accuracy; its advantage is query-efficient control over the coverage-error trade-off. Additional MATH inference baselines and diagnostics are in Appendices~\ref{sec:appendix_math_inference}--\ref{sec:appendix_math_inference_diagnostics}.

This shows that Reinforced Hesitation does not create a hierarchy of quality, but a spectrum of specialization. The $\lambda=10$ model is not inferior to $\lambda=0$; it serves a different role as a reliable gatekeeper versus a risk-tolerant problem solver. Where self-cascading exploits randomness within a single model and voting seeks agreement across identical samples, cascading combines models with different strengths into a collaborative pipeline.

\subsection{Self-Cascading: Scaling Through Multiple Attempts}
\vspace{-1mm}

When a model says ``I don't know,'' asking again can still help because stochastic decoding can explore a different reasoning path. Self-cascading treats abstention as a temporary early-exit signal rather than permanent failure: repeated samples from the same selective model may eventually produce a clean answer.

Self-cascading uses the same early-exit rule with a homogeneous model sequence $\{M_1 = M_2 = \cdots = M_k = \pi\}$, typically using a moderate-penalty specialist. Unlike majority voting, it stops at the first non-abstention, reducing expected queries and verification cost.

Figure~\ref{fig:self_cascading} shows that this converts some abstentions into answers on Knights \& Knaves: the $\lambda=1$ model improves from 77.5\% to 92.5\% accuracy as budget grows from $B=1$ to $B=64$, while abstention falls from 19.5\% to 0.5\%. Wrong answers rise but remain bounded, consistent with the high conditional accuracy of moderate-penalty specialists.
 
\textbf{MATH self-cascade and majority voting.} The same pattern transfers with smaller gains: at $B=16$, the MATH $\lambda=1$ specialist reaches 90.30\% correct with 1.43 average queries, compared with 91.05\% for 16-sample majority vote at $\lambda=0$. RH-based inference is therefore not a raw-accuracy replacement for majority voting in every regime; it provides early-exit mechanisms for trading query cost, coverage, and wrong-answer rate. Full MATH self-cascade and majority-vote plots are in Appendix~\ref{sec:appendix_math_inference}.

\textbf{Why cascading can help.} The threshold $\lambda/(1+\lambda)$ describes a Bayes decision rule for a calibrated agent, but trained fixed-$\lambda$ specialists are not merely thresholds over the same policy. They learn different answering policies whose answered subsets have different conditional accuracies. Cascading uses this learned diversity: high-$\lambda$ models answer first on the subset where they are most reliable, and lower-$\lambda$ models recover coverage only after conservative models abstain.

%% file: sections/conclusion.tex
\section{Conclusion, Limitations and Future Work}
\vspace{-2mm}
We showed that frontier LMs remain effectively penalty-blind under prompt-level error penalties, then introduced \emph{Reinforced Hesitation} (RH), a minimal RLVR modification with ternary rewards (+1 correct, 0 abstain, $-\lambda$ wrong). On Knights \& Knaves, MATH, and medical QA (Appendix~\ref{sec:appendix_medical_suite}), varying $\lambda$ yields risk-specialized policies, and abstention becomes a useful coordination signal for cascading and self-cascading inference.

\textbf{Limitations and future work.} Choosing $\lambda$ is empirical and should use development-set wrong-answer rate, abstention rate, conditional accuracy, and deployment reward. Fixed-$\lambda$ specialists currently outperform a single dynamic-$\lambda$ model, while cascades trade extra queries for risk control. Future work should extend RH beyond exact verification, penalize unnecessary abstention when appropriate, and improve single-model risk conditioning.

%% file: sections/acknowledgements.tex
\section{Acknowledgments}
We would like to thank Avrim Blum for helpful discussions and Danica J.\ Sutherland for her support. This work was enabled in part by support provided by the Natural Sciences and Engineering Research Council of Canada, the Canada CIFAR AI Chairs program, Advanced Research Computing at the University of British Columbia, Calcul Québec, the BC DRI Group, and the Digital Research Alliance of Canada.

%% file: appendix/related_work.tex
\section{Related Work}
\label{sec:related_work}

\textbf{Abstention and uncertainty quantification.} Selective prediction with reject options studies the accuracy--coverage tradeoff \citep{chow1970optimum, bartlett2008classification, geifman2017selective}, while reliable learning and don't-know prediction study learners that may say ``I don't know'' instead of making a mistake, and the tradeoff between mistakes and don't-know predictions \citep{RS88b,sayedi2010trading}. In question answering, reading comprehension benchmarks introduced unanswerable questions that require systems to abstain \citep{rajpurkar-etal-2018-know, kwiatkowski-etal-2019-natural}. Recent evaluations find that LLMs, including strong proprietary models, often answer when they should abstain \citep{kirichenko2025abstentionbenchreasoningllmsfail,madhusudhan-etal-2025-llms,Saadat2024WhenNotToAnswer,sun-etal-2024-benchmarking} and over-answer when errors are costly \citep{wu2025answer}; see \citet{wen-etal-2025-know} for a survey. Models can partly assess whether their own answers are correct \citep{Kadavath2022MostlyKnow} and can express calibrated confidence when prompted or fine-tuned \citep{Tian2023JustAskForCalibration,Lin2022UncertaintyInWords}, although prompted verbal confidence is often overconfident \citep{xiong2024can}. Other work reduces overconfidence during training, through knowledge-transfer fine-tuning \citep{Yang2024CanWeTrust} or calibrated reward models in RLHF \citep{Leng2025TamingOverconfidence}, or adds inference-time checks such as uncertainty-based abstention \citep{Tomani2024UncertaintyAbstention} and sampling-based consistency checks \citep{manakul-etal-2023-selfcheckgpt}. These inference-time signals help but remain imperfect \citep{Tomani2024UncertaintyAbstention}, and reasoning models often commit to a definitive answer even when their reasoning expresses uncertainty \citep{kirichenko2025abstentionbenchreasoningllmsfail}.

\noindent \textbf{Training paradigms and reward structures.} RLHF optimizes scalar rewards from reward models trained on human preference comparisons \citep{ouyang2022instructgpt}, typically with PPO \citep{schulman2017ppo}, while RLVR replaces the learned reward model with a verifier and a binary reward, 1 for a correct final answer and 0 otherwise \citep{DeepSeekAI2025R1,anonymous2025drGRPO,damani2025rlcr}, often optimized with GRPO \citep{Shao2024DeepSeekMath}; Tulu~3 uses the same binary signal with a larger constant in place of 1 \citep{lambert2025tulu3}. Frontier reasoning models are also post-trained with reinforcement learning \citep{jaech2024o1,google2025gemini25Pro}, although their reward functions are not documented. This binary structure rewards correct final answers even when the reasoning trace is unfaithful \citep{chen2025reasoningfaithfulness}. Theoretical analyses of calibrated language models show why confident wrong answers can arise under binary scoring and incomplete information \citep{kalai2024calibrated,Kalai2025WhyHallucinate}. RH changes the reward structure by making wrong answers explicitly costly and abstention a first-class action.

\noindent \textbf{Training abstention and verbalized confidence.} Several recent and concurrent works modify training so uncertainty becomes part of the model behavior. TruthRL \citep{wei2025truthrl} trains abstention with a fixed ternary reward, closely related to the $\lambda=1$ member of the RH family in verifiable-answer settings. BCRL \citep{wu2025bcrl} and Rewarding Doubt \citep{baniharouni2025rewardingdoubt} use related confidence- or calibration-oriented objectives. RLCR \citep{damani2025rlcr} trains models to emit answers with calibrated verbalized confidence using a Brier-style reward; we convert this confidence into decisions by thresholding it at inference time. We compare directly to an RLCR-style baseline in Appendix~\ref{sec:appendix_rlcr_comparison}. RH differs by training a family of risk-specialized answer/abstain policies indexed by $\lambda$, rather than learning a single confidence-producing policy and applying post-hoc thresholds. In concurrent work, \citet{jha2026rewarding} train with a ternary reward $(+1, r_{\text{abs}}, -1)$ and sweep the abstention reward $r_{\text{abs}}$ on MedMCQA and MATH. This is a reparameterization of the same decision problem: answering is optimal when the probability of a correct answer exceeds $(1+r_{\text{abs}})/2$, which corresponds to $\lambda = (1+r_{\text{abs}})/(1-r_{\text{abs}})$. Their swept range $r_{\text{abs}} \in [-0.95, 0.3]$ therefore corresponds to $\lambda \in [0.03, 1.86]$, between our $\lambda=0$ and $\lambda=2$ specialists, and $r_{\text{abs}}=0$ corresponds exactly to $\lambda=1$. They study single models for hallucination mitigation, while our sweep extends to $\lambda=20$ (threshold 0.95) and uses the resulting specialists for cascading. Abstain-R1 \citep{zhai2026abstain} rewards abstaining on questions that cannot be resolved from the given information, together with a verifiable clarification of what is missing. This is complementary to RH, which assumes that every question has a verifiable answer and asks whether the model can find it at an acceptable error cost. Table~\ref{tab:related_work_comparison} summarizes the closest training-time methods.

\begin{table}[!htbp]
\caption{\textbf{Closest training-time abstention and confidence methods.} RH is closest to fixed ternary abstention training, but the main empirical baseline in this paper is RLCR-style verbalized confidence \citep{damani2025rlcr}, which we threshold at the Bayes cutoff $\lambda/(1+\lambda)$ to obtain abstention.}
\label{tab:related_work_comparison}
\centering
\small
\setlength{\tabcolsep}{3.5pt}
\begin{tabular}{@{}p{0.16\textwidth}p{0.22\textwidth}p{0.25\textwidth}p{0.27\textwidth}@{}}
\toprule
\textbf{Method} & \textbf{Uncertainty signal} & \textbf{Training objective} & \textbf{Relation to RH} \\
\midrule
RH (ours) & Explicit answer/IDK action & Ternary utility $(+1,0,-\lambda)$ & Learns fixed-risk specialists and uses abstention as an inference-routing signal. \\
TruthRL \citep{wei2025truthrl} & Explicit abstention & Fixed ternary abstention reward & Closest to the $\lambda=1$ member of RH in verifiable-answer settings. \\
RLCR \citep{damani2025rlcr} & Verbalized confidence & Correctness plus Brier-style calibration & Direct confidence-threshold baseline; we threshold at $\lambda/(1+\lambda)$. \\
BCRL \citep{wu2025bcrl} & Verbalized confidence, thresholded post hoc at a user-specified risk level $t$ & Proper scoring rules; also a prompted-$t$ ternary reward $(+1, 0, -t/(1-t))$ & Adjacent single-model approach to user-specified risk; its prompted-$t$ variant did not adapt to $t$. \\
Rewarding Doubt \citep{baniharouni2025rewardingdoubt} & Verbalized integer confidence (0 to 10) & Logarithmic scoring rule on confidence, with answers held fixed & Complementary direction focused on calibrated confidence expression. \\
\bottomrule
\end{tabular}
\end{table}

\noindent \textbf{Direct abstention versus confidence thresholding.} Confidence-threshold methods and RH address adjacent selective-decision problems. If a model has already found a candidate answer and can assess or verify it well, thresholding calibrated confidence is natural: answer when confidence exceeds $\lambda/(1+\lambda)$ and abstain otherwise. In compute-bounded reasoning, however, the decision often arises earlier: whether continued search, verification, or guessing is worth the cost under a deployment penalty. A scalar confidence is then not only a property of the problem, but also of the solver, compute budget, and verification procedure. RH trains the deployment decision directly under the target cost, in the spirit of Vapnik's principle that one should avoid solving a more general problem as an intermediate step when the target decision is narrower \citep{vapnik1998statisticallearning}. This makes confidence methods complementary baselines rather than straw men: they communicate uncertainty, while RH makes abstention a trained action and a routing signal.

\noindent \textbf{Inference-time computation and model cascading.} Verifier-based methods rerank sampled candidates \citep{Cobbe2021TrainingVerifiers,Lightman2024LetsVerify} or iteratively verify and refine a solution until it passes repeated checks \citep{Huang2025imo}, while self-consistency methods aggregate sampled answers by majority voting \citep{Lewkowycz2022Minerva,Shao2024DeepSeekMath,Chen2023ProgramOfThoughts}. Sampling-based detectors instead use agreement among samples, including semantic clustering, or with a second verifier model to flag likely hallucinations \citep{manakul-etal-2023-selfcheckgpt,farquhar2024detecting,xue2025verify}. Model cascading routes queries based on confidence \citep{Khalili2022babybear}, cascade-aware training \citep{Wang2024cascadeaware}, calibrated deferral \citep{Rabanser2025gatekeeper}, ensemble agreement \citep{Kolawole2025abc}, or privacy constraints \citep{Zhang2024p3defer}. RH instead builds routing behavior into a set of learned specialists: high-$\lambda$ models answer a high-conditional-accuracy subset, and abstentions route the remaining queries to lower-penalty specialists. This connects to the selective-decision view of \citet{Kalai2025WhyHallucinate}, while extending it beyond thresholding a single calibrated policy: different RH penalties train different policies whose answered subsets have different conditional accuracies.

%% file: appendix/appendix.tex
\section{Experimental Setup and Training Implementation}
\label{sec:appendix_training_implementation}

\subsection{Training Suites and Shared Configuration}
\label{sec:appendix_hyperparameters}

We run two RL training suites, summarized in Table~\ref{tab:appendix_training_suites}. Within each suite, all non-reward hyperparameters are matched across the primary fixed-penalty RH runs so that behavioral differences are attributable to the reward structure.

\begin{table}[H]
\caption{\textbf{Training suites.} The two suites share the same answer/abstain decision structure while testing RH on different reasoning domains and model scales.}
\label{tab:appendix_training_suites}
\centering
\small
\setlength{\tabcolsep}{4pt}
\begin{tabular}{@{}p{0.19\textwidth}p{0.30\textwidth}p{0.43\textwidth}@{}}
\toprule
\textbf{Suite} & \textbf{Model and data} & \textbf{Reward variants} \\
\midrule
Knights \& Knaves & Qwen/Qwen3-1.7B \citep{qwenTeam2025qwen3}; 80,000 training puzzles and 10,000 test puzzles. & Fixed RH penalties $\lambda \in \{0,1,2,5,10,20\}$. \\
MATH & Qwen3.5-4B; MATH Levels 4--5 with 3,994 training problems and 2,538 validation problems. & Fixed RH penalties $\lambda \in \{0,1,2,5,10\}$, prompt-conditioned dynamic penalty over the same set, expected-penalty ablation at $\lambda=3.6$, and an RLCR baseline \citep{damani2025rlcr}. \\
\bottomrule
\end{tabular}
\end{table}

All MATH RH runs initialize from step 20 of a short RLVR format-warmup run. The warmup reward is +1 for correct answers, 0 for wrong answers, and -1 for format errors; its role is only to teach the required response format before penalty training.

\begin{table}[H]
\caption{\textbf{Core training configuration.} Shared settings are listed first, followed by suite-specific optimization and evaluation details.}
\label{tab:appendix_training_config}
\centering
\small
\setlength{\tabcolsep}{4pt}
\begin{tabular}{@{}p{0.24\textwidth}p{0.68\textwidth}@{}}
\toprule
\textbf{Setting} & \textbf{Value} \\
\midrule
\multicolumn{2}{@{}l}{\textbf{Shared settings}} \\
Training pipeline & verl \citep{sheng2024hybridflow} with SGLang rollouts \citep{zheng2024sglangefficientexecutionstructured}; MATH uses the same verl/DrGRPO stack. \\
Rollout sampling & Temperature 1.0 with nucleus sampling (top-p=1.0), 8 rollouts per prompt. \\
Validation decoding & Deterministic decoding, temperature 0, $n=1$. \\
Response limit & 4096 tokens. \\
\midrule
\multicolumn{2}{@{}l}{\textbf{Knights \& Knaves}} \\
Optimizer & AdamW, $\beta_1=0.9$, $\beta_2=0.999$, $\epsilon=10^{-8}$, weight decay $0$. \\
Learning rate and batch & Learning rate $2\times10^{-6}$; 256 prompts per iteration. \\
Training horizon & 1 epoch, 312 steps. \\
Prompt/stabilization & Prompt cap 650 tokens; gradient clipping at $\ell_2$ norm 1.0; KL coefficient disabled. \\
\midrule
\multicolumn{2}{@{}l}{\textbf{MATH L4--L5}} \\
Optimization & Actor learning rate $1\times10^{-6}$; 256 prompts per iteration; PPO mini-batch size 256. \\
Training horizon & Reported through step 150; checkpoints and validation every 10 steps. \\
Prompt/model selection & Prompt cap 2048 tokens; validation snapshot plots select the checkpoint with highest validation RH reward among steps $\leq150$. \\
Infrastructure & A single node with 4 or 8 GPUs per run (several runs moved to a different node partway through training); validation batch size 1024. \\
\bottomrule
\end{tabular}
\end{table}

\subsection{Knights \& Knaves Dataset Construction}
\label{sec:appendix_dataset}

Our Knights \& Knaves dataset contains logic puzzles where each character is either a knight (who always tells the truth) or a knave (who always lies). The dataset includes 80,000 training and 10,000 test samples, evenly distributed across 5, 6, and 7-person puzzles (approximately 33.3\% each).

\noindent \textbf{Difficulty Split.} Each puzzle is labeled as easy or hard based on the complexity of logical statements (not visible to the model, only used for evaluation purposes). The difficulty is determined by the number of nested conditionals and the presence of biconditionals. This 2:1 ratio of easy to hard problems ensures models encounter sufficient challenging cases while maintaining a stable training signal from simpler problems. Below we show representative examples of easy and hard puzzles to illustrate the complexity difference.

\vspace{2mm}
\begin{examplebox}[Example Easy Puzzle (5 people)]
\small
\emph{A very special island is inhabited only by knights and knaves. Knights always tell the truth, and knaves always lie. You meet 5 inhabitants: Quillan, Thorsten, Victoria, Eurydice, and Henry.}
\begin{itemize}[leftmargin=*,itemsep=1mm,parsep=0mm]
\item Quillan says: ``Henry is a knave''
\item Thorsten says: ``Eurydice is a knight or Quillan is a knave''
\item Victoria says: ``Eurydice is not a knave''
\item Eurydice says: ``Victoria is a knight and Victoria is a knave''
\item Henry says: ``Henry is a knight or Quillan is a knight''
\end{itemize}

\textbf{Solution:} Quillan is a knave, Thorsten is a knight, Victoria is a knave, Eurydice is a knave, Henry is a knight.
\end{examplebox}

\vspace{2mm}
\begin{examplebox}[Example Hard Puzzle (5 people)]
\small
\emph{A very special island is inhabited only by knights and knaves. Knights always tell the truth, and knaves always lie. You meet 5 inhabitants: Quillan, Thorsten, Victoria, Eurydice, and Henry.}
\begin{itemize}[leftmargin=*,itemsep=1mm,parsep=0mm]
\item Quillan says: ``if Henry is a knave if and only if Victoria is a knave then Eurydice is a knight if and only if Henry is a knight''
\item Thorsten says: ``if Henry is a knight if and only if Thorsten is a knight then Victoria is a knight or Eurydice is a knave''
\item Victoria says: ``if if Quillan is a knight then Eurydice is a knave then Thorsten is a knave''
\item Eurydice says: ``if Eurydice is a knight then Thorsten is a knight if and only if Victoria is a knave or Eurydice is a knight or Quillan is a knight''
\item Henry says: ``if Quillan is a knave or Thorsten is a knight then Thorsten is a knave if and only if Victoria is a knight''
\end{itemize}

\textbf{Solution:} Quillan is a knight, Thorsten is a knight, Victoria is a knight, Eurydice is a knight, Henry is a knave.
\end{examplebox}

\subsection{MATH Dataset, Protocol, and Checkpoint Selection}
\label{sec:appendix_math_protocol}

The MATH benchmark \citep{Hendrycks2021MATH} consists of competition-style mathematics problems labeled by difficulty. For this suite, we train and evaluate on the two hardest difficulty levels, yielding 3,994 training problems and 2,538 validation problems. The base model is Qwen3.5-4B. All fixed-penalty RH runs initialize from step 20 of a short RLVR format-warmup run whose purpose is to teach boxed-answer formatting before penalty training begins.

Each fixed-penalty specialist is trained with a single value of $\lambda \in \{0,1,2,5,10\}$ using the same answer/abstain reward as the main text: +1 for correct answers, 0 for ``I don't know,'' $-\lambda$ for wrong answers, and a format penalty for malformed outputs. Validation rollouts use 16 samples per problem on the full 2,538-problem validation set. Checkpoints are selected by validation RH reward among logical training steps $\leq150$: $\lambda=0$ at step 110, $\lambda=1$ at step 140, $\lambda=2$ at step 140, $\lambda=5$ at step 150, and $\lambda=10$ at step 110. The dynamic-penalty model and expected-penalty ablation are both selected at step 150; the RLCR baseline \citep{damani2025rlcr} is selected at step 120. We report these as validation-selected validation results; a fully held-out test split would be the cleaner protocol for final deployment estimates.

\subsection{Training Prompts}
\label{sec:appendix_training_prompts}

All models receive prompts that explicitly permit abstention and require parseable final answers. The model is presented with the problem after the system prompt or reward prefix.

\vspace{2mm}
\begin{promptbox}[System Prompt]
\small
You are a helpful assistant. The assistant first thinks about the reasoning process in the mind and then provides the user with the answer. The reasoning process and answer are enclosed within \texttt{<think> </think>} and \texttt{<answer> </answer>} tags, respectively, i.e., \texttt{<think>} reasoning process here \texttt{</think><answer>} answer here \texttt{</answer>}. After thinking, if you reach a clear and confident conclusion, state the identity of each character within \texttt{<answer> </answer>} tags. i.e., \texttt{<answer> (1) Mark is a knight} \texttt{(2) ... </answer>}. Critically, if you are not confident in the solution, if the problem seems too complex/contradictory, or if you cannot determine a clear answer within the 4000-token limit, you must output \texttt{<answer>I don't know</answer>}.
\end{promptbox}

\vspace{2mm}
\begin{promptbox}[MATH Reward Prefix]
\small
Correct: +1 \textbar{} Wrong: $-\lambda$ \textbar{} ``I don't know'': 0

Decide: solve or abstain. Put your final answer in \texttt{\textbackslash boxed\{\}}.
\end{promptbox}

\section{Penalty-Blind Prompting Study Details}
\label{sec:appendix_frontier_eval}

\subsection{Evaluation Prompts}
\label{sec:appendix_prompts}

We evaluate 11 frontier models across three datasets with systematically modified prompts that include explicit abstention options and varying penalty structures. For each dataset, we maintain consistent prompting across models to ensure comparability.

\noindent \textbf{GSM8K Modified Prompts.} Building on the standard chain-of-thought prompt, we add explicit abstention instructions and reward structure for each penalty condition:

\vspace{2mm}
\begin{promptbox}[GSM8K Baseline (No Penalty)]
\small
Q: \{question\} When you reach an answer, output it in a block of three number signs, for example if answer is 0, then `\#\#\# 0 \#\#\#'. If you don't know the answer, you can simply output `\#\#\# I don't know \#\#\#'.\\
A: Let's think step by step.
\end{promptbox}

\vspace{2mm}
\begin{promptbox}[GSM8K Penalty $\lambda{=}100$ (representative example)]
\small
Q: \{question\} When you reach an answer, output it in a block of three number signs, for example if answer is 0, then `\#\#\# 0 \#\#\#'. If you don't know the answer, you can simply output `\#\#\# I don't know \#\#\#'. Correct answer gets \textbf{+1 point}, saying I don't know gets \textbf{0 points} and a wrong answer gets \textbf{-100 points}. You should provide an answer that maximizes the expected reward.\\
A: Let's think step by step.
\end{promptbox}

\noindent \textbf{MedQA/GPQA Modified Prompts.} We add option (E) ``I don't know'' to the standard 4-option multiple choice format:

\vspace{2mm}
\begin{promptbox}[MedQA Baseline (No Penalty)]
\small
Question: \{question\}\\
Choices:\\
(A) \{choice1\}\\
(B) \{choice2\}\\
(C) \{choice3\}\\
(D) \{choice4\}\\
(E) I don't know\\
Let's think step by step:
\end{promptbox}

\vspace{2mm}
\begin{promptbox}[MedQA/GPQA With Penalty (example $\lambda{=}25$)]
\small
Question: \{question\}\\
Choices:\\
(A) \{choice1\}\\
(B) \{choice2\}\\
(C) \{choice3\}\\
(D) \{choice4\}\\
(E) I don't know\\
Correct answer gets \textbf{+1 point}, saying I don't know gets \textbf{0 points} and a wrong answer gets \textbf{-25 points}. You should provide an answer that maximizes the expected reward.\\
Let's think step by step:
\end{promptbox}

\subsection{MedQA, GPQA, and All-Dataset Results}
\label{sec:appendix_extended_results}

\begin{figure}[H]
    \centering
    \includegraphics[width=\textwidth]{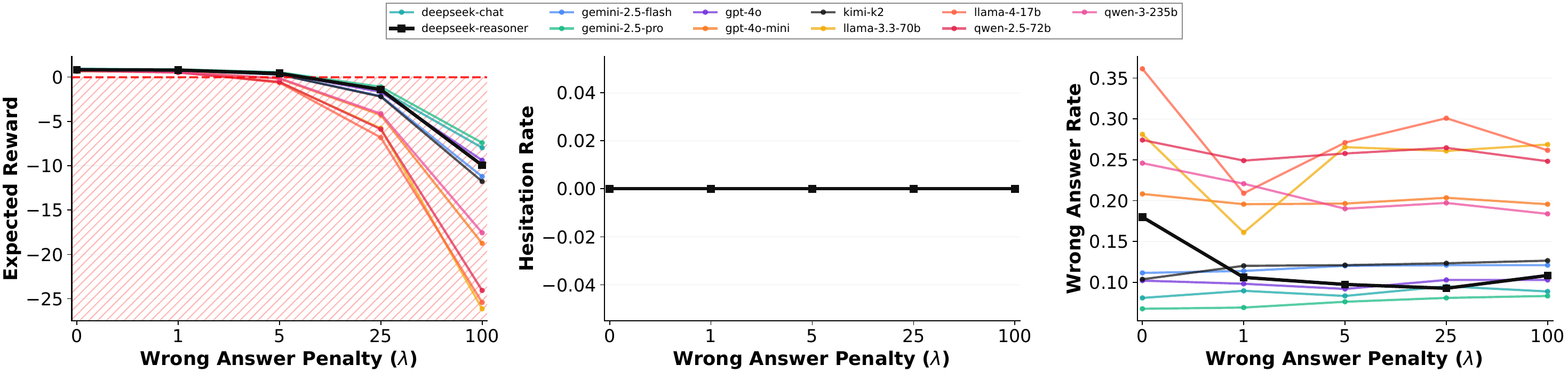}
    \caption{\textbf{MedQA results parallel to GSM8K (Figure~\ref{fig:frontier_eval_gsm8k}).} \textbf{Left:} Expected reward shows all models fall below zero for $\lambda \geq 5$. \textbf{Middle:} The hesitation curve is flat across all models and penalty conditions. \textbf{Right:} Wrong answer rates are largely unchanged (6-36\%) under prompt-level penalties.}
    \label{fig:medqa_results}
\end{figure}

\noindent \textbf{MedQA: Stable Answering Under Penalty Prompts.} Figure~\ref{fig:medqa_results} shows a consistent pattern: across 11 models, 5 penalty conditions, and 1,273 medical questions, we observe \textbf{exactly zero instances of abstention}. This provides a useful contrast case for our method: prompt-level penalties alone did not change answer/abstention behavior in this medical QA setting.

\begin{figure}[H]
    \centering
    \includegraphics[width=\textwidth]{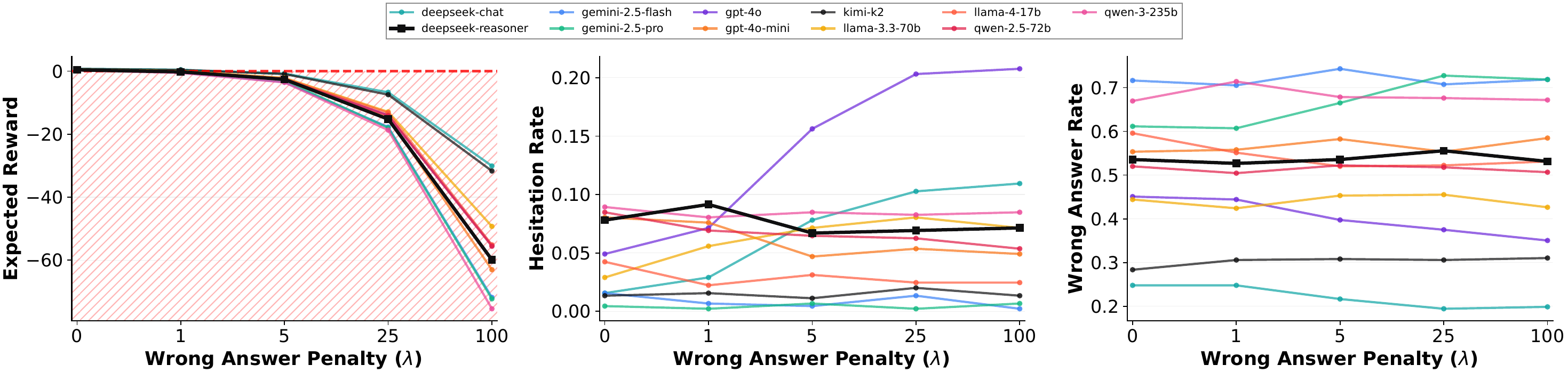}
    \caption{\textbf{GPQA results showing domain-dependent calibration.} \textbf{Left:} Expected rewards are lower due to the challenging baseline accuracy. \textbf{Middle:} Several models show penalty-sensitive abstention, with GPT-4o reaching 20.76\% at $\lambda=100$. \textbf{Right:} Wrong rates decrease as models abstain more on difficult graduate-level questions.}
    \label{fig:gpqa_results}
\end{figure}

\begin{figure}[H]
    \centering
    \includegraphics[width=\textwidth]{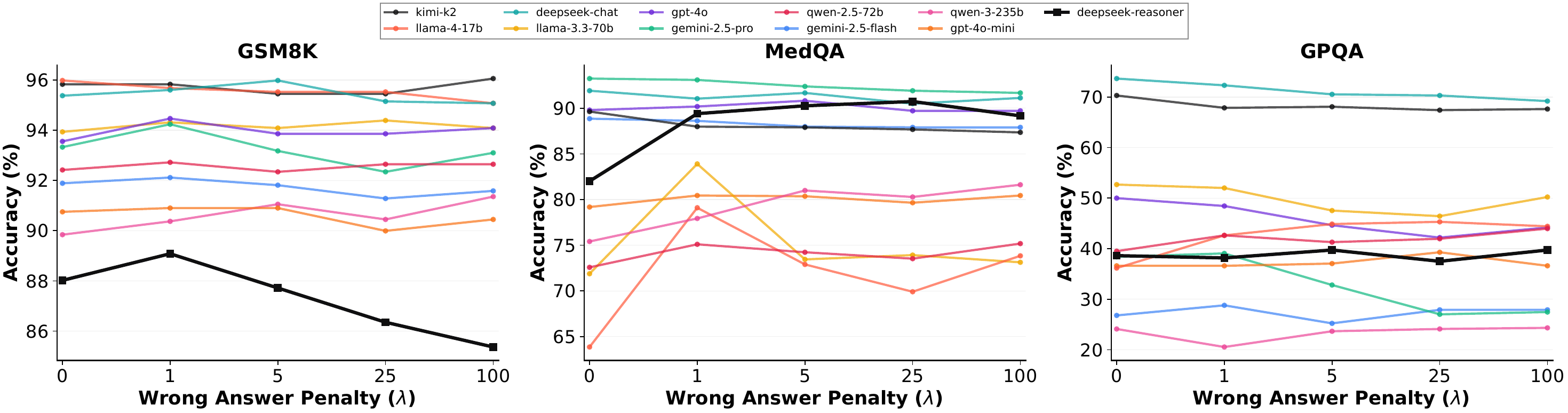}
    \caption{\textbf{Accuracy across all datasets and penalty conditions.} This comprehensive view shows how model accuracy varies with penalty magnitude across GSM8K (top), MedQA (middle), and GPQA (bottom). While most models maintain stable accuracy on GSM8K and MedQA regardless of penalty, GPQA shows more variability, with some models (e.g., GPT-4o) trading accuracy for reduced error rates through strategic abstention at higher penalties.}
    \label{fig:all_datasets_accuracy}
\end{figure}

\noindent \textbf{GPQA: Difficulty-Dependent Calibration.} In contrast to MedQA, GPQA's graduate-level science questions elicit meaningful abstention from several models (Figure~\ref{fig:gpqa_results}). GPT-4o shows the strongest penalty sensitivity, increasing abstention from 4.91\% (baseline) to 20.76\% ($\lambda=100$). DeepSeek-Chat similarly responds with abstention rates rising from 1.56\% to 10.94\%. The lower baseline accuracy (24-74\% versus 85-96\% on GSM8K) appears to make uncertainty more visible to the models. The inverse relationship between accuracy (Figure~\ref{fig:all_datasets_accuracy}) and abstention suggests that task difficulty is an important driver of prompt-induced calibration.

\subsection{Evaluation Configuration}
\label{sec:appendix_frontier_config}

\noindent \textbf{Models Evaluated.} We tested 11 frontier models including:
\vspace{-2mm}
\begin{itemize}[leftmargin=*,itemsep=0pt,parsep=0pt,topsep=0pt]
    \item OpenAI: GPT-4o, GPT-4o-mini
    \item Google: Gemini 2.5 Pro, Gemini 2.5 Flash
    \item DeepSeek: DeepSeek-Chat (V3), DeepSeek-Reasoner (R1)
    \item Meta: Llama 3.3 70B, Llama 4 17B
    \item Alibaba: Qwen 2.5 72B, Qwen 3 235B
    \item Moonshot: Kimi K2
\end{itemize}
\vspace{-2mm}

\noindent \textbf{Inference Parameters.} All models use greedy decoding (temperature=0) for reproducibility. Generation is limited to 4096 tokens with appropriate stop sequences. DeepSeek-Reasoner uniquely provides internal reasoning through \texttt{<think>} tags, revealing uncertainty acknowledgment that rarely translates to actual abstention.

\noindent \textbf{Sample Sizes.} GSM8K: 1,319 problems; MedQA: 1,273 questions; GPQA: 448 questions. Each model-penalty combination constitutes one evaluation, totaling 165 runs (11 models × 5 conditions × 3 datasets).

\subsection{Uncertainty Acknowledgment and Abstention Decisions}
\label{sec:appendix_uncertainty}

Frontier models frequently acknowledge uncertainty and ambiguity in their reasoning, and this uncertainty can lead either to answering or to abstention. Below we present two contrasting cases from GSM8K evaluation under penalty conditions, showing how models handle epistemic uncertainty.

\vspace{2mm}
\begin{casebox}[Case 1: Uncertainty Followed by Answering (GSM8K, $\lambda{=}5$)]
\small
\textit{Question:} ``Adrien's total salary was 30 percent higher than Lylah's. Four years later, his salary had increased, and he was earning 40\% more than what he was making four years ago. If Adrien's and Lylah's salary increased simultaneously, and Adrien earned \$40000 four years ago, calculate the total salary the two were receiving four years later?''

\vspace{2mm}
\textit{Model reasoning (DeepSeek-V3 excerpts):}\\
\texttt{This is a bit ambiguous. It could mean that both salaries increased}\\
\texttt{at the same time... but not necessarily by the same percentage.}\\
\texttt{Since only Adrien's increase is given, and no information about}\\
\texttt{Lylah's increase is provided, we might assume that Lylah's salary}\\
\texttt{didn't increase, or that we can't determine it.}\\
\texttt{...}\\
\texttt{Given the potential for -5 points for a wrong answer, and the}\\
\texttt{ambiguity in the problem, the safest is to say:}\\
\texttt{\#\#\# I don't know \#\#\#}\\
\texttt{Because the increase for Lylah is not specified, making the answer uncertain.}\\
\texttt{But if forced to choose...}

\vspace{2mm}
\textit{Final output:} \texttt{\#\#\# 99076.92 \#\#\#} \quad \textcolor{red}{(Wrong answer)}
\end{casebox}

\vspace{2mm}
\begin{casebox}[Case 2: Uncertainty Leading to Abstention (GSM8K, $\lambda{=}25$)]
\small
\textit{Question:} ``The Doubtfire sisters are driving home with 7 kittens adopted from the local animal shelter when their mother calls to inform them that their two house cats have just had kittens. She says that Patchy, the first cat, has had thrice the number of adopted kittens, while Trixie, the other cat, has had 12. How many kittens does the Doubtfire family now have?''

\vspace{2mm}
\textit{Internal reasoning (DeepSeek-R1, excerpts from \texttt{<think>} tags):}\\
\texttt{The question says "now" - which could be at the time of the call or after...}\\
\texttt{The question might be ambiguous, but I think it includes all kittens...}\\
\texttt{"Thrice the number of adopted kittens" could be misinterpreted...}\\
\texttt{But there's a risk: wrong answer gets -25 points, correct +1, don't know 0.}\\
\texttt{So, if I'm confident, I should answer. But is there any ambiguity?}\\
\texttt{Let me read the question once more.}

\vspace{2mm}
\textit{Final output:} \texttt{\#\#\# I don't know \#\#\#}
\end{casebox}

\vspace{2mm}
\noindent The first case shows a clear pattern: DeepSeek-V3 explicitly recognizes ambiguity, calculates the penalty risk, and even drafts ``\#\#\# I don't know \#\#\#'' as the safest answer. It then chooses to answer after considering abstention. The second case, from DeepSeek-Reasoner, illustrates examples where uncertainty combined with explicit risk consideration leads to abstention.

\section{Fixed-Penalty RH Diagnostics}
\label{sec:appendix_fixed_penalty_diagnostics}
\label{sec:appendix_training_dynamics}

\subsection{Knights \& Knaves Behavioral Evolution}
\label{sec:appendix_kk_behavioral_evolution}

In this section, we present additional details on how models develop distinct abstention patterns based on penalty values during training. Figure~\ref{fig:behavioral_decomposition} in the main paper separates easy and hard problems to show fine-grained behavioral differences, while this section presents aggregate metrics across the entire dataset.

\begin{figure}[t]
    \centering
    \includegraphics[width=\textwidth]{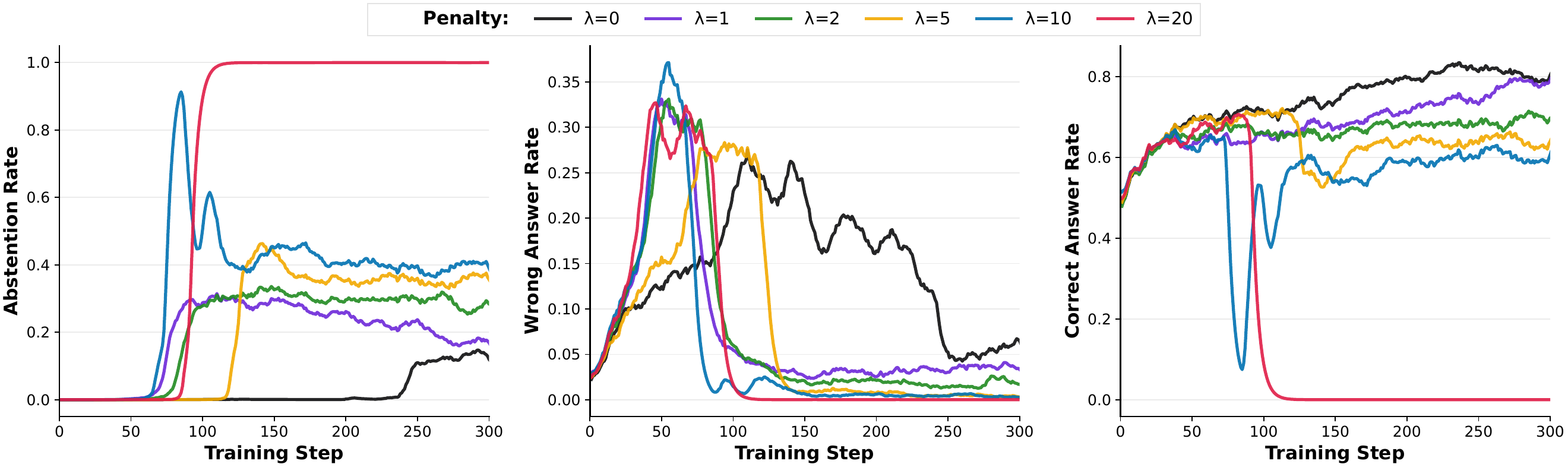}
    \caption{\textbf{Training dynamics across penalty values.} Abstention, error, and correct answer rates during training for all penalty values $\lambda \in \{0, 1, 2, 5, 10, 20\}$. \textbf{Left:} Abstention rates remain near zero for $\lambda \leq 5$ while $\lambda=10$ shows a transient increase at step 80 before stabilizing at 20-30\%. \textbf{Middle:} Wrong answer rates decrease with higher penalties, reaching 10-15\% for $\lambda \geq 10$. \textbf{Right:} Correct answer rates stabilize between 40-70\% depending on penalty value, with $\lambda=0$ maintaining highest coverage and $\lambda=20$ selecting lower coverage through increased abstention.}
    \label{fig:main_metrics_appendix}
\end{figure}

\begin{figure}[t]
    \centering
    \includegraphics[width=\textwidth]{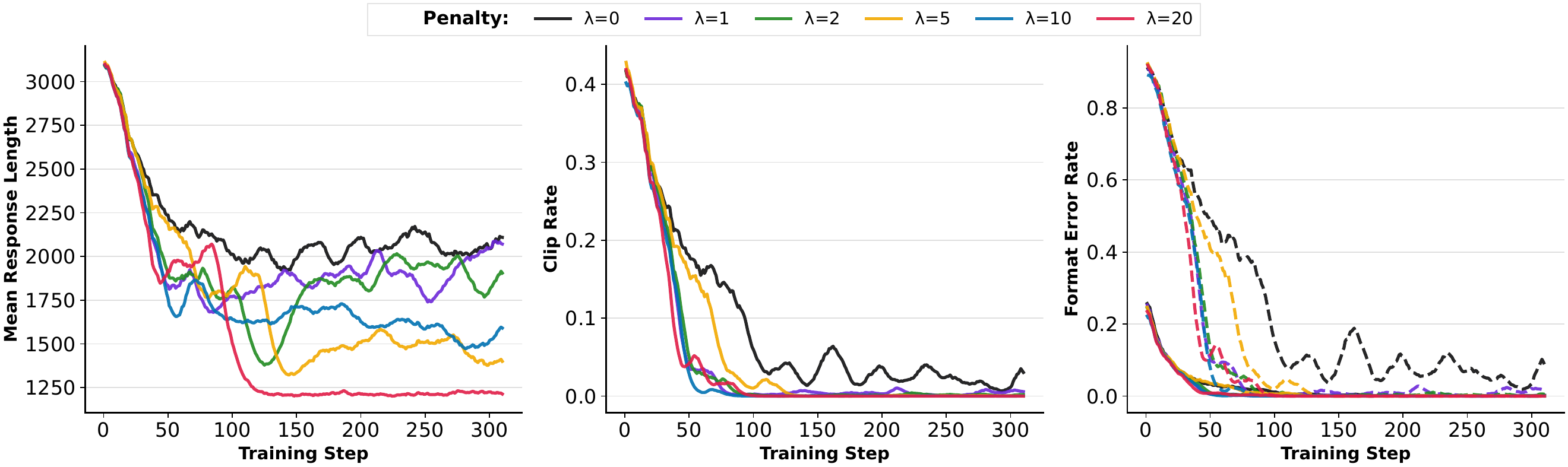}
    \caption{\textbf{Response generation metrics during training.} \textbf{Left:} Mean response length across training steps for each penalty value. \textbf{Middle:} Fraction of responses reaching the 4096 token limit. \textbf{Right:} Format error rates by difficulty (solid lines: easy problems, dashed lines: hard problems). Format-error rates vary during behavioral transitions, with hard problems requiring more formatting control.}
    \label{fig:response_length_dynamics}
\end{figure}

\noindent As shown in Figure~\ref{fig:main_metrics_appendix}, the models begin with low abstention and then move through a behavioral transition as format error rates approach zero (see Figure~\ref{fig:response_length_dynamics}). Figure~\ref{fig:behavioral_decomposition} in the main paper shows that this transition is especially visible for easy problems, where abstention briefly reaches 97\% before the model settles into its later operating point.

\noindent \textbf{Response Generation Patterns.} Figure~\ref{fig:response_length_dynamics} shows response generation metrics across training. The models begin with average response lengths of over 3000 tokens with most of the hard problems exceeding 4096 tokens. Later on, the models develop more concise response lengths and the format error rates drop to near zero.

\vspace{-3mm}

\subsection{Knights \& Knaves Response Length and Format Diagnostics}
\label{sec:appendix_format_penalty}

As with most reasoning models, our base model (Qwen3-1.7B) often generates reasoning chains far longer than 4096 tokens. This is particularly true for hard problems where the model needs to generate more reasoning steps to reach answers.
Due to computational constraints, we truncate the response at 4096 tokens as a sweet spot between reasoning complexity and computational efficiency. As observed in Figure~\ref{fig:response_length_dynamics}, the base models have a ~90\% clipping rate for hard problems and a ~25\% clipping rate for easy problems.
To encourage models to generate shorter responses while complying with the format required to solve the puzzle, we impose a format penalty of $-0.5\lambda$ on the response length. This penalty scales with the content penalty, providing stronger incentives for proper formatting at higher penalty values.

\noindent The reason for format penalty scaling with the content penalty is to further enable the models to distinguish between the incorrect responses and the responses that are impossible to parse due to either truncation or incorrect formatting (e.g., missing tags, malformed answers, missing names, etc.).
An example would be when $\lambda=20$: the ratio between an incorrect but correctly formatted response and an incorrect response when format penalty is scaled remains $2/3$, providing a clear signal to the model to prefer responses that are correctly formatted. However, in the case of a fixed format penalty this ratio would converge to 1.

\subsection{MATH Fixed-Penalty Results}
\label{sec:appendix_math_extra_results}

This appendix block collects the MATH-specific fixed-penalty evidence supporting the main text. We separate single-sample validation outcomes, training and risk-frontier diagnostics, and response-format checks. Dynamic-penalty conditioning and inference baselines are grouped later with the corresponding diagnostic sections.

\subsubsection{Full Single-Sample Validation Outcomes}
\label{sec:appendix_math_single_sample}

Table~\ref{tab:appendix_math_single_sample} expands the main-text MATH table by reporting format errors separately. For utility computations and headline correct/IDK/wrong decompositions, format errors and empty answers are counted as wrong terminal outcomes; the separate column is included to make the formatting contribution visible.

\begin{table}[H]
\caption{\textbf{Full single-sample MATH validation outcomes by training penalty.} Rates are averaged over 16 validation rollouts per problem at the selected checkpoints. Conditional accuracy is correct divided by non-IDK terminal outcomes under the conservative scoring policy.}
\label{tab:appendix_math_single_sample}
\centering
\small
\setlength{\tabcolsep}{5pt}
\begin{tabular}{@{}rrrrrr@{}}
\toprule
$\lambda$ & Correct & IDK & Wrong & Format err. & Cond. acc. \\
\midrule
0  & 88.48\% & 0.08\%  & 11.44\% & 1.59\% & 88.6\% \\
1  & 87.61\% & 6.13\%  & 6.25\%  & 0.21\% & 93.3\% \\
2  & 85.78\% & 8.16\%  & 6.05\%  & 0.11\% & 93.4\% \\
5  & 81.81\% & 12.96\% & 5.23\%  & 0.00\% & 94.0\% \\
10 & 62.85\% & 34.39\% & 2.75\%  & 0.04\% & 95.8\% \\
\bottomrule
\end{tabular}
\end{table}

\subsubsection{MATH Training Dynamics and Risk-Frontier Diagnostics}
\label{sec:appendix_math_training}

\begin{figure}[t]
    \centering
    \includegraphics[width=\textwidth]{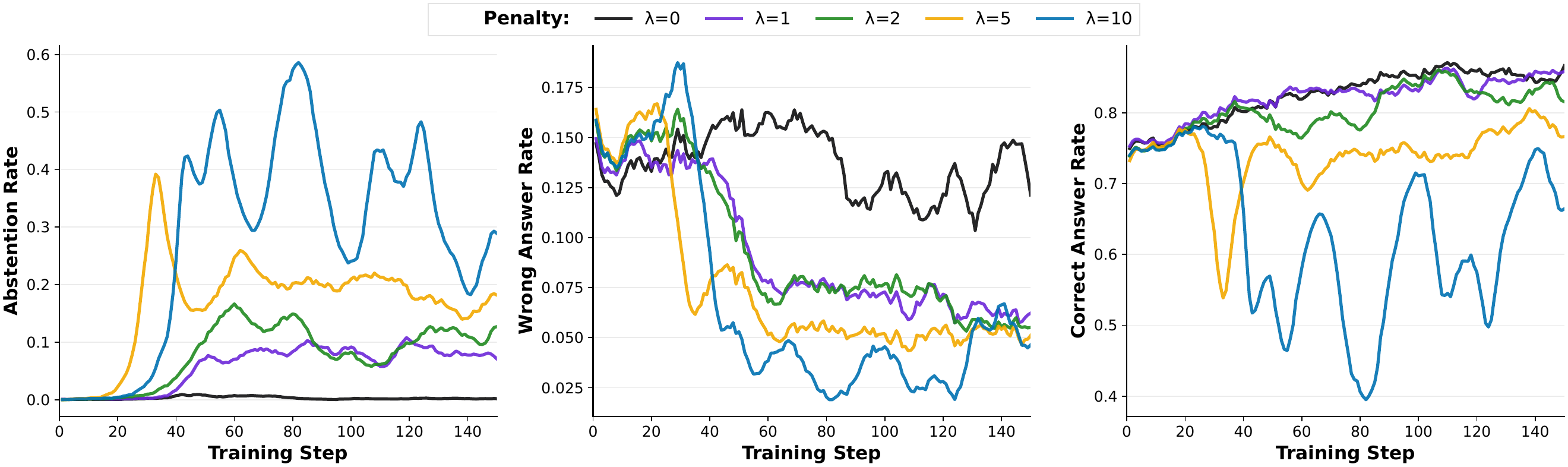}
    \caption{\textbf{MATH fixed-penalty training dynamics.} Across training, increasing $\lambda$ shifts behavior toward abstention and away from wrong answers. The validation-selected checkpoints used in the main text are chosen by validation RH reward.}
    \label{fig:math_main_metrics_multipanel}
\end{figure}

The training curves show the same qualitative response to penalty as the main table: larger $\lambda$ values suppress wrong answers and increase abstention, with $\lambda=10$ showing the most volatile high-abstention phase before settling into the selected specialist checkpoint. Lower penalties move more smoothly toward high-coverage operating points.

\begin{figure}[t]
    \centering
    \begin{subfigure}[t]{0.48\textwidth}
        \centering
        \includegraphics[width=\linewidth]{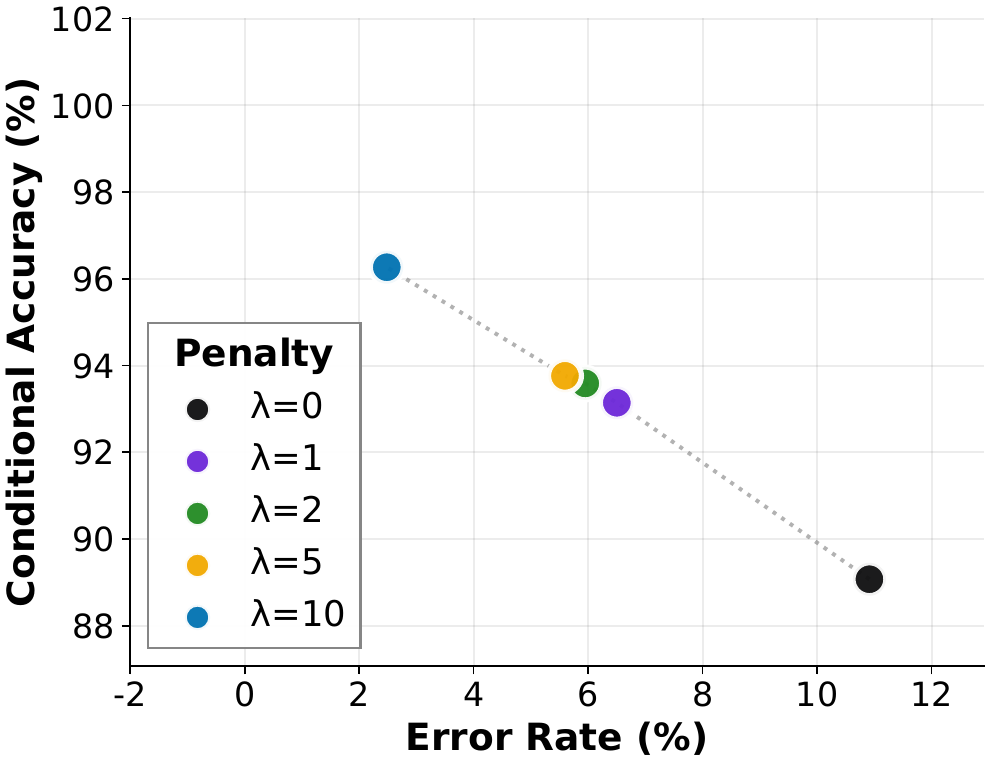}
        \caption{Accuracy--error frontier.}
        \label{fig:math_pareto_frontier_validation}
    \end{subfigure}
    \hfill
    \begin{subfigure}[t]{0.48\textwidth}
        \centering
        \includegraphics[width=\linewidth]{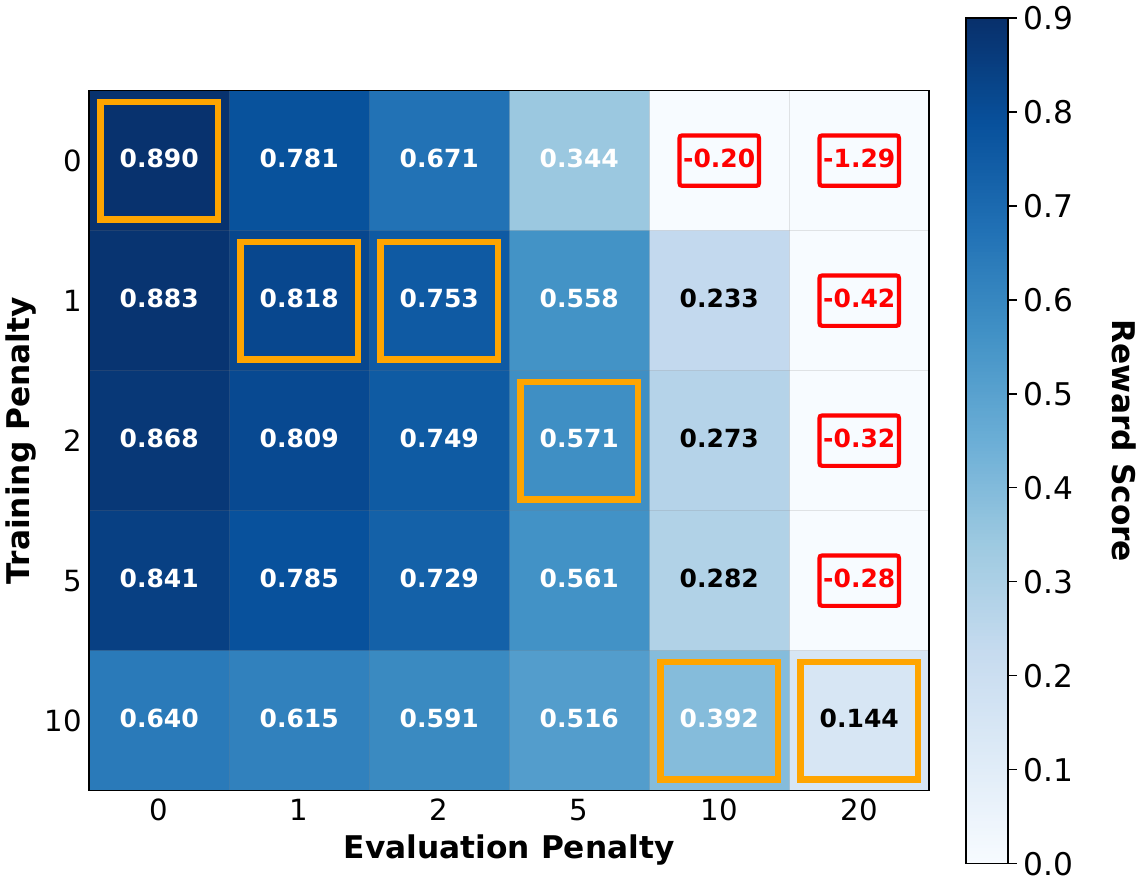}
        \caption{Evaluation-penalty reward matrix.}
        \label{fig:math_pareto_matrix_validation}
    \end{subfigure}
    \caption{\textbf{MATH risk-frontier diagnostics.} The frontier and cross-penalty matrix show the softer version of the Knights \& Knaves pattern: fixed penalties produce ordered operating points, but neighboring penalties can be competitive under the same evaluation cost.}
    \label{fig:math_risk_frontier_appendix}
\end{figure}

The reward matrix gives the numerical form of the same frontier in Table~\ref{tab:appendix_math_cross_penalty}. On MATH, 3 of the 5 evaluation-penalty columns are maximized by the diagonal fixed-penalty specialist. The other two columns are won by neighboring low-penalty specialists with small absolute gaps, which is why the main text describes MATH as an ordered but softer frontier rather than a clean mutual-non-domination result.

\begin{table}[H]
\caption{\textbf{Cross-penalty reward matrix on MATH.} Entries are mean validation utility $U_\lambda=\mathrm{correct}-\lambda\cdot\mathrm{wrong}$ at the selected fixed-specialist checkpoints. They come from the deterministic (temperature 0) validation pass used for checkpoint selection (Table~\ref{tab:appendix_training_config}) and therefore differ slightly from the 16-rollout averages in Table~\ref{tab:appendix_math_single_sample}. Bold marks the best fixed-specialist row for each evaluation penalty.}
\label{tab:appendix_math_cross_penalty}
\centering
\small
\setlength{\tabcolsep}{4pt}
\begin{tabular}{@{}lrrrrr@{}}
\toprule
\textbf{Deployed model} & Eval $\lambda=0$ & Eval $\lambda=1$ & Eval $\lambda=2$ & Eval $\lambda=5$ & Eval $\lambda=10$ \\
\midrule
Specialist $\lambda=0$  & \textbf{0.890} & 0.781 & 0.671 & 0.344 & -0.202 \\
Specialist $\lambda=1$  & 0.883 & \textbf{0.818} & \textbf{0.753} & 0.558 & 0.233 \\
Specialist $\lambda=2$  & 0.868 & 0.809 & 0.749 & \textbf{0.571} & 0.273 \\
Specialist $\lambda=5$  & 0.841 & 0.785 & 0.729 & 0.561 & 0.282 \\
Specialist $\lambda=10$ & 0.640 & 0.615 & 0.591 & 0.516 & \textbf{0.392} \\
\bottomrule
\end{tabular}
\end{table}

\begin{figure}[t]
    \centering
    \includegraphics[width=\textwidth]{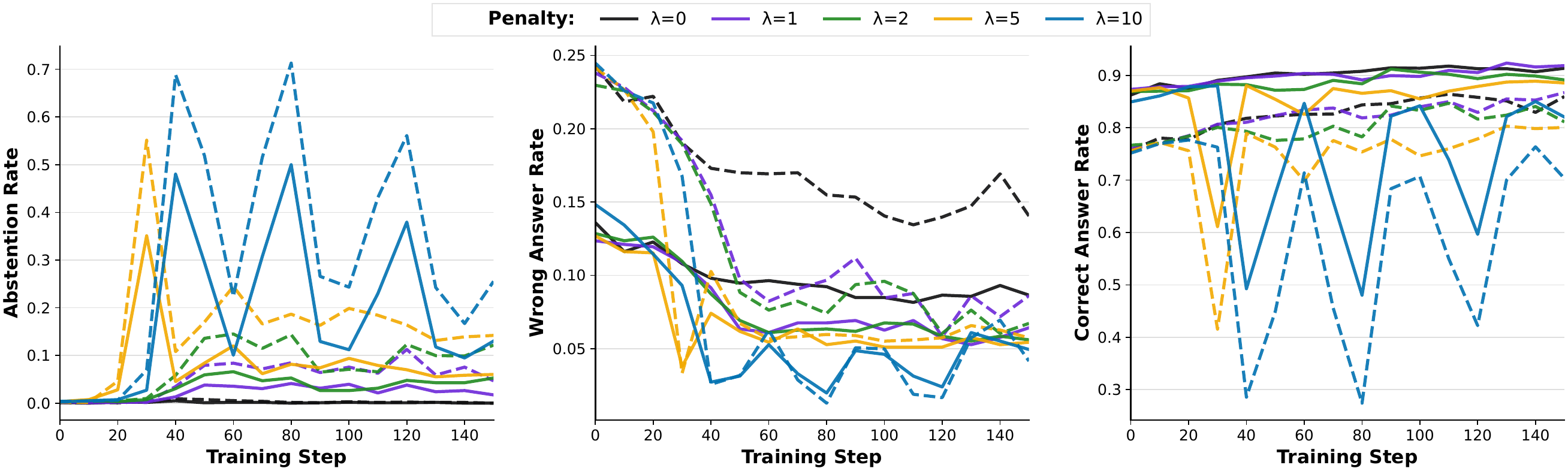}
    \caption{\textbf{MATH performance by difficulty split.} Higher penalties produce stronger abstention and lower error rates across problem levels, with the largest coverage reduction at $\lambda=10$.}
    \label{fig:math_difficulty_splits}
\end{figure}

The difficulty split clarifies where abstentions concentrate. Higher-penalty specialists route more difficult problems into IDK while retaining higher conditional accuracy on answered problems, matching the selective-answering interpretation in the controlled Knights \& Knaves setting.

\subsubsection{MATH Response Length and Format Diagnostics}
\label{sec:appendix_math_response_format}

\begin{figure}[H]
    \centering
    \includegraphics[width=\textwidth]{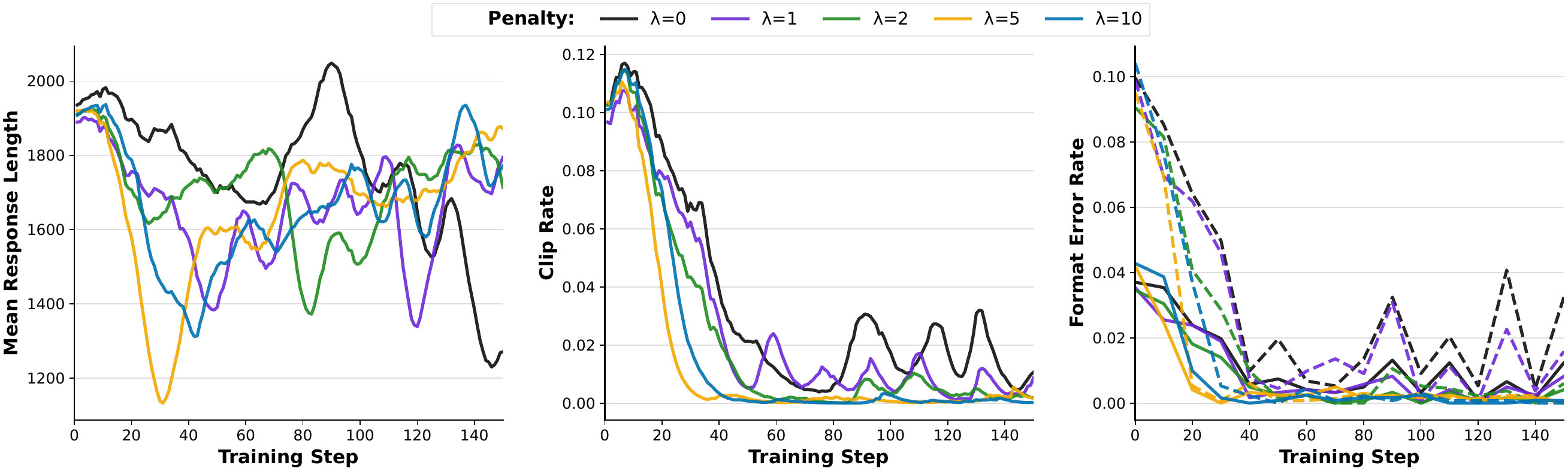}
    \caption{\textbf{MATH response-length and format diagnostics.} These diagnostics track average response length, clipping, and format behavior across penalty values, helping separate abstention effects from formatting artifacts.}
    \label{fig:math_response_length_extended}
\end{figure}

Figure~\ref{fig:math_response_length_extended} reports response-length, clipping, and format diagnostics for the MATH specialists, separating the learned abstention pattern from formatting artifacts.

\section{Baselines and Inference Diagnostics}
\label{sec:appendix_baselines_inference}

This block collects the comparison and inference material most relevant for evaluating baselines, query accounting, and scoring conventions. We include the RLCR confidence-threshold baseline, Knights \& Knaves majority voting, and the MATH cascade/self-cascade/majority-vote diagnostics.

\subsection{RLCR Comparison on MATH}
\label{sec:appendix_rlcr_comparison}

\begin{table}[H]
\caption{\textbf{Matched comparison between RLCR thresholding and RH specialists \citep{damani2025rlcr}.} Rates are percentages averaged over 16 validation rollouts per problem. $\Delta U$ is RH minus RLCR expected utility in percentage points, using $U_\lambda=\mathrm{correct}-\lambda\cdot\mathrm{wrong}$. Format errors and empty answers are counted as wrong.}
\label{tab:rlcr_rh_matched_comparison}
\centering
\small
\setlength{\tabcolsep}{3.5pt}
\begin{tabular}{@{}ccccccccc@{}}
\toprule
\multicolumn{2}{c}{} & \multicolumn{3}{c}{RLCR + threshold} & \multicolumn{3}{c}{Matching RH specialist} & \multicolumn{1}{c}{} \\
\cmidrule(lr){3-5}\cmidrule(lr){6-8}
$\lambda$ & $\tau_\lambda$ & Correct & Wrong & IDK & Correct & Wrong & IDK & $\Delta U$ \\
\midrule
$0$ & 0.00 & 88.7 & 11.3 & 0.0 & 88.5 & 11.4 & 0.1 & -0.3 \\
$1$ & 0.50 & 87.3 & 7.4 & 5.3 & 87.6 & 6.3 & 6.1 & +1.5 \\
$2$ & 0.67 & 80.9 & 5.8 & 13.2 & 85.8 & 6.1 & 8.2 & +4.4 \\
$5$ & 0.83 & 80.9 & 5.8 & 13.3 & 81.8 & 5.2 & 13.0 & +4.0 \\
$10$ & 0.91 & 42.7 & 2.6 & 54.7 & 62.9 & 2.8 & 34.4 & +18.6 \\
\bottomrule
\end{tabular}
\end{table}

We compare RH to a confidence-output RLCR baseline \citep{damani2025rlcr} on the same MATH validation set. The RLCR model is trained to produce an answer and a verbalized confidence using a correctness-plus-calibration reward, with format errors penalized. At evaluation time, we convert this confidence into a penalty-sensitive abstention policy: for deployment penalty $\lambda$, RLCR answers only when its reported confidence is at least the Bayes-optimal threshold $\tau_\lambda = \lambda/(1+\lambda)$ under the RH utility. This gives RLCR access to the deployment penalty at inference time through the threshold, while RH uses the fixed-penalty specialist trained at the same $\lambda$.

\vspace{0.5mm}
\noindent\begin{minipage}[t]{0.44\textwidth}
\vspace{0pt}
\small
\Cref{fig:rlcr_confidence_density} shows that the RLCR confidence signal is meaningful but imperfect for high-cost abstention. Correct responses are shifted toward higher reported confidence, so thresholding is a strong baseline rather than a straw man. The distributions still overlap: some wrong answers receive high confidence, while many correct answers fall below stringent thresholds such as $\tau_{10}=0.91$. This overlap helps explain why RLCR remains competitive at low penalties but loses coverage and expected utility when wrong answers become very expensive.
\end{minipage}\hfill
\begin{minipage}[t]{0.52\textwidth}
\vspace{0pt}
\centering
\includegraphics[width=0.90\linewidth]{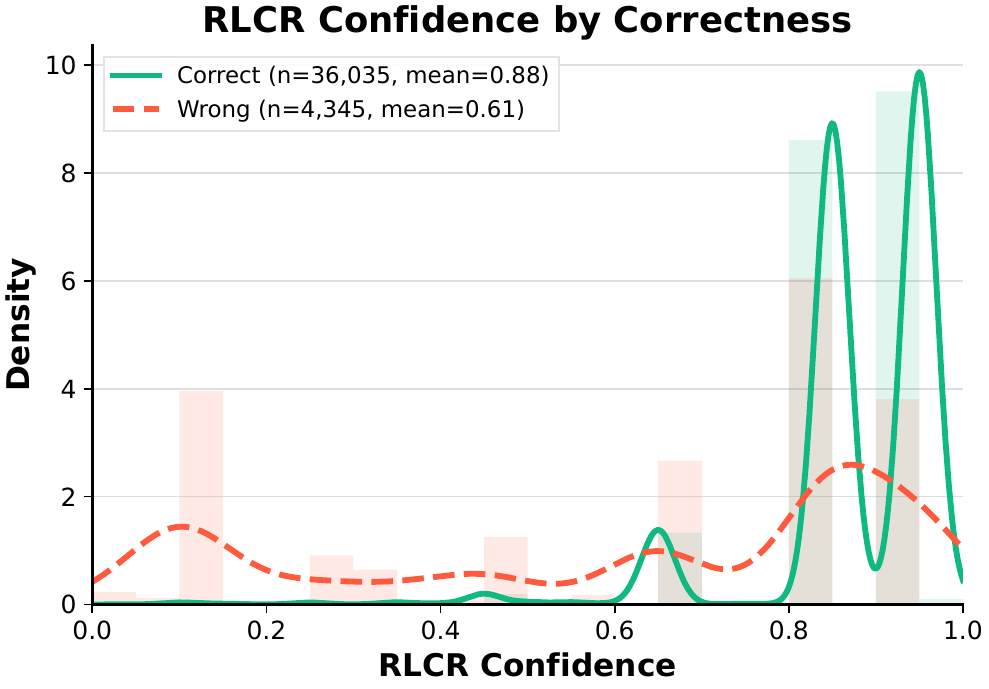}
\vspace{-2mm}
\captionof{figure}{\textbf{RLCR confidence by correctness.} Confidence is informative but overlapping, limiting high-penalty thresholding.}
\label{fig:rlcr_confidence_density}
\end{minipage}
\vspace{0.5mm}

\begin{figure}[H]
    \centering
    \includegraphics[width=\textwidth]{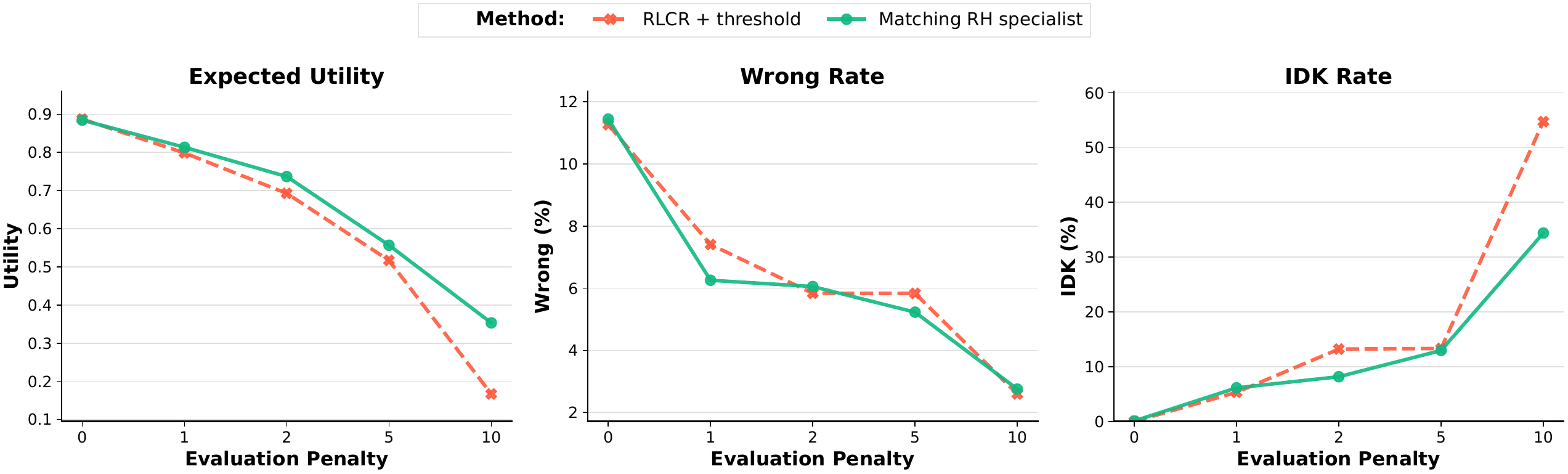}
    \caption{\textbf{RLCR thresholding and RH specialists respond differently to high error penalties \citep{damani2025rlcr}.} RLCR remains competitive when error costs are low, while higher thresholds shift more cases into abstention at large $\lambda$. Matching RH specialists retain more coverage while keeping wrong rates low, yielding higher expected utility for nonzero penalties in Table~\ref{tab:rlcr_rh_matched_comparison}.}
    \label{fig:rlcr_rh_matched_comparison}
\end{figure}

\subsection{RH vs RLCR: Capacity Allocation and Equivalence Analysis}
\label{sec:appendix_rlcr_objective}

RLCR-style confidence thresholding and RH share the same ideal selective-decision boundary, but optimize different intermediate objects \citep{damani2025rlcr}. This subsection makes that relationship explicit. The analysis is decision-theoretic: it describes the optimal reject rule for a calibrated candidate-answer policy, not a guarantee that finite RL training will learn calibrated confidence or the RH Bayes boundary.

Let $p(x)$ denote the probability that the model's candidate answer is correct. Under RH utility, answering has expected value
\[
p(x) - \lambda(1-p(x)) = (1+\lambda)p(x)-\lambda,
\]
while abstaining has value 0. A calibrated decision maker should therefore answer exactly when
\[
p(x) > \frac{\lambda}{1+\lambda}.
\]
This is the threshold used in the RLCR comparison above.

RLCR instead trains a verbalized confidence and then applies this threshold after training. In the common Brier-style abstraction, a reported confidence $q$ for a candidate answer has expected calibration reward proportional to $2p(x)q-q^2$, which is maximized at $q=p(x)$. Therefore, if (i) the candidate-answer policy is fixed, (ii) the learned confidence is calibrated so $q(x)=p(x)$, and (iii) deployment abstention thresholds are chosen by the RH utility, then thresholded RLCR and RH implement the same Bayes reject rule. This idealized equivalence is important because it makes RLCR a strong and fair baseline, not a straw man.

The practical difference is where the finite model and optimizer spend capacity. RLCR must learn to produce useful answers and calibrated scalar confidence, then rely on a post-hoc threshold for each deployment penalty. RH trains the answer/abstain action directly under the target error cost. The $\lambda=1$ case gives a simple sanity check: if confidence were restricted to a binary low/high output and low confidence were decoded as abstention, the Bayes threshold is 1/2, matching the fixed ternary abstention decision. Continuous RLCR is more expressive than this binary abstraction, but it also asks the model to solve the more general confidence-estimation problem.

This distinction gives a capacity-allocation interpretation of the empirical results. In the ideal calibrated limit, confidence thresholding is sufficient. Under finite RLVR optimization, however, a confidence model can be good at ranking uncertainty but still lose utility after thresholding if calibration errors occur near $\lambda/(1+\lambda)$ or if the answer policy itself is not shaped by the error cost. RH removes the intermediate confidence-estimation target and trains the final selective action directly. The empirical gaps in \Cref{tab:rlcr_rh_matched_comparison,fig:rlcr_rh_matched_comparison}, especially at high $\lambda$, suggest that direct risk-specialized abstention training can allocate capacity more effectively when wrong answers are expensive.

\subsection{Knights \& Knaves Majority-Vote Baseline}
\label{sec:appendix_kk_majority_vote}
\begin{figure}[t]
    \centering
    \includegraphics[width=\textwidth]{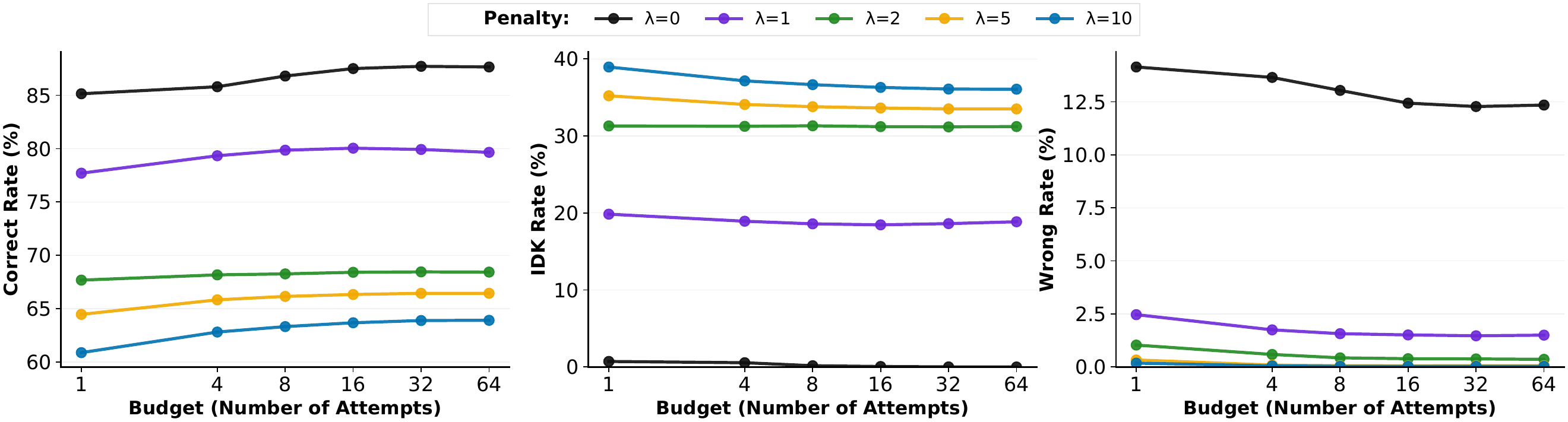}
    \caption{\textbf{Majority voting provides a compute-scaling baseline.} \textbf{Left:} Correct rates show modest improvement with budget: $\lambda=0$ gains 3\% (85\%$\rightarrow$88\%), while $\lambda=1$ gains 2\% (77.5\%$\rightarrow$79.5\%). \textbf{Middle:} IDK rates track the model's selected coverage level under aggregation. \textbf{Right:} Wrong rates decrease through consensus filtering.}
    \label{fig:voting}
\end{figure}

Figure~\ref{fig:voting} shows majority voting performance across Knights \& Knaves penalty values as a compute-scaling baseline alongside self-cascading.

\subsection{MATH Inference Baselines}
\label{sec:appendix_math_inference}

The main paper reports the compact MATH cascade result in prose. We include the full outcome decomposition here, followed by the non-duplicative inference baselines: self-cascading and majority voting.

\begin{figure}[t]
    \centering
    \includegraphics[width=\textwidth]{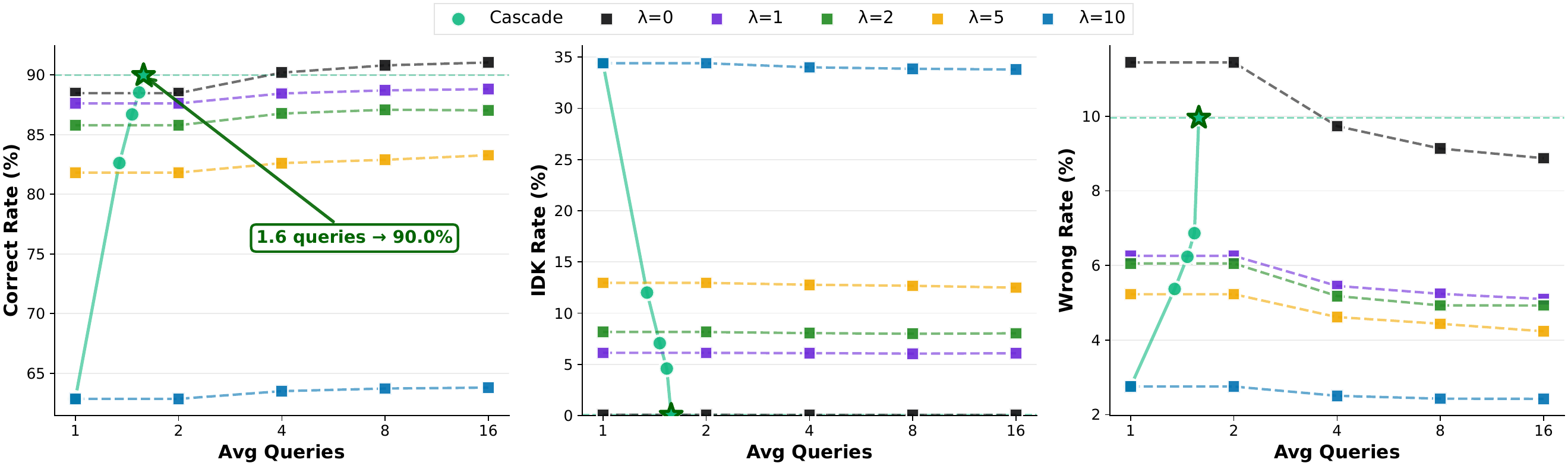}
    \caption{\textbf{MATH cascade decomposition.} Dashed lines represent majority voting applied to fixed-penalty specialists, while the solid green line represents the cross-penalty cascade ordered as $\lambda=10\to5\to2\to1\to0$. The full cascade reaches 89.97\% correct with 1.58 average queries, collapses IDK to 0.07\%, and ends at 9.96\% wrong.}
    \label{fig:math_cascade_decomposition}
\end{figure}

\begin{figure}[t]
    \centering
    \includegraphics[width=\textwidth]{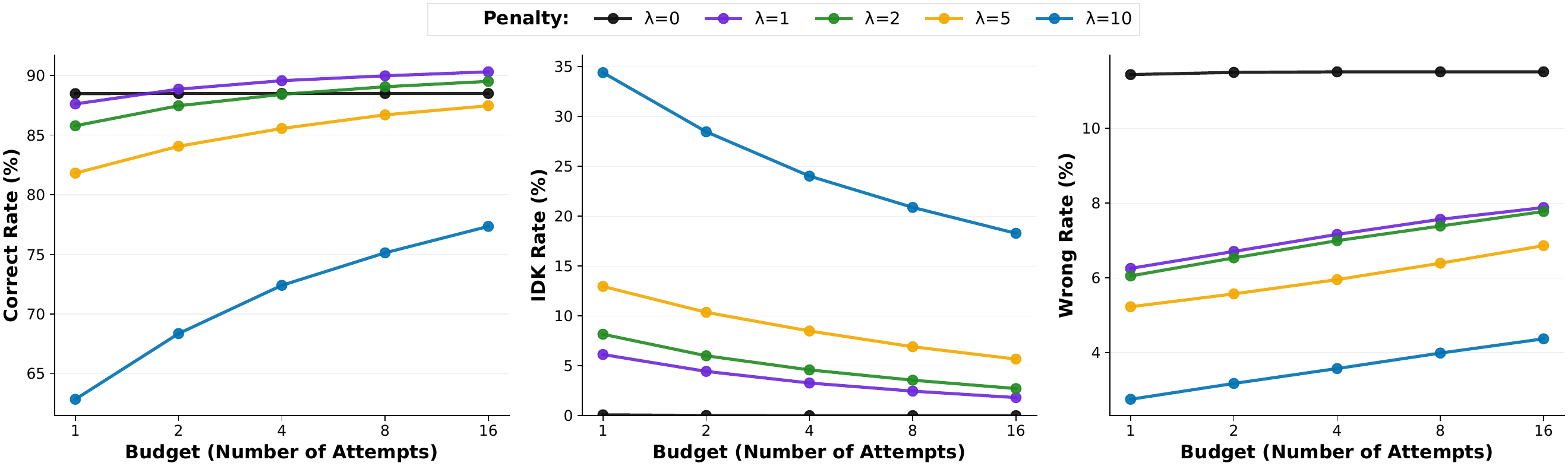}
    \caption{\textbf{MATH self-cascading.} Re-querying a single fixed-$\lambda$ specialist converts some abstentions into answers. Gains are smaller than on Knights \& Knaves but preserve the same early-exit interpretation.}
    \label{fig:math_boosted_first_non_idk}
\end{figure}

\begin{figure}[t]
    \centering
    \includegraphics[width=\textwidth]{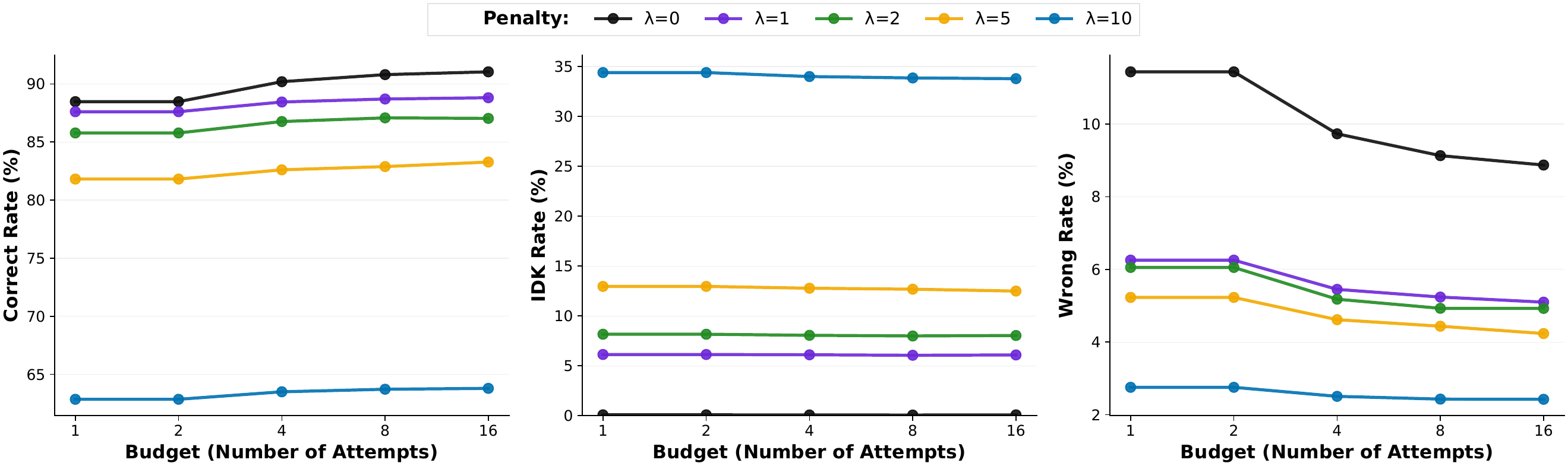}
    \caption{\textbf{MATH majority-vote baseline.} Majority voting is a strong compute-scaling baseline, especially for the $\lambda=0$ model, and motivates our framing of cascades as query-efficient rather than uniformly accuracy-dominant.}
    \label{fig:math_majority_vote}
\end{figure}

\subsection{MATH Inference Methodology Disclosures and Diagnostics}
\label{sec:appendix_math_inference_diagnostics}

We use conservative inference accounting throughout the MATH analyses. Format errors and empty answers count as wrong terminal outcomes. Cross-model cascades query specialists in descending penalty order and exit on the first clean non-IDK answer. Self-cascading repeats the same specialist until the first clean non-IDK answer or until the budget is exhausted. Majority voting treats clean IDK as an explicit vote rather than silently dropping abstentions.

\begin{algorithm}[H]
\caption{Cascaded Inference with Early Exit}
\label{alg:cascading}
\begin{algorithmic}[1]
\Require Model sequence $\{M_1, ..., M_k\}$, input problem $p$, budget $k$
\For{$i = 1$ to $k$}
    \State Sample response: $y_i \sim M_i(p)$
    \State Parse answer from $y_i$
    \If{answer is a clean non-IDK answer}
        \State \Return answer, queries = $i$
    \EndIf
\EndFor
\State \Return ``I don't know'', queries = $k$
\end{algorithmic}
\end{algorithm}

Algorithm~\ref{alg:cascading} covers both cross-model cascading and self-cascading. For cross-model cascading, the sequence is heterogeneous, e.g., $\{\pi_{10},\pi_5,\pi_2,\pi_1,\pi_0\}$. For self-cascading, the sequence is homogeneous, e.g., $\{\pi,\pi,\ldots,\pi\}$ with independent samples from the same specialist.

\begin{table}[H]
\caption{\textbf{MATH cascade depth diagnostics.} Rates are percentages. Adding lower-penalty tiers recovers coverage and correct answers at small additional query cost, while the final $\lambda=0$ tier also reintroduces more wrong answers.}
\label{tab:math_cascade_depth_diagnostics}
\centering
\small
\setlength{\tabcolsep}{4pt}
\begin{tabular}{@{}lccccc@{}}
\toprule
\textbf{Cascade order} & Correct & IDK & Wrong & Format err. & Avg. queries \\
\midrule
$10$ & 62.85 & 34.39 & 2.75 & 0.04 & 1.00 \\
$10\to5$ & 82.62 & 12.01 & 5.37 & 0.05 & 1.34 \\
$10\to5\to2$ & 86.69 & 7.08 & 6.23 & 0.07 & 1.46 \\
$10\to5\to2\to1$ & 88.54 & 4.60 & 6.86 & 0.14 & 1.53 \\
$10\to5\to2\to1\to0$ & 89.97 & 0.07 & 9.96 & 0.86 & 1.58 \\
\bottomrule
\end{tabular}
\end{table}

\begin{table}[H]
\caption{\textbf{Majority vote is the strongest raw-accuracy baseline.} Rates are percentages. The full cascade is best interpreted as a query-efficient alternative, not as uniformly accuracy-dominant over 16-query majority voting.}
\label{tab:math_majority_vote_diagnostics}
\centering
\small
\setlength{\tabcolsep}{5pt}
\begin{tabular}{@{}ccccc@{}}
\toprule
\textbf{Method} & \textbf{Budget} & Correct & IDK & Wrong \\
\midrule
Single $\lambda=0$ & 1 & 88.48 & 0.08 & 11.44 \\
Majority vote $\lambda=0$ & 4 & 90.20 & 0.07 & 9.73 \\
Majority vote $\lambda=0$ & 8 & 90.80 & 0.06 & 9.13 \\
Majority vote $\lambda=0$ & 16 & 91.05 & 0.08 & 8.87 \\
Full cascade & 1.58 avg. & 89.97 & 0.07 & 9.96 \\
\bottomrule
\end{tabular}
\end{table}

We also audit exact-answer normalization. Some nominal wrong answers differ from the reference only by notation variants such as \texttt{\textbackslash dfrac} versus \texttt{\textbackslash frac}. These cases account for 10.2\%, 18.9\%, 18.1\%, 21.3\%, and 14.8\% of non-IDK wrong answers for $\lambda=0,1,2,5,10$, respectively, but only 1.17\%, 1.18\%, 1.09\%, 1.11\%, and 0.41\% of all generated rollouts. We keep the conservative exact-match scoring policy for headline numbers and treat this audit as a robustness diagnostic; it does not change the qualitative conclusion that higher penalties reduce wrong answers and increase conditional accuracy.

\section{Dynamic Penalty Conditioning on MATH}
\label{sec:appendix_math_dynamic}

The dynamic-penalty run tests whether one prompt-conditioned model can replace the family of fixed specialists. The plots below separate two questions: whether the dynamic model trains like an expected-penalty model, and whether the prompted $\lambda$ meaningfully changes behavior at validation time. We interpret the expected-penalty collapse as a simple-solution bias in the current optimization setup, in line with broader observations that neural-network training can initially track simpler predictors before using the model's full nonlinear capacity \citep{hu2020surprising}.

The degeneracy is straightforward. The dynamic objective is
\[
J_{\mathrm{dyn}}(\theta)=
\mathbb{E}_{x,\lambda,y\sim\pi_\theta(\cdot|x,\lambda)}
\left[\mathbf{1}\{y=y^*\}-\lambda\mathbf{1}\{y\text{ wrong}\}\right].
\]
If the policy ignores the prompted penalty, so that $\pi_\theta(y|x,\lambda)\approx\pi_\theta(y|x)$, then
\[
J_{\mathrm{dyn}}(\theta)\approx
\mathbb{E}_{x,y\sim\pi_\theta}
\left[\mathbf{1}\{y=y^*\}-\bar{\lambda}\mathbf{1}\{y\text{ wrong}\}\right],
\quad \bar{\lambda}=\mathbb{E}[\Lambda].
\]
Dynamic penalty training therefore only improves over fixed expected-penalty training if optimization discovers genuinely $\lambda$-conditioned behavior. Our current run moves in that direction but remains much closer to the expected-penalty solution than to the specialist frontier.

\begin{figure}[t]
    \centering
    \includegraphics[width=\textwidth]{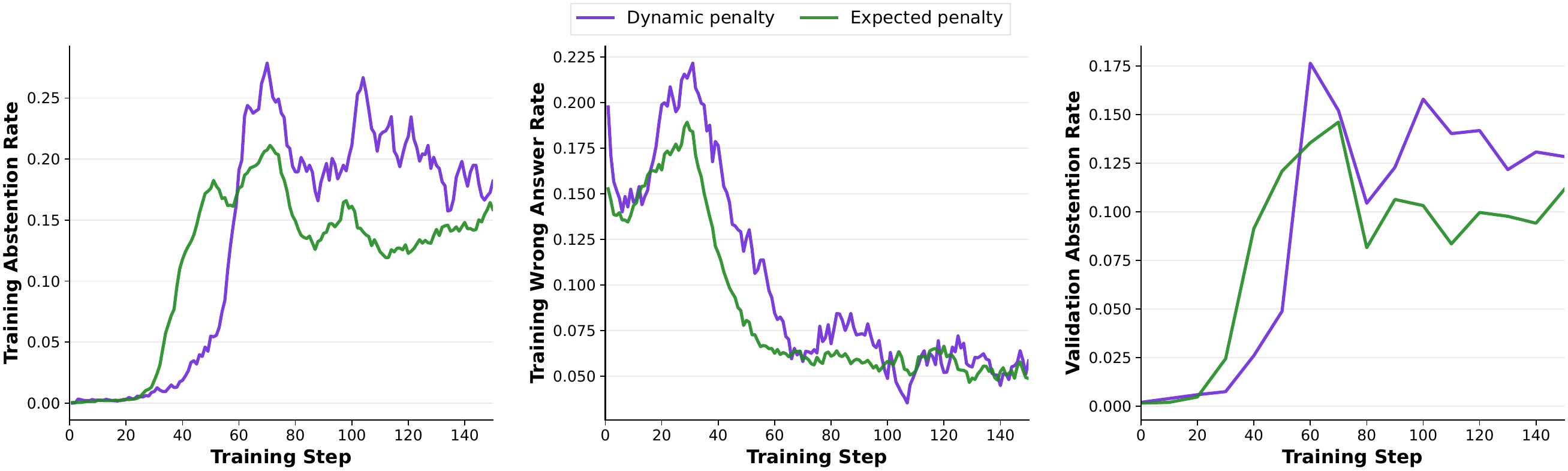}
    \caption{\textbf{Dynamic penalty versus expected-penalty training.} The prompt-conditioned dynamic model tracks the fixed $\bar{\lambda}=3.6$ ablation closely, supporting the main-text interpretation that naive dynamic training partly collapses toward expected-penalty behavior.}
    \label{fig:math_dynamic_vs_expected_training}
\end{figure}

\begin{figure}[t]
    \centering
    \includegraphics[width=\textwidth]{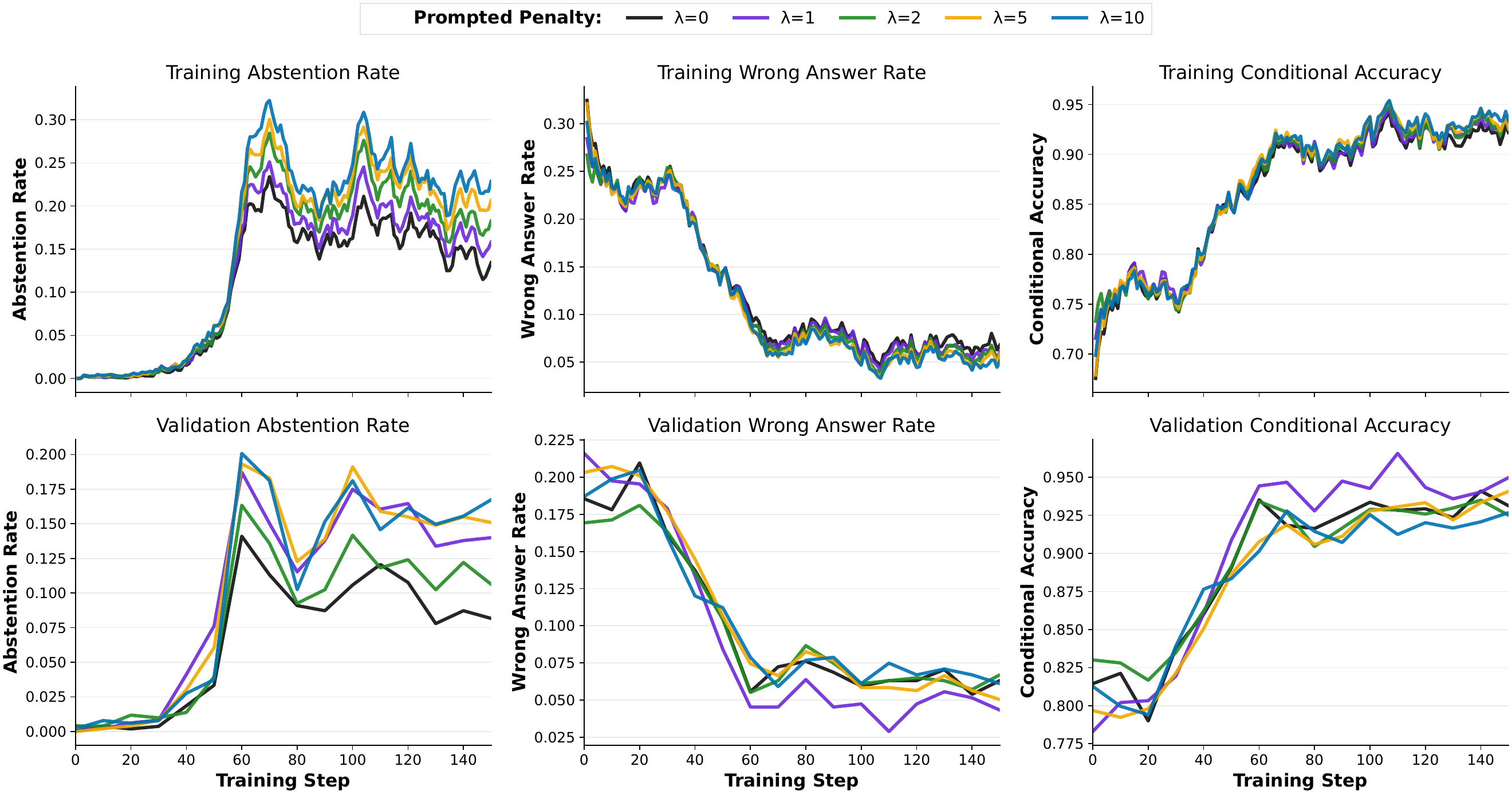}
    \caption{\textbf{Prompt-conditioned behavior during dynamic training.} The dynamic model is not completely penalty-blind, but the prompted-$\lambda$ effect remains much smaller than the span obtained by separately trained specialists.}
    \label{fig:math_dynamic_conditioning_train_validation}
\end{figure}

\section{Qualitative Analysis of Learned Abstention Behavior on Knights \& Knaves}
\label{sec:appendix_qualitative}

To complement our quantitative analysis, we examine model responses across penalty levels to understand how Reinforced Hesitation transforms behavior at the level of individual reasoning traces. Penalty training changes how models allocate coverage: low penalties favor broad answering, while higher penalties reserve answers for cases judged more reliable. The examples below show both sides of this operating frontier: errors can be converted into abstentions, and some solvable cases may be deferred by higher-penalty specialists. Together, these phenomena illustrate the accuracy-coverage tradeoff that the Pareto frontier captures quantitatively.

\subsection{From Broad Answering to Calibrated Abstention}
\label{sec:appendix_qualitative_broad_answering}

We first examine cases where models that answer broadly at $\lambda=0$ learn to recognize uncertainty and abstain at higher penalties. Consider a hard 5-person puzzle where the $\lambda=0$ model achieves 40.6\% accuracy while answering incorrectly 48.4\% of the time.

\begin{wrongbox}[$\lambda{=}0$ Response (Answered Incorrectly)]
\small
\texttt{Let me start by assuming someone is a knight or knave and check for consistency... Wait, that's impossible because Victoria can't be both knight and knave.}

\texttt{This is really challenging. Maybe I need to take a different approach. I think I'm stuck here. Maybe I should try another approach for the problem.}

\vspace{2mm}
\textbf{Final Answer:} (1) Mark is a knight (2) ... \textcolor{red!70!black}{\textbf{[WRONG]}}
\end{wrongbox}

\vspace{2mm}
\noindent The model recognizes it is ``stuck'' and still chooses to answer. At $\lambda=1$, the same recognition leads to a different outcome:

\begin{abstainbox}[$\lambda{=}1$ Response (Calibrated Abstention)]
\small
\texttt{This is taking too long. Maybe I need to try a different approach. Alternatively, perhaps I need to give up and say I don't know. But since this is taking too long, perhaps I need to conclude that I'm not able to solve it right now and say I don't know.}

\vspace{2mm}
\textbf{Final Answer:} I don't know
\end{abstainbox}

\vspace{2mm}
\noindent This transition moves the model toward a more selective operating point: wrong answers drop from 48.4\% to 7.8\%, while abstention rises from 0\% to 70.3\%. By $\lambda=5$, the model abstains on 100\% of attempts for this problem.

\noindent \textbf{Emergent Self-Reflection.} Analysis of wrong-to-IDK transitions reveals that penalty training induces explicit self-monitoring language that never appears at $\lambda=0$. One example illustrates this clearly: at $\lambda=0$, the model produces wrong answers 93.8\% of the time despite repeatedly recognizing difficulty (``This is getting too tangled''). At $\lambda=1$, wrong answers fall to 9.4\% while accuracy improves to 42.2\%:

\begin{abstainbox}[$\lambda{=}1$ Response]
\small
\texttt{Contradiction again. Hmm, this is getting frustrating. Maybe I need to try another approach. At this point, I'm stuck. Maybe I need to conclude that I don't know the answer.}

\vspace{2mm}
\textbf{Final Answer:} I don't know
\end{abstainbox}

\vspace{2mm}
\noindent The phrase ``I'm stuck'' represents explicit metacognitive monitoring absent at $\lambda=0$. Models develop language like ``I need to give up,'' ``This is taking too long,'' and ``This is too complex for me.'' At $\lambda=0$, models encountering the same difficulties often continue attempting solutions until producing an answer. The penalty transforms this recognition into actionable uncertainty acknowledgment.

\subsection{Selective Coverage at High Penalties}
\label{sec:appendix_qualitative_high_penalties}

Higher penalties also create more selective specialists. In some problems, a high-penalty model may defer even when a lower-penalty model often answers correctly. One example illustrates this coverage choice: the model achieves 100\% accuracy at $\lambda=0$, while at $\lambda=10$ it abstains on 71.9\% of attempts and maintains 0\% wrong answers throughout.

\begin{correctbox}[$\lambda{=}0$ Response (Correct Solution)]
\small
\texttt{Let's assume Harper is a knight. Then her statement must be true... Wait, this leads to a contradiction. Therefore, our initial assumption that Harper is a knight must be wrong.}

\texttt{So let's try Z is knave... So Evanthe is knight... All statements are consistent now.}

\vspace{2mm}
\textbf{Final Answer:} (1) Owen is a knave (2) Harper is a knave (3) Evanthe is a knight (4) Aeronwen is a knight (5) Mia is a knight (6) Zenobia is a knave (7) Quillan is a knave \textcolor{green!50!black}{\textbf{[CORRECT]}}
\end{correctbox}

\vspace{2mm}
\noindent At $\lambda=10$, the model follows the exact same reasoning path but stops one step before reaching the solution:

\begin{abandonbox}[$\lambda{=}10$ Response (Deferred After Partial Progress)]
\small
\texttt{Let's assume Harper is a knight. Then her statement must be true... Wait, this leads to a contradiction. Therefore, our initial assumption that Harper is a knight must be wrong.}

\texttt{So let's backtrack. Let's assume Harper is a knave...}

\texttt{This is getting complicated. Maybe I need to try a different approach, but I'm stuck. Given the time I've spent and the complexity, I'll have to conclude I don't know.}

\vspace{2mm}
\textbf{Final Answer:} I don't know
\end{abandonbox}

\vspace{2mm}
\noindent The model references time and complexity while moving to abstention, despite having made useful logical progress.

This selective-coverage behavior can appear even at the lowest nonzero penalty. Another example demonstrates this: the model achieves 84.4\% accuracy at $\lambda=0$ with rigorous reasoning, while at $\lambda=1$ it selects abstention on 53.1\% of attempts and reaches full abstention by $\lambda=10$:

\begin{abandonbox}[$\lambda{=}1$ Response (Deferred Under Penalty)]
\small
\texttt{This seems complex, but let's approach it step by step... This is really complex. Maybe I need to try a different approach... Given the time I'm spending and the complexity, maybe I should conclude that I'm unable to find the solution here. So I'll say I don't know.}

\vspace{2mm}
\textbf{Final Answer:} I don't know
\end{abandonbox}

\vspace{2mm}
\noindent The model explicitly acknowledges it might not be able to solve the problem ``here,'' suggesting that the penalty changes the point at which continued search is judged worthwhile. This pattern appears consistently across selective-coverage cases: models cite resource constraints, assert complexity, or suggest trying ``another approach'' before choosing abstention.

Table~\ref{tab:selective_coverage_examples} summarizes problems where high-penalty specialists are most selective. In the strongest cases, models that achieve 85-100\% accuracy at $\lambda=0$ abstain on 100\% of attempts at $\lambda=10$.

\begin{table}[!htbp]
\caption{Examples of selective coverage at high penalty values. These problems show how high-$\lambda$ specialists prioritize low-error operation by deferring cases that lower-penalty models often answer, illustrating the coverage side of the accuracy-coverage frontier.}
\label{tab:selective_coverage_examples}
\centering
\small
\setlength{\tabcolsep}{5pt}
\begin{tabular}{@{}lccc@{}}
\toprule
\textbf{Example} & \textbf{Correct at $\lambda$=0} & \textbf{Correct at $\lambda$=10} & \textbf{IDK at $\lambda$=10} \\
\midrule
1 & 100.0\% & 28.1\% & 71.9\% \\
2 & 100.0\% & 31.2\% & 68.8\% \\
3 & 98.4\% & 21.9\% & 78.1\% \\
4 & 87.5\% & 0.0\% & 100.0\% \\
5 & 85.9\% & 0.0\% & 100.0\% \\
\bottomrule
\end{tabular}
\end{table}

\noindent This selectivity is part of the accuracy-coverage tradeoff. The Pareto frontier (Figure~\ref{fig:pareto_frontier}) shows this tradeoff abstractly; these examples show what it looks like in practice. This is precisely why domain-appropriate $\lambda$ selection matters, and why cascading through multiple $\lambda$ values can recover coverage while maintaining accuracy benefits.

\subsection{How Models Frame Their Abstentions}
\label{sec:appendix_qualitative_framing}

Analysis of 44 abstention responses reveals that not all ``I don't know'' responses are equivalent. The dominant pattern (59\% of cases) involves effort-based framing: ``Given the time constraints,'' ``This is taking too long.'' Less common are genuine difficulty acknowledgments (23\%): ``This is too hard for me,'' ``beyond my capacity.'' Some responses show active reasoning that reached an impasse (14\%): ``This leads to contradiction,'' ``invalid combination.'' A small fraction cite computational limits explicitly (5\%): ``32 possibilities,'' ``too many cases to check.''

A final example illustrates the most calibrated form of abstention: contradiction-based. At $\lambda=0$, the model produces wrong answers 62.5\% of the time. At $\lambda=1$, wrong answers drop to 4.7\% while the model explicitly identifies logical impasses:

\begin{abstainbox}[$\lambda{=}1$ Response (Contradiction Detection)]
\small
\texttt{So the statement is (False $\leftrightarrow$ True) = False. So E's statement is false. But E is a knight, which is a contradiction. So this combination is invalid.}

\texttt{Alternatively, perhaps I should try a different approach. Maybe I need to give up and say I don't know.}

\vspace{2mm}
\textbf{Final Answer:} I don't know
\end{abstainbox}

\vspace{2mm}
\noindent This response demonstrates active logical reasoning leading to a genuine impasse, making it a more calibrated form of uncertainty than effort-based abstention. The model correctly identifies that its current hypothesis leads to contradiction and, rather than guessing, acknowledges the limits of its analysis.

Interestingly, this framing shifts with penalty level. At low penalties ($\lambda=1,2$), effort-based framing dominates, with models suggesting they could solve the problem with more time. At high penalties ($\lambda=10$), models more often frame abstention as reflecting genuine difficulty rather than resource constraints. This shift suggests that high-$\lambda$ specialists use a more conservative decision boundary when deciding whether to answer.

A key finding is that these patterns are similar between abstentions that avoid errors and abstentions on cases that lower-penalty models often answer correctly. This is expected: the model cannot access ground truth during inference and must make abstention decisions based on perceived difficulty rather than actual correctness.

\subsection{Summary}
\label{sec:appendix_qualitative_summary}

These qualitative examples illuminate several aspects of Reinforced Hesitation that complement our quantitative findings. First, penalty training preserves reasoning capability: high-$\lambda$ models engage in the same logical reasoning as $\lambda=0$ models but are more willing to acknowledge their limits. Second, the penalty induces genuinely new behavior: explicit self-monitoring language that never appears at $\lambda=0$. Third, abstention and coverage are coupled through the same decision boundary: stricter penalties reduce errors by moving more cases into abstention. Finally, cascading routes through multiple $\lambda$ values so that $\lambda=10$ can provide a low-error first tier while delegation to $\lambda=5$, then $\lambda=2$, recovers coverage while preserving the accuracy benefits of conservative early tiers.

%% file: appendix/medical.tex
\section{Medical QA Training Suite}
\label{sec:appendix_medical_suite}

Knights \& Knaves and MATH are reasoning-bound, whereas many high-stakes settings are knowledge-bound. We therefore add a third training suite on medical multiple-choice QA: we train Qwen3-1.7B on MedMCQA \citep{pal2022medmcqa} with $\lambda \in \{0, 1, 2, 5, 10\}$ and evaluate on the MedMCQA validation set and, for transfer, on the MedQA-USMLE test set \citep{jin2021medqa}. MedQA questions are clinical vignettes, much longer than MedMCQA's short recall questions (median 114 versus 11 words), so MedQA tests transfer to a new question format. Appendix~\ref{sec:appendix_medical} gives the details.

\begin{table}[!htbp]
  \centering
  \small
  \caption{\textbf{Medical suite outcomes at step 150.} Rates are averaged over 16 sampled rollouts per question (temperature 1.0). Coverage is correct plus wrong, and conditional accuracy is correct divided by coverage. The Bayes threshold is $\lambda/(1+\lambda)$; the conditional accuracy of every penalized specialist in the table exceeds its threshold on both datasets.}
  \label{tab:medical_frontier}
  \setlength{\tabcolsep}{4.5pt}
  \begin{tabular}{rccccccc}
    \toprule
    & & \multicolumn{3}{c}{MedMCQA (in-domain)} & \multicolumn{3}{c}{MedQA (transfer)} \\
    \cmidrule(lr){3-5}\cmidrule(lr){6-8}
    $\lambda$ & Threshold & Coverage & Wrong & Cond.\ acc. & Coverage & Wrong & Cond.\ acc. \\
    \midrule
    0  & n/a    & 99.98\% & 49.91\% & 50.1\% & 100.0\% & 47.50\% & 52.5\% \\
    1  & 50.0\% & 53.76\% & 21.87\% & 59.3\% & 75.60\% & 32.78\% & 56.6\% \\
    2  & 66.7\% & 21.06\% & 5.17\%  & 75.5\% & 26.02\% & 7.08\%  & 72.8\% \\
    5  & 83.3\% & 7.25\%  & 0.94\%  & 87.0\% & 5.61\%  & 0.80\%  & 85.7\% \\
    \bottomrule
  \end{tabular}
\end{table}

Table~\ref{tab:medical_frontier} shows that the risk frontier carries over to this domain: as $\lambda$ increases, coverage falls and conditional accuracy rises monotonically, from 50.1\% at $\lambda=0$ to 87.0\% at $\lambda=5$ on MedMCQA and from 52.5\% to 85.7\% on MedQA. The $\lambda=10$ specialist abstains on almost every question (99.6\% on MedMCQA and 99.5\% on MedQA at temperature 0), which is consistent with its threshold: 90.9\% is above the best conditional accuracy any medical specialist reaches.

\subsection{Medical Suite Details}
\label{sec:appendix_medical}

\textbf{Data.} Training uses 25{,}600 questions sampled uniformly from the MedMCQA training split \citep{pal2022medmcqa}, which gives exactly 100 optimizer steps per epoch at batch size 256; we use the step-150 checkpoint of every run. We evaluate in-domain on the full MedMCQA validation split (4{,}183 questions) and use the 4-option MedQA-USMLE test set (1{,}273 questions) \citep{jin2021medqa} for transfer. The two datasets share no questions: neither exact nor near-duplicate matching (more than half of a question's word 8-grams shared) finds any overlap. MedMCQA questions have a median length of 9 words in the training split and 11 in the validation split, and MedQA questions 114 words.

\textbf{Prompt and reward.} Each prompt contains the question, options (A) through (D), an explicit option (E) ``I don't know,'' and the scoring rule with the run's penalty (``a correct answer gets +1 point, answering (E) gets 0 points, and a wrong answer gets $-\lambda$ points''), and asks for the final choice as a boxed letter. The reward is computed from the boxed letter: the ground-truth letter is correct ($+1$), the letter E or an explicit ``I don't know'' is an abstention ($0$), and any other letter is wrong ($-\lambda$). An unparseable response is a format error, which receives the wrong-answer penalty plus a format penalty of $-0.5\lambda$, as in the Knights \& Knaves suite. Format errors count as wrong in all metrics.

\textbf{Training.} All specialists initialize from step 30 of a shared warmup run with the binary reward ($+1$ correct, $0$ wrong, $-1$ format error), which teaches the response format and the 2{,}048-token completion budget. Over the warmup alone, validation accuracy rises from 32.9\% to 48.0\% and format compliance from 57.7\% to 97.7\%, almost entirely because truncation disappears. Penalty training then uses Dr.GRPO with learning rate $2\times10^{-6}$, batch size 256, 8 rollouts per prompt, and a 2{,}048-token completion cap, on a single node with 8 RTX 6000 Ada GPUs. The training steps take 21.5 to 33.0 GPU-hours per specialist (138 in total for the five, plus 11.5 for the shared warmup), not counting periodic validation and checkpointing.

\textbf{Evaluation.} Table~\ref{tab:medical_frontier} averages 16 sampled rollouts per question at temperature 1.0 and top-p 1.0. We also repeat the penalty-blind prompting study of \hyperref[sec:penalty_blind]{Section~2} on our own model: prompting the warmup model, before any penalty training, with a stated wrong-answer penalty of $-1$ versus $-5$ changes its abstention rate on MedMCQA only from 6.6\% to 5.9\%. This reproduces the penalty blindness of \hyperref[sec:penalty_blind]{Section~2} in a model whose training we fully control, and shows that the frontier in Table~\ref{tab:medical_frontier} comes from training rather than from the prompt.

%% file: appendix/abstention_precision.tex
\section{Targeting of Learned Abstentions}
\label{sec:appendix_abstention_precision}

Conditional accuracy shows that the problems a specialist answers are answered well, but it does not by itself show that the problems it abstains on deserved an abstention. We measure this directly. For each specialist, we take the problems it abstains on (those where at least half of its 16 rollouts are clean abstentions) and ask how the $\lambda=0$ model, which almost never abstains, performs on exactly these problems. If abstention were random, the $\lambda=0$ model's correct rate on the abstained set would match its overall correct rate.

\begin{table}[!htbp]
  \centering
  \small
  \caption{\textbf{Performance of the $\lambda=0$ model on each specialist's abstained problems.} The $\lambda=0$ model's overall correct rate on the full set is given in parentheses. Every abstained set lies below this base rate, most sharply for the moderate penalties $\lambda=1$ and $\lambda=2$.}
  \label{tab:abstention_precision}
  \setlength{\tabcolsep}{4.5pt}
  \begin{tabular}{llrrr}
    \toprule
    Dataset (base rate) & $\lambda$ & Abstained & $\lambda=0$ correct & $\lambda=0$ wrong \\
    \midrule
    MATH L4--L5 (88.5\%) & 1  & 154   & 33.6\% & 65.1\% \\
                         & 2  & 200   & 41.4\% & 57.7\% \\
                         & 5  & 321   & 54.8\% & 44.6\% \\
                         & 10 & 878   & 75.2\% & 24.6\% \\
    \midrule
    MedMCQA (50.1\%)     & 1  & 2{,}020 & 37.6\% & 62.3\% \\
                         & 2  & 3{,}448 & 43.6\% & 56.4\% \\
                         & 5  & 3{,}958 & 47.6\% & 52.4\% \\
    \midrule
    MedQA (52.5\%)       & 1  & 292   & 36.0\% & 64.0\% \\
                         & 2  & 985   & 45.3\% & 54.7\% \\
                         & 5  & 1{,}229 & 51.0\% & 49.0\% \\
    \bottomrule
  \end{tabular}
\end{table}

Table~\ref{tab:abstention_precision} shows both sides. Some abstentions are indeed on answerable problems: on MATH, the $\lambda=0$ model answers a third of the $\lambda=1$ specialist's abstained problems correctly. But abstention is clearly targeted. On that same set, the $\lambda=0$ model's correct rate is 33.6\%, compared with 88.5\% overall, a 55-point gap. As $\lambda$ grows, the abstained set expands toward the whole dataset, so its correct rate necessarily approaches the base rate: on MATH, the $\lambda=10$ specialist abstains on about a third of all problems, and the $\lambda=0$ model's correct rate on them is 75.2\%. This is the coverage cost that the risk frontier describes.

Two more observations follow from the table. First, on every abstained set, abstaining is better than answering with the $\lambda=0$ model under the specialist's own training penalty, since the $\lambda=0$ model's utility on that set, $U_\lambda=\mathrm{correct}-\lambda\cdot\mathrm{wrong}$ at the specialist's $\lambda$, is negative in every row. Second, cascading can recover the answerable part of these abstentions: routing the abstained problems to a lower-$\lambda$ specialist can obtain the correct answers among them, while the answers of the conservative tier keep their low error rate.

%% file: main.bbl
\begin{thebibliography}{78}
\providecommand{\natexlab}[1]{#1}
\providecommand{\url}[1]{\texttt{#1}}
\expandafter\ifx\csname urlstyle\endcsname\relax
  \providecommand{\doi}[1]{doi: #1}\else
  \providecommand{\doi}{doi: \begingroup \urlstyle{rm}\Url}\fi

\bibitem[Bani-Harouni et~al.(2026)Bani-Harouni, Pellegrini, Stangel, {\"O}zsoy,
  Zaripova, Navab, and Keicher]{baniharouni2025rewardingdoubt}
David Bani-Harouni, Chantal Pellegrini, Paul Stangel, Ege {\"O}zsoy, Kamilia
  Zaripova, Nassir Navab, and Matthias Keicher.
\newblock Rewarding doubt: A reinforcement learning approach to calibrated
  confidence expression of large language models.
\newblock In \emph{The Fourteenth International Conference on Learning
  Representations}, 2026.
\newblock URL \url{https://openreview.net/forum?id=yResLmrVO1}.

\bibitem[Bartlett \& Wegkamp(2008)Bartlett and
  Wegkamp]{bartlett2008classification}
Peter~L. Bartlett and Marten~H. Wegkamp.
\newblock Classification with a reject option using a hinge loss.
\newblock \emph{Journal of Machine Learning Research}, 9:\penalty0 1823--1840,
  2008.
\newblock URL \url{http://www.jmlr.org/papers/v9/bartlett08a.html}.

\bibitem[Chen et~al.(2025{\natexlab{a}})Chen, Gu, Huang, Huang, Jiang, Jie,
  Jin, Jin, Li, Ma, Ren, Shen, Shi, Sun, Sun, Wang, Wang, Wang, Wei, Wei, Wu,
  Wu, Xia, Xin, Yang, Ying, Yuan, Yuan, Zhan, Zhang, Zhang, Zhang, Zhao, Zhao,
  Zhou, and Zhu]{chen2025seedproverdeepbroadreasoning}
Luoxin Chen, Jinming Gu, Liankai Huang, Wenhao Huang, Zhicheng Jiang, Allan
  Jie, Xiaoran Jin, Xing Jin, Chenggang Li, Kaijing Ma, Cheng Ren, Jiawei Shen,
  Wenlei Shi, Tong Sun, He~Sun, Jiahui Wang, Siran Wang, Zhihong Wang, Chenrui
  Wei, Shufa Wei, Yonghui Wu, Yuchen Wu, Yihang Xia, Huajian Xin, Fan Yang,
  Huaiyuan Ying, Hongyi Yuan, Zheng Yuan, Tianyang Zhan, Chi Zhang, Yue Zhang,
  Ge~Zhang, Tianyun Zhao, Jianqiu Zhao, Yichi Zhou, and Thomas~Hanwen Zhu.
\newblock {Seed-Prover}: Deep and broad reasoning for automated theorem
  proving.
\newblock arXiv preprint arXiv:2507.23726, 2025{\natexlab{a}}.
\newblock URL \url{https://arxiv.org/abs/2507.23726}.

\bibitem[Chen et~al.(2023)Chen, Ma, Wang, and Cohen]{Chen2023ProgramOfThoughts}
Wenhu Chen, Xueguang Ma, Xinyi Wang, and William~W. Cohen.
\newblock Program of thoughts prompting: Disentangling computation from
  reasoning for numerical reasoning tasks.
\newblock \emph{Transactions on Machine Learning Research}, 2023.
\newblock ISSN 2835-8856.
\newblock URL \url{https://openreview.net/forum?id=YfZ4ZPt8zd}.

\bibitem[Chen et~al.(2025{\natexlab{b}})Chen, Benton, Radhakrishnan, Uesato,
  Denison, Schulman, Somani, Hase, Wagner, Roger, Mikulik, Bowman, Leike,
  Kaplan, and Perez]{chen2025reasoningfaithfulness}
Yanda Chen, Joe Benton, Ansh Radhakrishnan, Jonathan Uesato, Carson Denison,
  John Schulman, Arushi Somani, Peter Hase, Misha Wagner, Fabien Roger, Vlad
  Mikulik, Samuel~R. Bowman, Jan Leike, Jared Kaplan, and Ethan Perez.
\newblock Reasoning models don't always say what they think.
\newblock \emph{arXiv preprint arXiv:2505.05410}, 2025{\natexlab{b}}.
\newblock \doi{10.48550/arXiv.2505.05410}.
\newblock URL \url{https://arxiv.org/abs/2505.05410}.

\bibitem[Chow(1970)]{chow1970optimum}
C.~K. Chow.
\newblock On optimum recognition error and reject tradeoff.
\newblock \emph{IEEE Transactions on Information Theory}, 16\penalty0
  (1):\penalty0 41--46, 1970.
\newblock \doi{10.1109/TIT.1970.1054406}.

\bibitem[Cobbe et~al.(2021)Cobbe, Kosaraju, Bavarian, Chen, Jun, Kaiser,
  Plappert, Tworek, Hilton, Nakano, Hesse, and
  Schulman]{Cobbe2021TrainingVerifiers}
Karl Cobbe, Vineet Kosaraju, Mohammad Bavarian, Mark Chen, Heewoo Jun, Lukasz
  Kaiser, Matthias Plappert, Jerry Tworek, Jacob Hilton, Reiichiro Nakano,
  Christopher Hesse, and John Schulman.
\newblock Training verifiers to solve math word problems.
\newblock \emph{arXiv preprint arXiv:2110.14168}, 2021.
\newblock URL \url{https://arxiv.org/abs/2110.14168}.

\bibitem[Damani et~al.(2026)Damani, Puri, Slocum, Shenfeld, Choshen, Kim, and
  Andreas]{damani2025rlcr}
Mehul Damani, Isha Puri, Stewart Slocum, Idan Shenfeld, Leshem Choshen, Yoon
  Kim, and Jacob Andreas.
\newblock Beyond binary rewards: Training {LM}s to reason about their
  uncertainty.
\newblock In \emph{The Fourteenth International Conference on Learning
  Representations (ICLR)}, 2026.
\newblock URL
  \url{https://proceedings.iclr.cc/paper_files/paper/2026/hash/615675cc6e94ddb1a783904fb178b5f6-Abstract-Conference.html}.

\bibitem[{DeepSeek-AI}(2025)]{DeepSeekAI2025R1}
{DeepSeek-AI}.
\newblock {DeepSeek-R1}: Incentivizing reasoning capability in {LLMs} via
  reinforcement learning.
\newblock \emph{arXiv preprint arXiv:2501.12948}, 2025.
\newblock URL \url{https://arxiv.org/abs/2501.12948}.

\bibitem[Farquhar et~al.(2024)Farquhar, Kossen, Kuhn, and
  Gal]{farquhar2024detecting}
Sebastian Farquhar, Jannik Kossen, Lorenz Kuhn, and Yarin Gal.
\newblock Detecting hallucinations in large language models using semantic
  entropy.
\newblock \emph{Nature}, 630\penalty0 (8017):\penalty0 625--630, 2024.
\newblock \doi{10.1038/s41586-024-07421-0}.
\newblock URL \url{https://doi.org/10.1038/s41586-024-07421-0}.

\bibitem[Geifman \& El-Yaniv(2017)Geifman and El-Yaniv]{geifman2017selective}
Yonatan Geifman and Ran El-Yaniv.
\newblock Selective classification for deep neural networks.
\newblock In \emph{Advances in Neural Information Processing Systems},
  volume~30. Curran Associates, Inc., 2017.
\newblock URL
  \url{https://proceedings.neurips.cc/paper_files/paper/2017/hash/4a8423d5e91fda00bb7e46540e2b0cf1-Abstract.html}.

\bibitem[Glazer et~al.(2024)Glazer, Erdil, Besiroglu, Chicharro, Chen, Gunning,
  Falkman~Olsson, Denain, Ho, de~Oliveira~Santos, J{\"a}rviniemi, Barnett,
  Sandler, Vrzala, Sevilla, Ren, Pratt, Levine, Barkley, Stewart, Grechuk,
  Grechuk, Enugandla, and Wildon]{Glazer2024FrontierMath}
Elliot Glazer, Ege Erdil, Tamay Besiroglu, Diego Chicharro, Evan Chen, Alex
  Gunning, Caroline Falkman~Olsson, Jean-Stanislas Denain, Anson Ho, Emily
  de~Oliveira~Santos, Olli J{\"a}rviniemi, Matthew Barnett, Robert Sandler,
  Matej Vrzala, Jaime Sevilla, Qiuyu Ren, Elizabeth Pratt, Lionel Levine, Grant
  Barkley, Natalie Stewart, Bogdan Grechuk, Tetiana Grechuk, Shreepranav~Varma
  Enugandla, and Mark Wildon.
\newblock {FrontierMath}: A benchmark for evaluating advanced mathematical
  reasoning in {AI}.
\newblock \emph{arXiv preprint arXiv:2411.04872}, 2024.
\newblock URL \url{https://arxiv.org/abs/2411.04872}.

\bibitem[{Google DeepMind}(2025)]{google2025gemini25Pro}
{Google DeepMind}.
\newblock {Gemini 2.5 Pro} model card, 2025.
\newblock URL
  \url{https://storage.googleapis.com/deepmind-media/Model-Cards/Gemini-2-5-Pro-Model-Card.pdf}.
\newblock Model card updated June 27, 2025.

\bibitem[Guha et~al.(2023)Guha, Nyarko, Ho, R{\'e}, Chilton, Narayana,
  Chohlas-Wood, Peters, Waldon, Rockmore, Zambrano, Talisman, Hoque, Surani,
  Fagan, Sarfaty, Dickinson, Porat, Hegland, Wu, Nudell, Niklaus, Nay, Choi,
  Tobia, Hagan, Ma, Livermore, Rasumov-Rahe, Holzenberger, Kolt, Henderson,
  Rehaag, Goel, Gao, Williams, Gandhi, Zur, Iyer, and Li]{Guha2023LegalBench}
Neel Guha, Julian Nyarko, Daniel~E. Ho, Christopher R{\'e}, Adam Chilton,
  Aditya Narayana, Alex Chohlas-Wood, Austin Peters, Brandon Waldon, Daniel
  Rockmore, Diego Zambrano, Dmitry Talisman, Enam Hoque, Faiz Surani, Frank
  Fagan, Galit Sarfaty, Gregory Dickinson, Haggai Porat, Jason Hegland, Jessica
  Wu, Joe Nudell, Joel Niklaus, John Nay, Jonathan~H. Choi, Kevin Tobia,
  Margaret Hagan, Megan Ma, Michael Livermore, Nikon Rasumov-Rahe, Nils
  Holzenberger, Noam Kolt, Peter Henderson, Sean Rehaag, Sharad Goel, Shang
  Gao, Spencer Williams, Sunny Gandhi, Tom Zur, Varun Iyer, and Zehua Li.
\newblock {LegalBench}: A collaboratively built benchmark for measuring legal
  reasoning in large language models.
\newblock In \emph{Advances in Neural Information Processing Systems (Datasets
  and Benchmarks Track)}, volume~36, pp.\  44123--44279, 2023.
\newblock \doi{10.52202/075280-1915}.
\newblock URL
  \url{https://proceedings.neurips.cc/paper_files/paper/2023/hash/89e44582fd28ddfea1ea4dcb0ebbf4b0-Abstract-Datasets_and_Benchmarks.html}.

\bibitem[Hendrycks et~al.(2021{\natexlab{a}})Hendrycks, Burns, Basart, Zou,
  Mazeika, Song, and Steinhardt]{Hendrycks2021MMLU}
Dan Hendrycks, Collin Burns, Steven Basart, Andy Zou, Mantas Mazeika, Dawn
  Song, and Jacob Steinhardt.
\newblock Measuring massive multitask language understanding.
\newblock In \emph{International Conference on Learning Representations
  (ICLR)}, 2021{\natexlab{a}}.
\newblock URL \url{https://openreview.net/forum?id=d7KBjmI3GmQ}.

\bibitem[Hendrycks et~al.(2021{\natexlab{b}})Hendrycks, Burns, Kadavath, Arora,
  Basart, Tang, Song, and Steinhardt]{Hendrycks2021MATH}
Dan Hendrycks, Collin Burns, Saurav Kadavath, Akul Arora, Steven Basart, Eric
  Tang, Dawn Song, and Jacob Steinhardt.
\newblock Measuring mathematical problem solving with the {MATH} dataset.
\newblock In \emph{Proceedings of the Neural Information Processing Systems
  Track on Datasets and Benchmarks}, volume~1, 2021{\natexlab{b}}.
\newblock URL
  \url{https://datasets-benchmarks-proceedings.neurips.cc/paper/2021/hash/be83ab3ecd0db773eb2dc1b0a17836a1-Abstract-round2.html}.

\bibitem[Hu et~al.(2020)Hu, Xiao, Adlam, and Pennington]{hu2020surprising}
Wei Hu, Lechao Xiao, Ben Adlam, and Jeffrey Pennington.
\newblock The surprising simplicity of the early-time learning dynamics of
  neural networks.
\newblock In H.~Larochelle, M.~Ranzato, R.~Hadsell, M.F. Balcan, and H.~Lin
  (eds.), \emph{Advances in Neural Information Processing Systems}, volume~33,
  pp.\  17116--17128. Curran Associates, Inc., 2020.
\newblock URL
  \url{https://proceedings.neurips.cc/paper_files/paper/2020/hash/c6dfc6b7c601ac2978357b7a81e2d7ae-Abstract.html}.

\bibitem[Huang \& Yang(2025)Huang and Yang]{Huang2025imo}
Yichen Huang and Lin~F. Yang.
\newblock Winning gold at {IMO} 2025 with a model-agnostic
  verification-and-refinement pipeline.
\newblock \emph{arXiv preprint arXiv:2507.15855}, 2025.
\newblock \doi{10.48550/arXiv.2507.15855}.
\newblock URL \url{https://arxiv.org/abs/2507.15855}.

\bibitem[Jaech et~al.(2024)]{jaech2024o1}
Aaron Jaech et~al.
\newblock {OpenAI} o1 system card.
\newblock \emph{arXiv preprint arXiv:2412.16720}, 2024.
\newblock \doi{10.48550/arXiv.2412.16720}.
\newblock URL \url{https://arxiv.org/abs/2412.16720}.

\bibitem[Jha et~al.(2026)Jha, Mahajan, Aravindan, Saravanan, Policharla, and
  Gehlot]{jha2026rewarding}
Abha Jha, Akanksha Mahajan, Ashwath~Vaithinathan Aravindan, Praveen Saravanan,
  Sai~Sailaja Policharla, and Sonal~Chaturbhuj Gehlot.
\newblock Rewarding intellectual humility: Learning when not to answer in large
  language models.
\newblock \emph{arXiv preprint arXiv:2601.20126}, 2026.

\bibitem[Jin et~al.(2021)Jin, Pan, Oufattole, Weng, Fang, and
  Szolovits]{jin2021medqa}
Di~Jin, Eileen Pan, Nassim Oufattole, Wei-Hung Weng, Hanyi Fang, and Peter
  Szolovits.
\newblock What disease does this patient have? {A} large-scale open domain
  question answering dataset from medical exams.
\newblock \emph{Applied Sciences}, 11\penalty0 (14):\penalty0 6421, 2021.
\newblock \doi{10.3390/app11146421}.
\newblock URL \url{https://doi.org/10.3390/app11146421}.

\bibitem[Kadavath et~al.(2022)Kadavath, Conerly, Askell, Henighan, Drain,
  Perez, Schiefer, Hatfield-Dodds, DasSarma, Tran-Johnson, Johnston, El-Showk,
  Jones, Elhage, Hume, Chen, Bai, Bowman, Fort, Ganguli, Hernandez, Jacobson,
  Kernion, Kravec, Lovitt, Ndousse, Olsson, Ringer, Amodei, Brown, Clark,
  Joseph, Mann, McCandlish, Olah, and Kaplan]{Kadavath2022MostlyKnow}
Saurav Kadavath, Tom Conerly, Amanda Askell, Tom Henighan, Dawn Drain, Ethan
  Perez, Nicholas Schiefer, Zac Hatfield-Dodds, Nova DasSarma, Eli
  Tran-Johnson, Scott Johnston, Sheer El-Showk, Andy Jones, Nelson Elhage,
  Tristan Hume, Anna Chen, Yuntao Bai, Sam Bowman, Stanislav Fort, Deep
  Ganguli, Danny Hernandez, Josh Jacobson, Jackson Kernion, Shauna Kravec,
  Liane Lovitt, Kamal Ndousse, Catherine Olsson, Sam Ringer, Dario Amodei, Tom
  Brown, Jack Clark, Nicholas Joseph, Ben Mann, Sam McCandlish, Chris Olah, and
  Jared Kaplan.
\newblock Language models (mostly) know what they know.
\newblock arXiv preprint arXiv:2207.05221, 2022.
\newblock URL \url{https://arxiv.org/abs/2207.05221}.

\bibitem[Kalai \& Vempala(2024)Kalai and Vempala]{kalai2024calibrated}
Adam~Tauman Kalai and Santosh~S. Vempala.
\newblock Calibrated language models must hallucinate.
\newblock In \emph{Proceedings of the 56th Annual ACM Symposium on Theory of
  Computing}, STOC '24, pp.\  160--171, New York, NY, USA, 2024. Association
  for Computing Machinery.
\newblock \doi{10.1145/3618260.3649777}.
\newblock URL \url{https://doi.org/10.1145/3618260.3649777}.

\bibitem[Kalai et~al.(2025)Kalai, Nachum, Vempala, and
  Zhang]{Kalai2025WhyHallucinate}
Adam~Tauman Kalai, Ofir Nachum, Santosh~S. Vempala, and Edwin Zhang.
\newblock Why language models hallucinate.
\newblock \emph{arXiv preprint arXiv:2509.04664}, 2025.
\newblock URL \url{https://arxiv.org/abs/2509.04664}.

\bibitem[Kapoor et~al.(2024)Kapoor, Gruver, Roberts, Collins, Pal, Bhatt,
  Weller, Dooley, Goldblum, and Wilson]{Kapoor2024LLMMustBeTaught}
Sanyam Kapoor, Nate Gruver, Manley Roberts, Katherine Collins, Arka Pal, Umang
  Bhatt, Adrian Weller, Samuel Dooley, Micah Goldblum, and Andrew~Gordon
  Wilson.
\newblock Large language models must be taught to know what they don't know.
\newblock In A.~Globerson, L.~Mackey, D.~Belgrave, A.~Fan, U.~Paquet,
  J.~Tomczak, and C.~Zhang (eds.), \emph{Advances in Neural Information
  Processing Systems}, volume~37, pp.\  85932--85972. Curran Associates, Inc.,
  2024.
\newblock \doi{10.52202/079017-2729}.
\newblock URL
  \url{https://proceedings.neurips.cc/paper_files/paper/2024/hash/9c20f16b05f5e5e70fa07e2a4364b80e-Abstract-Conference.html}.

\bibitem[Khalili et~al.(2022)Khalili, You, and Bohannon]{Khalili2022babybear}
Leila Khalili, Yao You, and John Bohannon.
\newblock {BabyBear}: Cheap inference triage for expensive language models.
\newblock \emph{arXiv preprint arXiv:2205.11747}, 2022.
\newblock \doi{10.48550/arXiv.2205.11747}.
\newblock URL \url{https://arxiv.org/abs/2205.11747}.

\bibitem[Kirichenko et~al.(2025)Kirichenko, Ibrahim, Chaudhuri, and
  Bell]{kirichenko2025abstentionbenchreasoningllmsfail}
Polina Kirichenko, Mark Ibrahim, Kamalika Chaudhuri, and Samuel~J. Bell.
\newblock {AbstentionBench}: Reasoning {LLMs} fail on unanswerable questions.
\newblock In \emph{Advances in Neural Information Processing Systems (Datasets
  and Benchmarks Track)}, volume~38. Curran Associates, Inc., 2025.
\newblock \doi{10.52202/085713-5729}.
\newblock URL
  \url{https://proceedings.neurips.cc/paper_files/paper/2025/hash/fb122bfc3f0127a94ded048b5b03496f-Abstract-Datasets_and_Benchmarks_Track.html}.

\bibitem[Kolawole et~al.(2025)Kolawole, Dennis, Talwalkar, and
  Smith]{Kolawole2025abc}
Steven Kolawole, Don Dennis, Ameet Talwalkar, and Virginia Smith.
\newblock Agreement-based cascading for efficient inference.
\newblock \emph{Transactions on Machine Learning Research}, 2025.
\newblock ISSN 2835-8856.
\newblock URL \url{https://openreview.net/forum?id=jn9B7LMlzk}.

\bibitem[Krishnan et~al.(2024)Krishnan, Khanna, and
  Tickoo]{Krishnan2024EnhancingTrust}
Ranganath Krishnan, Piyush Khanna, and Omesh Tickoo.
\newblock Enhancing trust in large language models via uncertainty-calibrated
  fine-tuning.
\newblock arXiv preprint arXiv:2412.02904, 2024.
\newblock URL \url{https://arxiv.org/abs/2412.02904}.

\bibitem[Kwiatkowski et~al.(2019)Kwiatkowski, Palomaki, Redfield, Collins,
  Parikh, Alberti, Epstein, Polosukhin, Devlin, Lee, Toutanova, Jones, Kelcey,
  Chang, Dai, Uszkoreit, Le, and Petrov]{kwiatkowski-etal-2019-natural}
Tom Kwiatkowski, Jennimaria Palomaki, Olivia Redfield, Michael Collins, Ankur
  Parikh, Chris Alberti, Danielle Epstein, Illia Polosukhin, Jacob Devlin,
  Kenton Lee, Kristina Toutanova, Llion Jones, Matthew Kelcey, Ming-Wei Chang,
  Andrew~M. Dai, Jakob Uszkoreit, Quoc Le, and Slav Petrov.
\newblock {Natural Questions}: A benchmark for question answering research.
\newblock \emph{Transactions of the Association for Computational Linguistics},
  7:\penalty0 453--466, 2019.
\newblock \doi{10.1162/tacl_a_00276}.
\newblock URL \url{https://aclanthology.org/Q19-1026/}.

\bibitem[Lambert et~al.(2025)Lambert, Morrison, Pyatkin, Huang, Ivison,
  Brahman, Miranda, Liu, Dziri, Lyu, Gu, Malik, Graf, Hwang, Yang, Le~Bras,
  Tafjord, Wilhelm, Soldaini, Smith, Wang, Dasigi, and
  Hajishirzi]{lambert2025tulu3}
Nathan Lambert, Jacob Morrison, Valentina Pyatkin, Shengyi Huang, Hamish
  Ivison, Faeze Brahman, Lester James~Validad Miranda, Alisa Liu, Nouha Dziri,
  Xinxi Lyu, Yuling Gu, Saumya Malik, Victoria Graf, Jena~D. Hwang, Jiangjiang
  Yang, Ronan Le~Bras, Oyvind Tafjord, Christopher Wilhelm, Luca Soldaini,
  Noah~A. Smith, Yizhong Wang, Pradeep Dasigi, and Hannaneh Hajishirzi.
\newblock Tulu 3: Pushing frontiers in open language model post-training.
\newblock In \emph{Conference on Language Modeling (COLM)}, 2025.
\newblock URL \url{https://openreview.net/forum?id=i1uGbfHHpH}.

\bibitem[Leng et~al.(2025)Leng, Huang, Zhu, and
  Huang]{Leng2025TamingOverconfidence}
Jixuan Leng, Chengsong Huang, Banghua Zhu, and Jiaxin Huang.
\newblock Taming overconfidence in {LLM}s: Reward calibration in {RLHF}.
\newblock In \emph{The Thirteenth International Conference on Learning
  Representations (ICLR)}, 2025.
\newblock URL \url{https://arxiv.org/abs/2410.09724}.

\bibitem[Lewkowycz et~al.(2022)Lewkowycz, Andreassen, Dohan, Dyer, Michalewski,
  Ramasesh, Slone, Anil, Schlag, Gutman-Solo, Wu, Neyshabur, Gur-Ari, and
  Misra]{Lewkowycz2022Minerva}
Aitor Lewkowycz, Anders Andreassen, David Dohan, Ethan Dyer, Henryk
  Michalewski, Vinay~V. Ramasesh, Ambrose Slone, Cem Anil, Imanol Schlag, Theo
  Gutman-Solo, Yuhuai Wu, Behnam Neyshabur, Guy Gur-Ari, and Vedant Misra.
\newblock Solving quantitative reasoning problems with language models.
\newblock In S.~Koyejo, S.~Mohamed, A.~Agarwal, D.~Belgrave, K.~Cho, and A.~Oh
  (eds.), \emph{Advances in Neural Information Processing Systems}, volume~35,
  pp.\  3843--3857. Curran Associates, Inc., 2022.
\newblock \doi{10.52202/068431-0278}.
\newblock URL
  \url{https://proceedings.neurips.cc/paper_files/paper/2022/hash/18abbeef8cfe9203fdf9053c9c4fe191-Abstract-Conference.html}.

\bibitem[Li et~al.(2024)Li, Balachandran, Feng, Ilgen, Pierson, Koh, and
  Tsvetkov]{li2024mediq}
Shuyue~Stella Li, Vidhisha Balachandran, Shangbin Feng, Jonathan~S. Ilgen, Emma
  Pierson, Pang~Wei Koh, and Yulia Tsvetkov.
\newblock {MediQ}: Question-asking {LLMs} and a benchmark for reliable
  interactive clinical reasoning.
\newblock In \emph{Advances in Neural Information Processing Systems},
  volume~37, pp.\  28858--28888, 2024.
\newblock \doi{10.52202/079017-0908}.
\newblock URL
  \url{https://proceedings.neurips.cc/paper_files/paper/2024/hash/32b80425554e081204e5988ab1c97e9a-Abstract-Conference.html}.

\bibitem[Lightman et~al.(2024)Lightman, Kosaraju, Burda, Edwards, Baker, Lee,
  Leike, Schulman, Sutskever, and Cobbe]{Lightman2024LetsVerify}
Hunter Lightman, Vineet Kosaraju, Yuri Burda, Harrison Edwards, Bowen Baker,
  Teddy Lee, Jan Leike, John Schulman, Ilya Sutskever, and Karl Cobbe.
\newblock Let's verify step by step.
\newblock In \emph{The Twelfth International Conference on Learning
  Representations}, 2024.
\newblock URL \url{https://openreview.net/forum?id=v8L0pN6EOi}.

\bibitem[Lin et~al.(2022{\natexlab{a}})Lin, Hilton, and
  Evans]{Lin2022UncertaintyInWords}
Stephanie Lin, Jacob Hilton, and Owain Evans.
\newblock Teaching models to express their uncertainty in words.
\newblock \emph{Transactions on Machine Learning Research}, 2022{\natexlab{a}}.
\newblock ISSN 2835-8856.
\newblock URL \url{https://openreview.net/forum?id=8s8K2UZGTZ}.

\bibitem[Lin et~al.(2022{\natexlab{b}})Lin, Hilton, and
  Evans]{lin-etal-2022-truthfulqa}
Stephanie Lin, Jacob Hilton, and Owain Evans.
\newblock {T}ruthful{QA}: Measuring how models mimic human falsehoods.
\newblock In Smaranda Muresan, Preslav Nakov, and Aline Villavicencio (eds.),
  \emph{Proceedings of the 60th Annual Meeting of the Association for
  Computational Linguistics (Volume 1: Long Papers)}, pp.\  3214--3252, Dublin,
  Ireland, May 2022{\natexlab{b}}. Association for Computational Linguistics.
\newblock \doi{10.18653/v1/2022.acl-long.229}.
\newblock URL \url{https://aclanthology.org/2022.acl-long.229/}.

\bibitem[Liu et~al.(2023)Liu, Yao, Ton, Zhang, Guo, Cheng, Klochkov, Taufiq,
  and Li]{liu2024trustworthy}
Yang Liu, Yuanshun Yao, Jean-Francois Ton, Xiaoying Zhang, Ruocheng Guo, Hao
  Cheng, Yegor Klochkov, Muhammad~Faaiz Taufiq, and Hang Li.
\newblock Trustworthy {LLMs}: a survey and guideline for evaluating large
  language models' alignment.
\newblock arXiv preprint arXiv:2308.05374, 2023.
\newblock URL \url{https://arxiv.org/abs/2308.05374}.

\bibitem[Liu et~al.(2025)Liu, Chen, Li, Qi, Pang, Du, Lee, and
  Lin]{anonymous2025drGRPO}
Zichen Liu, Changyu Chen, Wenjun Li, Penghui Qi, Tianyu Pang, Chao Du, Wee~Sun
  Lee, and Min Lin.
\newblock Understanding {R1-Zero}-like training: A critical perspective.
\newblock In \emph{Conference on Language Modeling (COLM)}, 2025.
\newblock URL \url{https://openreview.net/forum?id=5PAF7PAY2Y}.

\bibitem[Madhusudhan et~al.(2025)Madhusudhan, Madhusudhan, Yadav, and
  Hashemi]{madhusudhan-etal-2025-llms}
Nishanth Madhusudhan, Sathwik~Tejaswi Madhusudhan, Vikas Yadav, and Masoud
  Hashemi.
\newblock Do {LLM}s know when to {NOT} answer? {I}nvestigating abstention
  abilities of large language models.
\newblock In \emph{Proceedings of the 31st International Conference on
  Computational Linguistics}, pp.\  9329--9345, Abu Dhabi, UAE, January 2025.
  Association for Computational Linguistics.
\newblock URL \url{https://aclanthology.org/2025.coling-main.627/}.

\bibitem[Manakul et~al.(2023)Manakul, Liusie, and
  Gales]{manakul-etal-2023-selfcheckgpt}
Potsawee Manakul, Adian Liusie, and Mark Gales.
\newblock {S}elf{C}heck{GPT}: Zero-resource black-box hallucination detection
  for generative large language models.
\newblock In Houda Bouamor, Juan Pino, and Kalika Bali (eds.),
  \emph{Proceedings of the 2023 Conference on Empirical Methods in Natural
  Language Processing}, pp.\  9004--9017, Singapore, December 2023. Association
  for Computational Linguistics.
\newblock \doi{10.18653/v1/2023.emnlp-main.557}.
\newblock URL \url{https://aclanthology.org/2023.emnlp-main.557}.

\bibitem[Mazeika et~al.(2024)Mazeika, Phan, Yin, Zou, Wang, Mu, Sakhaee, Li,
  Basart, Li, Forsyth, and Hendrycks]{mazeika2024harmbench}
Mantas Mazeika, Long Phan, Xuwang Yin, Andy Zou, Zifan Wang, Norman Mu, Elham
  Sakhaee, Nathaniel Li, Steven Basart, Bo~Li, David Forsyth, and Dan
  Hendrycks.
\newblock {HarmBench}: A standardized evaluation framework for automated red
  teaming and robust refusal.
\newblock In \emph{Proceedings of the 41st International Conference on Machine
  Learning}, volume 235 of \emph{Proceedings of Machine Learning Research},
  pp.\  35181--35224. PMLR, 2024.
\newblock URL \url{https://proceedings.mlr.press/v235/mazeika24a.html}.

\bibitem[Muennighoff et~al.(2025)Muennighoff, Yang, Shi, Li, Li, Hajishirzi,
  Zettlemoyer, Liang, Cand{\`e}s, and Hashimoto]{Muennighoff2025s1}
Niklas Muennighoff, Zitong Yang, Weijia Shi, Xiang~Lisa Li, Fei-Fei Li,
  Hannaneh Hajishirzi, Luke Zettlemoyer, Percy Liang, Emmanuel Cand{\`e}s, and
  Tatsunori Hashimoto.
\newblock {s1}: Simple test-time scaling.
\newblock In \emph{Proceedings of the 2025 Conference on Empirical Methods in
  Natural Language Processing}, pp.\  20275--20321, Suzhou, China, November
  2025. Association for Computational Linguistics.
\newblock \doi{10.18653/v1/2025.emnlp-main.1025}.
\newblock URL \url{https://aclanthology.org/2025.emnlp-main.1025/}.

\bibitem[Ouyang et~al.(2022)Ouyang, Wu, Jiang, Almeida, Wainwright, Mishkin,
  Zhang, Agarwal, Slama, Ray, Schulman, Hilton, Kelton, Miller, Simens, Askell,
  Welinder, Christiano, Leike, and Lowe]{ouyang2022instructgpt}
Long Ouyang, Jeff Wu, Xu~Jiang, Diogo Almeida, Carroll~L. Wainwright, Pamela
  Mishkin, Chong Zhang, Sandhini Agarwal, Katarina Slama, Alex Ray, John
  Schulman, Jacob Hilton, Fraser Kelton, Luke Miller, Maddie Simens, Amanda
  Askell, Peter Welinder, Paul~F. Christiano, Jan Leike, and Ryan Lowe.
\newblock Training language models to follow instructions with human feedback.
\newblock In S.~Koyejo, S.~Mohamed, A.~Agarwal, D.~Belgrave, K.~Cho, and A.~Oh
  (eds.), \emph{Advances in Neural Information Processing Systems}, volume~35,
  pp.\  27730--27744. Curran Associates, Inc., 2022.
\newblock \doi{10.52202/068431-2011}.
\newblock URL
  \url{https://proceedings.neurips.cc/paper_files/paper/2022/hash/b1efde53be364a73914f58805a001731-Abstract-Conference.html}.

\bibitem[Pal et~al.(2022)Pal, Umapathi, and Sankarasubbu]{pal2022medmcqa}
Ankit Pal, Logesh~Kumar Umapathi, and Malaikannan Sankarasubbu.
\newblock Med{MCQA}: A large-scale multi-subject multi-choice dataset for
  medical domain question answering.
\newblock In Gerardo Flores, George~H. Chen, Tom Pollard, Joyce~C. Ho, and
  Tristan Naumann (eds.), \emph{Proceedings of the Conference on Health,
  Inference, and Learning}, volume 174 of \emph{Proceedings of Machine Learning
  Research}, pp.\  248--260. PMLR, 2022.
\newblock URL \url{https://proceedings.mlr.press/v174/pal22a.html}.

\bibitem[{Qwen Team}(2025)]{qwenTeam2025qwen3}
{Qwen Team}.
\newblock {Qwen3} technical report.
\newblock \emph{arXiv preprint arXiv:2505.09388}, 2025.
\newblock URL \url{https://arxiv.org/abs/2505.09388}.

\bibitem[Rabanser et~al.(2025)Rabanser, Rauschmayr, Kulshrestha, Poklukar,
  Jitkrittum, Augenstein, Wang, and Tombari]{Rabanser2025gatekeeper}
Stephan Rabanser, Nathalie Rauschmayr, Achin Kulshrestha, Petra Poklukar,
  Wittawat Jitkrittum, Sean Augenstein, Congchao Wang, and Federico Tombari.
\newblock Gatekeeper: Improving model cascades through confidence tuning.
\newblock In \emph{Advances in Neural Information Processing Systems},
  volume~38. Curran Associates, Inc., 2025.
\newblock \doi{10.52202/085713-0656}.
\newblock URL
  \url{https://proceedings.neurips.cc/paper_files/paper/2025/file/1c23f6ce43a7db43dfe31f26ceabfbfe-Paper-Conference.pdf}.

\bibitem[Rajpurkar et~al.(2018)Rajpurkar, Jia, and
  Liang]{rajpurkar-etal-2018-know}
Pranav Rajpurkar, Robin Jia, and Percy Liang.
\newblock Know what you don{'}t know: Unanswerable questions for {SQ}u{AD}.
\newblock In Iryna Gurevych and Yusuke Miyao (eds.), \emph{Proceedings of the
  56th Annual Meeting of the Association for Computational Linguistics (Volume
  2: Short Papers)}, pp.\  784--789, Melbourne, Australia, July 2018.
  Association for Computational Linguistics.
\newblock \doi{10.18653/v1/P18-2124}.
\newblock URL \url{https://aclanthology.org/P18-2124/}.

\bibitem[Rein et~al.(2024)Rein, Hou, Cooper~Stickland, Petty, Pang, Dirani,
  Michael, and Bowman]{Rein2024GPQA}
David Rein, Betty~Li Hou, Asa Cooper~Stickland, Jackson Petty, Richard~Yuanzhe
  Pang, Julien Dirani, Julian Michael, and Samuel~R. Bowman.
\newblock {GPQA}: A graduate-level {Google}-proof {Q\&A} benchmark.
\newblock In \emph{First Conference on Language Modeling (COLM)}, 2024.
\newblock URL \url{https://openreview.net/forum?id=Ti67584b98}.

\bibitem[Rivest \& Sloan(1988)Rivest and Sloan]{RS88b}
Ronald~L. Rivest and Robert~H. Sloan.
\newblock Learning complicated concepts reliably and usefully.
\newblock In \emph{Proceedings of the Seventh National Conference on Artificial
  Intelligence (AAAI-88)}, pp.\  635--640, Saint Paul, Minnesota, USA, August
  1988. AAAI Press.
\newblock URL \url{https://people.csail.mit.edu/rivest/pubs/RS88b.pdf}.

\bibitem[Saadat et~al.(2024)Saadat, Binte~Sogir, Chowdhury, and
  Aziz]{Saadat2024WhenNotToAnswer}
Asir Saadat, Tasmia Binte~Sogir, Md~Taukir~Azam Chowdhury, and Syem Aziz.
\newblock When not to answer: Evaluating prompts on {GPT} models for effective
  abstention in unanswerable math word problems.
\newblock \emph{arXiv preprint arXiv:2410.13029}, 2024.
\newblock URL \url{https://arxiv.org/abs/2410.13029}.

\bibitem[Sayedi et~al.(2010)Sayedi, Zadimoghaddam, and Blum]{sayedi2010trading}
Amin Sayedi, Morteza Zadimoghaddam, and Avrim Blum.
\newblock Trading off mistakes and don't-know predictions.
\newblock In J.~Lafferty, C.~Williams, J.~Shawe-Taylor, R.~Zemel, and
  A.~Culotta (eds.), \emph{Advances in Neural Information Processing Systems},
  volume~23. Curran Associates, Inc., 2010.
\newblock URL
  \url{https://proceedings.neurips.cc/paper/2010/hash/286674e3082feb7e5afb92777e48821f-Abstract.html}.

\bibitem[Schulman et~al.(2017)Schulman, Wolski, Dhariwal, Radford, and
  Klimov]{schulman2017ppo}
John Schulman, Filip Wolski, Prafulla Dhariwal, Alec Radford, and Oleg Klimov.
\newblock Proximal policy optimization algorithms.
\newblock \emph{arXiv preprint arXiv:1707.06347}, 2017.
\newblock \doi{10.48550/arXiv.1707.06347}.
\newblock URL \url{https://arxiv.org/abs/1707.06347}.

\bibitem[Shao et~al.(2024)Shao, Wang, Zhu, Xu, Song, Bi, Zhang, Zhang, Li, Wu,
  and Guo]{Shao2024DeepSeekMath}
Zhihong Shao, Peiyi Wang, Qihao Zhu, Runxin Xu, Junxiao Song, Xiao Bi, Haowei
  Zhang, Mingchuan Zhang, Y.~K. Li, Y.~Wu, and Daya Guo.
\newblock {DeepSeekMath}: Pushing the limits of mathematical reasoning in open
  language models.
\newblock \emph{arXiv preprint arXiv:2402.03300}, 2024.
\newblock URL \url{https://arxiv.org/abs/2402.03300}.

\bibitem[Sheng et~al.(2025)Sheng, Zhang, Ye, Wu, Zhang, Zhang, Peng, Lin, and
  Wu]{sheng2024hybridflow}
Guangming Sheng, Chi Zhang, Zilingfeng Ye, Xibin Wu, Wang Zhang, Ru~Zhang,
  Yanghua Peng, Haibin Lin, and Chuan Wu.
\newblock {HybridFlow}: A flexible and efficient {RLHF} framework.
\newblock In \emph{Proceedings of the Twentieth European Conference on Computer
  Systems (EuroSys '25)}, pp.\  1279--1297. ACM, 2025.
\newblock \doi{10.1145/3689031.3696075}.
\newblock URL \url{https://doi.org/10.1145/3689031.3696075}.

\bibitem[Srivastava et~al.(2023)Srivastava, Rastogi, Rao, Shoeb, Abid, Fisch,
  Brown, Santoro, Gupta, Garriga-Alonso, et~al.]{srivastava2022beyond}
Aarohi Srivastava, Abhinav Rastogi, Abhishek Rao, Abu Awal~Md Shoeb, Abubakar
  Abid, Adam Fisch, Adam~R. Brown, Adam Santoro, Aditya Gupta, Adri{\`a}
  Garriga-Alonso, et~al.
\newblock Beyond the imitation game: Quantifying and extrapolating the
  capabilities of language models.
\newblock \emph{Transactions on Machine Learning Research}, 2023.
\newblock ISSN 2835-8856.
\newblock URL \url{https://openreview.net/forum?id=uyTL5Bvosj}.

\bibitem[Sun et~al.(2024)Sun, Yin, Guo, Wu, Qiu, and
  Zhao]{sun-etal-2024-benchmarking}
YuHong Sun, Zhangyue Yin, Qipeng Guo, Jiawen Wu, Xipeng Qiu, and Hui Zhao.
\newblock Benchmarking hallucination in large language models based on
  unanswerable math word problem.
\newblock In Nicoletta Calzolari, Min-Yen Kan, Veronique Hoste, Alessandro
  Lenci, Sakriani Sakti, and Nianwen Xue (eds.), \emph{Proceedings of the 2024
  Joint International Conference on Computational Linguistics, Language
  Resources and Evaluation (LREC-COLING 2024)}, pp.\  2178--2188, Torino,
  Italia, May 2024. ELRA and ICCL.
\newblock URL \url{https://aclanthology.org/2024.lrec-main.196/}.

\bibitem[Thirunavukarasu et~al.(2023)Thirunavukarasu, Ting, Elangovan,
  Gutierrez, Tan, and Ting]{Thirunavukarasu2023LLMMedicine}
Arun~James Thirunavukarasu, Darren Shu~Jeng Ting, Kabilan Elangovan, Laura
  Gutierrez, Ting~Fang Tan, and Daniel Shu~Wei Ting.
\newblock Large language models in medicine.
\newblock \emph{Nature Medicine}, 29\penalty0 (8):\penalty0 1930--1940, 2023.
\newblock \doi{10.1038/s41591-023-02448-8}.
\newblock URL \url{https://www.nature.com/articles/s41591-023-02448-8}.

\bibitem[Tian et~al.(2023)Tian, Mitchell, Zhou, Sharma, Rafailov, Yao, Finn,
  and Manning]{Tian2023JustAskForCalibration}
Katherine Tian, Eric Mitchell, Allan Zhou, Archit Sharma, Rafael Rafailov,
  Huaxiu Yao, Chelsea Finn, and Christopher~D. Manning.
\newblock Just ask for calibration: Strategies for eliciting calibrated
  confidence scores from language models fine-tuned with human feedback.
\newblock In Houda Bouamor, Juan Pino, and Kalika Bali (eds.),
  \emph{Proceedings of the 2023 Conference on Empirical Methods in Natural
  Language Processing}, pp.\  5433--5442, Singapore, December 2023. Association
  for Computational Linguistics.
\newblock \doi{10.18653/v1/2023.emnlp-main.330}.
\newblock URL \url{https://aclanthology.org/2023.emnlp-main.330/}.

\bibitem[Tomani et~al.(2024)Tomani, Chaudhuri, Evtimov, Cremers, and
  Ibrahim]{Tomani2024UncertaintyAbstention}
Christian Tomani, Kamalika Chaudhuri, Ivan Evtimov, Daniel Cremers, and Mark
  Ibrahim.
\newblock Uncertainty-based abstention in {LLMs} improves safety and reduces
  hallucinations.
\newblock \emph{arXiv preprint arXiv:2404.10960}, 2024.
\newblock URL \url{https://arxiv.org/abs/2404.10960}.

\bibitem[Tong et~al.(2025)Tong, Xia, Shang, and Zhou]{tong2025measuring}
Bingkui Tong, Jiaer Xia, Sifeng Shang, and Kaiyang Zhou.
\newblock Measuring epistemic humility in multimodal large language models.
\newblock \emph{arXiv preprint arXiv:2509.09658}, 2025.
\newblock URL \url{https://arxiv.org/abs/2509.09658}.

\bibitem[Vapnik(1998)]{vapnik1998statisticallearning}
Vladimir~N. Vapnik.
\newblock \emph{Statistical Learning Theory}.
\newblock Wiley, New York, 1998.

\bibitem[Wang et~al.(2024)Wang, Augenstein, Rush, Jitkrittum, Narasimhan,
  Rawat, Menon, and Go]{Wang2024cascadeaware}
Congchao Wang, Sean Augenstein, Keith Rush, Wittawat Jitkrittum, Harikrishna
  Narasimhan, Ankit~Singh Rawat, Aditya~Krishna Menon, and Alec Go.
\newblock Cascade-aware training of language models.
\newblock \emph{arXiv preprint arXiv:2406.00060}, 2024.
\newblock \doi{10.48550/arXiv.2406.00060}.
\newblock URL \url{https://arxiv.org/abs/2406.00060}.

\bibitem[Wei et~al.(2026)Wei, Yang, Sun, Wang, Shao, Chen, Kachuee, Gollapudi,
  Liao, Scheffer, Wanga, Kumar, Meng, Yih, and Dong]{wei2025truthrl}
Zhepei Wei, Xiao Yang, Kai Sun, Jiaqi Wang, Rulin Shao, Jingxiang Chen,
  Mohammad Kachuee, Teja Gollapudi, Yiwei Liao, Nicolas Scheffer, Rakesh Wanga,
  Anuj Kumar, Yu~Meng, Wen-tau Yih, and Xin~Luna Dong.
\newblock {TruthRL}: Incentivizing truthful {LLM}s via reinforcement learning.
\newblock In \emph{Proceedings of the 43rd International Conference on Machine
  Learning (ICML)}, 2026.
\newblock URL \url{https://arxiv.org/abs/2509.25760}.

\bibitem[Wen et~al.(2025)Wen, Yao, Feng, Xu, Tsvetkov, Howe, and
  Wang]{wen-etal-2025-know}
Bingbing Wen, Jihan Yao, Shangbin Feng, Chenjun Xu, Yulia Tsvetkov, Bill Howe,
  and Lucy~Lu Wang.
\newblock Know your limits: A survey of abstention in large language models.
\newblock \emph{Transactions of the Association for Computational Linguistics},
  13:\penalty0 529--556, 2025.
\newblock \doi{10.1162/tacl_a_00754}.
\newblock URL \url{https://doi.org/10.1162/tacl_a_00754}.

\bibitem[Wu et~al.(2025)Wu, Tam, Lin, Chen, and Lee]{wu2025answer}
Cheng-Kuang Wu, Zhi~Rui Tam, Chieh-Yen Lin, Yun-Nung Chen, and Hung-yi Lee.
\newblock Answer, refuse, or guess? {I}nvestigating risk-aware decision making
  in language models.
\newblock \emph{arXiv preprint arXiv:2503.01332}, 2025.
\newblock URL \url{https://arxiv.org/abs/2503.01332}.

\bibitem[Wu et~al.(2026)Wu, Liu, Zeng, Zhan, Cai, and Huang]{wu2025bcrl}
Jiayun Wu, Jiashuo Liu, Zhiyuan Zeng, Tianyang Zhan, Tianle Cai, and Wenhao
  Huang.
\newblock Mitigating {LLM} hallucination via behaviorally calibrated
  reinforcement learning.
\newblock arXiv preprint arXiv:2512.19920, 2026.
\newblock URL \url{https://arxiv.org/abs/2512.19920}.

\bibitem[Wu et~al.(2023)Wu, Irsoy, Lu, Dabravolski, Dredze, Gehrmann, Kambadur,
  Rosenberg, and Mann]{Wu2023BloombergGPT}
Shijie Wu, Ozan Irsoy, Steven Lu, Vadim Dabravolski, Mark Dredze, Sebastian
  Gehrmann, Prabhanjan Kambadur, David Rosenberg, and Gideon Mann.
\newblock {BloombergGPT}: A large language model for finance.
\newblock \emph{arXiv preprint arXiv:2303.17564}, 2023.
\newblock URL \url{https://arxiv.org/abs/2303.17564}.

\bibitem[Xie et~al.(2025{\natexlab{a}})Xie, Huang, Zhang, Yu, Chen, Lin, Li,
  Ghazi, and Kumar]{xie2024memorization}
Chulin Xie, Yangsibo Huang, Chiyuan Zhang, Da~Yu, Xinyun Chen, Bill~Yuchen Lin,
  Bo~Li, Badih Ghazi, and Ravi Kumar.
\newblock On memorization of large language models in logical reasoning.
\newblock In \emph{Proceedings of the 14th International Joint Conference on
  Natural Language Processing and the 4th Conference of the Asia-Pacific
  Chapter of the Association for Computational Linguistics}, pp.\  2742--2785,
  Mumbai, India, 2025{\natexlab{a}}. The Asian Federation of Natural Language
  Processing and The Association for Computational Linguistics.
\newblock \doi{10.18653/v1/2025.ijcnlp-long.148}.
\newblock URL \url{https://aclanthology.org/2025.ijcnlp-long.148/}.

\bibitem[Xie et~al.(2025{\natexlab{b}})Xie, Gao, Ren, Luo, Hong, Dai, Zhou,
  Qiu, Wu, and Luo]{xie2025logicRL}
Tian Xie, Zitian Gao, Qingnan Ren, Haoming Luo, Yuqian Hong, Bryan Dai, Joey
  Zhou, Kai Qiu, Zhirong Wu, and Chong Luo.
\newblock {Logic-RL}: Unleashing {LLM} reasoning with rule-based reinforcement
  learning.
\newblock arXiv preprint arXiv:2502.14768, 2025{\natexlab{b}}.
\newblock URL \url{https://arxiv.org/abs/2502.14768}.

\bibitem[Xiong et~al.(2024)Xiong, Hu, Lu, Li, Fu, He, and Hooi]{xiong2024can}
Miao Xiong, Zhiyuan Hu, Xinyang Lu, Yifei Li, Jie Fu, Junxian He, and Bryan
  Hooi.
\newblock Can {LLM}s express their uncertainty? {An} empirical evaluation of
  confidence elicitation in {LLM}s.
\newblock In \emph{The Twelfth International Conference on Learning
  Representations}, 2024.
\newblock URL \url{https://openreview.net/forum?id=gjeQKFxFpZ}.

\bibitem[Xue et~al.(2026)Xue, Greenewald, Mroueh, and
  Mirzasoleiman]{xue2025verify}
Yihao Xue, Kristjan Greenewald, Youssef Mroueh, and Baharan Mirzasoleiman.
\newblock Verify when uncertain: Beyond self-consistency in black box
  hallucination detection.
\newblock \emph{Transactions on Machine Learning Research}, 2026.
\newblock ISSN 2835-8856.
\newblock URL \url{https://openreview.net/forum?id=6tlLISSgiu}.

\bibitem[Yang et~al.(2024)Yang, Wang, Xu, Zhang, and Bian]{Yang2024CanWeTrust}
Haoyan Yang, Yixuan Wang, Xingyin Xu, Hanyuan Zhang, and Yirong Bian.
\newblock Can we trust {LLM}s? {M}itigate overconfidence bias in {LLM}s through
  knowledge transfer.
\newblock \emph{arXiv preprint arXiv:2405.16856}, 2024.
\newblock URL \url{https://arxiv.org/abs/2405.16856}.

\bibitem[Yao et~al.(2025)Yao, Liu, Chen, Chen, Fang, Hou, Li, and
  Chua]{yao-etal-2025-are}
Zijun Yao, Yantao Liu, Yanxu Chen, Jianhui Chen, Junfeng Fang, Lei Hou, Juanzi
  Li, and Tat-Seng Chua.
\newblock Are reasoning models more prone to hallucination?
\newblock \emph{arXiv preprint arXiv:2505.23646}, 2025.
\newblock URL \url{https://arxiv.org/abs/2505.23646}.

\bibitem[Yunis et~al.(2026)Yunis, Huo, Gunasekara, and
  Contractor]{yunis2024reducing}
David Yunis, Siyu Huo, Chulaka Gunasekara, and Danish Contractor.
\newblock Reducing the scope of language models.
\newblock In \emph{Proceedings of the AAAI Conference on Artificial
  Intelligence}, volume~40, pp.\  34548--34556, 2026.
\newblock \doi{10.1609/aaai.v40i41.40754}.
\newblock URL \url{https://ojs.aaai.org/index.php/AAAI/article/view/40754}.

\bibitem[Zhai et~al.(2026)Zhai, Liang, and Kang]{zhai2026abstain}
Haotian Zhai, Jingcheng Liang, and Dongyeop Kang.
\newblock Abstain-{R1}: Calibrated abstention and post-refusal clarification
  via verifiable {RL}.
\newblock In \emph{Findings of the Association for Computational Linguistics:
  ACL 2026}, pp.\  19674--19695, San Diego, California, United States, 2026.
  Association for Computational Linguistics.
\newblock \doi{10.18653/v1/2026.findings-acl.985}.
\newblock URL \url{https://aclanthology.org/2026.findings-acl.985/}.

\bibitem[Zhang et~al.(2026)Zhang, Wang, Peng, Go, and Liu]{Zhang2024p3defer}
Kai Zhang, Congchao Wang, Liqian Peng, Alec Go, and Xiaozhong Liu.
\newblock Privacy-preserved {LLM} cascade via {CoT}-enhanced policy learning.
\newblock In \emph{Proceedings of the Nineteenth ACM International Conference
  on Web Search and Data Mining (WSDM '26)}, pp.\  955--964. ACM, 2026.
\newblock \doi{10.1145/3773966.3777920}.
\newblock URL \url{https://doi.org/10.1145/3773966.3777920}.

\bibitem[Zheng et~al.(2024)Zheng, Yin, Xie, Sun, Huang, Yu, Cao, Kozyrakis,
  Stoica, Gonzalez, Barrett, and
  Sheng]{zheng2024sglangefficientexecutionstructured}
Lianmin Zheng, Liangsheng Yin, Zhiqiang Xie, Chuyue Sun, Jeff Huang, Cody~Hao
  Yu, Shiyi Cao, Christos Kozyrakis, Ion Stoica, Joseph~E. Gonzalez, Clark
  Barrett, and Ying Sheng.
\newblock {SGLang}: Efficient execution of structured language model programs.
\newblock In A.~Globerson, L.~Mackey, D.~Belgrave, A.~Fan, U.~Paquet,
  J.~Tomczak, and C.~Zhang (eds.), \emph{Advances in Neural Information
  Processing Systems}, volume~37, pp.\  62557--62583. Curran Associates, Inc.,
  2024.
\newblock \doi{10.52202/079017-2000}.
\newblock URL
  \url{https://proceedings.neurips.cc/paper_files/paper/2024/hash/724be4472168f31ba1c9ac630f15dec8-Abstract-Conference.html}.

\end{thebibliography}
